\documentclass[12pt]{article}
\usepackage[utf8]{inputenc}

\usepackage[margin=1.0in, top=0.5in]{geometry}

\usepackage{amsmath,amsfonts,amssymb}
\usepackage{tikz}
\usepackage{newtxmath}
\DeclareMathAlphabet{\mathpzc}{T1}{pzc}{m}{it}

\usepackage{import}

\usepackage[explicit]{titlesec}

\usepackage{titling}
\usepackage{enumitem}
\setlist[itemize]{noitemsep, topsep=0pt}

\usepackage{caption}
\usepackage{changepage}

\usepackage[
    backend=biber,
    sortcites=true,
    style=nature,
    maxbibnames=99,
]{biblatex}

\usepackage{xr}

\usepackage{array}
\usepackage{siunitx}
\usepackage{colortbl}
\usepackage{multirow}
\usepackage{hhline}
\usepackage{calc}
\usepackage{tabularx}
\usepackage{threeparttable}
\usepackage{adjustbox}

\usepackage{lscape}

\usepackage{graphicx}

\usepackage{amsmath}
\usepackage{amssymb}
\DeclareMathAlphabet{\mathpzc}{T1}{pzc}{m}{it}
\def\given{\mid}
\usepackage{bm}

\newcommand{\vect}[1]{\boldsymbol{#1}}

\newcommand{\beginsupplement}{%
        \setcounter{table}{0}
        \renewcommand{\thetable}{S\arabic{table}}%
        \setcounter{figure}{0}
        \renewcommand{\thefigure}{S\arabic{figure}}%
     }

\def\ie{i.e.,~}

\def\SI{Supporting Information}

\usepackage{times}

\usepackage{authblk}

\begin{document}
\title{Unsupervised embedding of trajectories captures the latent structure of scientific migration} 

\author[a,b,*]{Dakota Murray}
\author[a,c,d,*]{Jisung Yoon}
\author[a,*]{Sadamori Kojaku}
\author[e,f]{Rodrigo Costas}
\author[g,h]{Woo-Sung Jung}
\author[a]{Staša Milojević}
\author[a,i]{Yong-Yeol Ahn}

\affil[a]{Center for Complex Networks and Systems Research, Luddy School of Informatics, Computing, and Engineering, Indiana University, Bloomington, IN 47408, US}
\affil[b]{Network Science Institute at Northeastern University, Boston, MA, 02115, US}
\affil[c]{Kellogg School of Management at Northwestern University, Evanston, IL 60208, US}
\affil[d]{Northwestern Institute on Complex Systems, Evanston, IL 60208, USA.}
\affil[e]{Centre for Science and Technology Studies, Leiden University, 2300 AX Leiden, NL}
\affil[f]{DST-NRF Centre of Excellence in Scientometrics and Science, Technology and Innovation Policy, Stellenbosch University, Stellenbosch, ZA}
\affil[g]{Department of Physics, Pohang University of Science and Technology,
	Pohang 37673, SK}
\affil[h]{Department of Industrial and Management Engineering, Pohang University of Science and Technology,
	Pohang 37673, SK}
\affil[*]{These authors contributed equally to this work: D.M., J.Y., S.K.}

\date{\today}


\baselineskip24pt

\maketitle 

%
%
\newpage

\begin{abstract}
Human migration and mobility drives major societal phenomena including epidemics, economies, innovation, and the diffusion of ideas.
Although human mobility and migration have been heavily constrained by geographic distance throughout the history, advances and globalization are making other factors such as language and culture increasingly more important.
Advances in neural embedding models, originally designed for natural language, provide an opportunity to tame this complexity and open new avenues for the study of migration.
Here, we demonstrate the ability of the model \textit{word2vec} to encode nuanced relationships between discrete locations from migration trajectories, producing an accurate, dense, continuous, and meaningful vector-space representation.
The resulting representation provides a functional distance between locations, as well as a ``digital double'' that can be distributed, re-used, and itself interrogated to understand the many dimensions of migration.
We show that the unique power of \textit{word2vec} to encode migration patterns stems from its mathematical equivalence with the gravity model of mobility.
Focusing on the case of scientific migration, we apply \textit{word2vec} to a database of three million migration trajectories of scientists derived from the affiliations listed on their publication records. 
Using techniques that leverage its semantic structure, we demonstrate that embeddings can learn the rich structure that underpins scientific migration, such as cultural, linguistic, and prestige relationships at multiple levels of granularity.
Our results provide a theoretical foundation and methodological framework for using neural embeddings to represent and understand migration both within and beyond science.
\end{abstract}

How far apart are two places?The question is surprisingly hard to answer when it involves human migration and mobility.
Although geographic distance has historically constrained human movements, it is becoming less relevant in a world increasingly interconnected by rapid communications and travel.
For instance, a person living in Australia is more likely to migrate to the United Kingdom, a far-away country with similar language and culture, than to a much closer country such as Indonesia~\cite{pew2018migration}.
Similarly, a student in South Korea is more likely to attend a university in Canada than one in neighboring North Korea~\cite{unesco2019students}.
Although geographic distance has been used as the most prominent basis for models of migration and mobility, such as the gravity~\cite{zipf1946gravity} and radiation~\cite{simini2012universal} models, the diminishing relevance of geography calls for alternative ways of conceptualizing ``distance''~\cite{boschma2005proximity, brown1970migration, kim2018functional}.

Yet, functional distances are often low-resolution, computed at the level of countries rather than regions, cities, or organizations, and have focused on only a single facet of migration at a time.
By contrast, real-world migration is multi-faceted, influenced simultaneously by geography, language, culture, history, and economic opportunity.
Low dimensional distance alone cannot represent the multitude of inter-related factors that drive migration.
Although networks have been explored as a solution to representing many dimensions of migration, edges only encode simple, dyadic relationships between connected entities.
Capturing the complexity of migration requires moving beyond simple functional distances and networks, to learning high-dimensional landscapes of migration that incorporate many facets of migration into a single fine-grained and continuous representation.
Such a representation can be used not only to measure distances at multiple scales, but also to act as a convenient ``digital double'', an entire functional topology that can be distributed, incorporated into future analyses, and itself interrogated to reveal fundamental insights into patterns of global migration. 

Here, we demonstrate that the \textit{word2vec} model (Skip-Gram Negative Sampling)~\cite{mikolov2013word2vec} is equivalent to the gravity law of mobility, a fundamental  framework used to model migration across many domains.
We then empirically test the resulting representation by its ability to derive the functional distances between locations from migration trajectories.
After validating its accurate representation of real-world data, we apply a variety of techniques that leverage the unique and powerful semantic structure of the embedding space to study scientific migration.
Doing so demonstrates \textit{word2vec}'s capacity to encode rich information related to geography, culture, language, and even prestige, at multiple scales of analysis.

While the \textit{word2vec} model shown here can be applied across domains of migration, here we demonstrate its applicability by applying it to study scientific migration.
Scientific migration is a central driver of the globalized scientific enterprise~\cite{czaika2018globalisation, box2008competition} and it is strongly related to innovation~\cite{braunerhjelm2020labor, kaiser2018innovation}, impact~\cite{sugimoto2017mostimpact, petersen2018multiscale}, collaboration~\cite{rodrigues2016mobility}, and the diffusion of knowledge~\cite{braunerhjelm2020labor, morgan2018prestige}.
Researchers migrate between organizations as they attain new roles throughout their careers, often motivated by the desire to expand their professional networks~\cite{bauder_international_2020}, to gain access to prestigious institutions~\cite{boring_international_2015}, to gain entry into high-performing research groups~\cite{azoulay_mobility_2017}, or to obtain resources for research~\cite{hunter_elite_2009}.
Their choice of destination is however constrained by many factors, including rigid prestige hierarchies that shape faculty hiring~\cite{clauset2015hierarchy, deville2014career}, language~\cite{brandi_reasons_2011}, visa \& immigration policies~\cite{kerr_america_2018}, and family considerations~\cite{azoulay_mobility_2017,ackers_internationalisation_2008}.
In spite of its importance, holistic understandings of global scientific migration have been limited by the sheer scope and complexity of the phenomenon~\cite{robinson2019mobility, deville2014career}, being further confounded by the diminishing role of geography in shaping the landscape of scientific migration. 
Its known structural properties combined with the difficulty of its study at scale make scientific migration the ideal case study for application of \textit{word2vec}. 

Trajectories of scientific migration are constructed using more than three million name-disambiguated authors who were \textit{mobile}---having more than one affiliation---between 2008 and 2019, as evidenced by their publications indexed in the Web of Science database (see Methods).
As a scientist's career progresses, they move between organizations or pick up additional (simultaneous) affiliations forming \textit{affiliation trajectories} (Fig.~\ref{fig:gravity_performance}a).
Thus, the trajectories encode both migration and co-affiliation---the holding of multiple simultaneous affiliations involving the sharing of time and capital between locations---that is typical of scientific migration~\cite{rodrigues2016mobility, sugimoto2017mostimpact} (see \SI).
This particular intricacy of scientific migration further illustrates how \textit{word2vec} can be applied to even the most complex domains.
We also apply this technique to U.S. passenger flight itinerary records and Korean accommodation reservations (Detailed descriptions are available in the Methods) in order to demonstrate its applicability to incredibly distinct domains of migration and mobility.

Here, we study the skip-gram negative sampling (SGNS), or \textit{word2vec} neural-network architecture (see Methods).
This neural embedding model, originally designed as a  language model~\cite{mikolov2013word2vec}, made breakthroughs by revealing novel insights into texts~\cite{tshitoyan2019mat2vec, garg2018gender, kozlowski2018geometry, hamilton2016diachronic, le2014doc2vec, nakandala2017gendered}, networks~\cite{grover2016node2vec, perozzi2014deepwalk, linzhuo2020hyperbolic} and trajectories~\cite{liu2016exploring, feng2017poi2vec, yao2018representing, cao2019habit2vec, crivellari2019motion, solomon2018predict}.
It works under the notion that a good representation should facilitate prediction, learning a mapping between words can predict a target word based on its context (surrounding words).
The model is also computationally efficient, robust to noise, and can encode relations between entities as geometric relationships in the vector space~\cite{levy2014neural, nakandala2017gendered, kozlowski2018geometry, an2018semaxis, hashimoto2016word}.
When applied to the trajectory data, each location is encoded into a vector space, and vectors relate to one another based on the likelihood of locations appearing adjacent to one another in the same trajectory. 
Also, \textit{word2vec} can be interpreted as a kind of metric recovery, which recovers the underlying metric of the semantic manifold ~\cite{hashimoto2016word}.
Although more sophisticated embedding techniques~\cite{devlin2018bert, pennington2014glove}, some adapted towards migration data~\cite{pennington2014glove, zhai2019beyond,gao2018place2vec, bo2017place2vec, simini2021deep}, have been developed, the standard \textit{word2vec} remains a powerful model for representing migration data, owing to its simplicity, intuitiveness, and accessibility.
Establishing a theoretical and methodological foundation for \textit{word2vec} is essential for better understanding and application of other more sophisticated models.

The gravity model framework~\cite{zipf1946gravity} is a widely-used, fundamental migration model~\cite{curiel2018citygravity, jung2008highwaygravity, hong2016busgravity, truscott2012epidemicgravity} that connects the \textit{expected flux}, $\hat{T}_{ij}$, between locations based on their populations and distance:

\begin{equation}
	\label{eq:gravity_basic}
	\hat{T}_{ij} = Cm_{i}m_{j}f(r_{ij}),
\end{equation}
where $m_{i}$ is the population of location $i$,  $f(r_{ij})$ is a decay function with respect to distance between locations, and $C$ is a constant estimated from data (see Methods).
Here, we use the mean annual number of unique mobile and non-mobile authors who were affiliated with each organization.
$\hat{T}_{ij}$ or  ``expected flux''~\cite{simini2012universal},  as the expected frequency of the co-occurrence of location $i$ and $j$ in the trajectory in the gravity model.

The gravity model posits that the expected flow, $\hat{T}_{ij}$, ($\hat{T}_{ij }= \hat{T}_{ji}$), is proportional to the locations' population, $\hat{T}_{ij} \propto m_{i} m_{j}$, and decays as a function of their distance, $f(r_{ij})$.
Traditionally, the decay function has been defined in terms of geographic distance, due to its intuitiveness and availability.
Here, we also consider the \textit{embedding distance}, calculated as the cosine distance between location vectors modeled by \textit{word2vec}, to test the ability to encode migration data.
The decay function $f(r_{ij})$ defines the effect of distance, and different decay functions can model fundamentally different mechanisms \cite{barthelemy2011spatial} such as the cost functions for a given distance and the spatial granularity of the observation.
For geographic distance, we define $f(r_{ij})$ as the standard power-law function, and for the embedding distance, we use the exponential function, selected as the best performing for each case (See \SI{}, Fig.~\ref{fig:supp:predict_geo_bymodel} and Fig.~\ref{fig:supp:predict_emb_bymodel} for more information).

\section*{Results}
%
\subsection*{word2vec and the gravity model}

We first demonstrate the mathematical equivalence between the SGNS model and the gravity model.
The {\it word2vec} model takes a location trajectory, denoted by ($a_{1}, a_{2}, \ldots, a_{T}$), as input.
A target location $a_t=i$ is considered to have a context location $a_{t'}=j$ that appears in the previous or subsequent $w$ locations in the trajectory, \ie  $j \in [a_{t-w}, \ldots, a_{t-1}, a_{t+1},\ldots, a_{t+w}]$.
	{\it word2vec} learns an embedding by estimating the probability that location $i$ has context $j$:
\begin{align}
	P\left(j \given i \right):= \frac{\exp(\bm{u}_j \cdot \bm{v}_{i})}{Z_i}, \label{eq:prob_w2v_ng}
\end{align}
where the denominator $Z_i=\sum_{j' \in \mathcal{A}} \exp(\bm{u}_{j'} \cdot \bm{v}_{i})$ is a normalization constant, and $\mathcal{A}$ is the set of all locations.
Although {\it word2vec} generates two embedding vectors $v_{i}$ and $u_i$---referred to as the in-vector and out-vector, respectively---we follow convention to use the in-vector $\bm{v}_i$ as an embedding of location $i$.
Training {\it word2vec} is computationally expensive because of $Z_i$ that extends over all $|{\cal A}|$ locations.

Negative sampling is a widely used heuristics to efficiently train {\it word2vec} without explicitly calculating $Z_i$.
Negative sampling was introduced as a \textit{simplified} version of Noise Contrastive Estimation (NCE)~\cite{mikolov2013word2vec,Gutmann2010}. 
We show that this simplification gives rise to a biased est, which subsequently lead to the equivalence between SGNS word2vec and the gravity model. 

NCE is a generic estimator for probability model~\cite{Gutmann2010}
\begin{align}
	\label{eq:nce-model}
	P_m(x) = \frac{ f(x) }{\sum_{ x' \in {\cal X}} f(x') },
\end{align}
where $f$ is a positive real-valued likelihood function of data $x$, and ${\cal X}$ is the set of all data.
Note that {\it word2vec} belongs to this class of probability models, with $x=\vect{u}_i\cdot \vect{v}_j$ and $f(x)=\exp(x)$.
To train {\it word2vec} with NCE~\cite{mikolov2013word2vec}, one samples a center-context pair $(i,j)$ from the given data and labels the pair as $Y=1$. 
Then, one replaces the context location $j$ with a random location $j'$ sampled from a noise distribution $p_0(j')$ and labels the pair as $Y=0$. 
NCE finds the embedding that can classify the center-context pairs using a logistic function  (see \SI)

\begin{align}
	\label{eq:nce}
	P^{\text{NCE}}\left(Y_{j}=1 \vert j\right) = \frac{
		1
	}{
		1 + \exp\left[ - \ln f(\vect{u}_j \cdot \vect{v}_{i})  + \ln p_0(j) \right]
	},
\end{align}
by maximizing the log-likelihood 
\begin{align}
	\label{eq:log-likelihood-logistic-regress}
	{\cal J}^{\text{NCE}} = \sum_{i \in {\cal A}}\sum_{j \in {\cal D}} \left[ Y_{j} \log P^{\text{NCE}}(Y_{j} = 1\vert j)  \right. \nonumber \\ + \left. (1-Y_{j}) \log P^{\text{NCE}}(Y_{j} = 0 \vert j)\right].
\end{align}

Note that NCE is an unbiased estimator that has asymptomatic convergence to the optimal embedding in terms of the original word2vec's objective function, ${\cal J}$~\cite{Chia2010,Dyer2014}. Let us revisit negative sampling from the perspective of NCE. Negative sampling simplifies NCE by dropping $ \ln p_0(j)$ in the logistic function, i.e.,
\begin{align}
	\label{eq:logistic-regress1}
	P^{\text{NS}}(Y_{j} = 1; \vect{v}_{i}, \vect{u}_j) = \frac{1}{1 + \exp(-\vect{u}_j \cdot \vect{v}_{i})}.
\end{align}
Despite its innocuous appearance, this simplification produces substantial biases. 
To see this, we rewrite $P^{\text{NS}}$ in the form of $P^{\text{NCE}}$ as 
\begin{align}
	&P^{\text{NS}}\left(Y_{j}=1 \vert j\right) \nonumber\\& = \frac{
		1
	}{
		1 + \exp\left[ - \left( \vect{u}_j \cdot \vect{v}_{i} + \ln p_0(j) + c \right) + \ln p_0(j) + c \right]
	}                                             \\
	                                   & = \frac{
		1
	}{
		1 + \exp\left[ - \ln f(\vect{u}_j \cdot \vect{v}_{i}) + \ln p_0(j) + c \right]
	},
\end{align}
where we define the likelihood function $f$ by
\begin{align}
	f(\vect{u}_j \cdot \vect{v}_{i}) = \exp\left( \vect{u}_j \cdot \vect{v}_{i} + \ln p_0(j) + c\right),
\end{align}
which is the unbiased estimator for the probability model
\begin{align}
	\label{eq:unbiasef}
	P_m^{\text{NS}}(\vect{u}_j \cdot \vect{v}_{i}) & = \frac{ f(\vect{u}_j \cdot \vect{v}_{i}) }{\sum_{ j' \in {\cal A}} f(\vect{u}_{j'} \cdot \vect{v}_{i}) },                                                                                               \\
	                                        & = \frac{  p_0(j)\exp( \vect{u}_j \cdot \vect{v}_{i} ) }{\sum_{j' \in \mathcal{A}}  p_0(j') \exp(\vect{u}_{j'} \cdot \vect{v}_{i})},                                                                      \\
	                                        & = \frac{  P^\gamma(j)\exp( \vect{u}_j \cdot \vect{v}_{i} ) }{\sum_{j' \in \mathcal{A}}  P^\gamma(j') \exp(\vect{u}_{j'} \cdot \vect{v}_{i})} \hspace{0.3cm}\\&(\because p_0(\ell) \propto P^\gamma (\ell)), \nonumber
\end{align}
where $P^\gamma (\ell)$ is the $\gamma$ power of the frequency of the location $\ell$.

Taken together, the conditional probability that SGNS \textit{word2vec} actually optimizes is
\begin{align}
\label{eq:ns_gamma}
	P^{\text{NS}}(j\given i) = P_m^{\text{NS}}(\vect{u}_j \cdot \vect{v}_{i})  =  \frac{  P^\gamma(j)\exp( \vect{u}_j \cdot \vect{v}_{i} ) }{Z'_i},
\end{align}
where $Z'_i=\sum_{j' \in \mathcal{A}}  P^\gamma(j') \exp(\vect{u}_{j'} \cdot \vect{v}_{i})$.
\eqref{eq:ns_gamma} clarifies the bias due to negative sampling, i.e., 
the noise distribution $p_0(j)=P^\gamma(j)$ appears in the numerator and, thus, is a part of the {\it word2vec} model.

Armed with this result, we can now show the equivalence between the SGNS {\it word2vec} model and the gravity model.
We set the window length to $w=1$ to restrict {\it word2vec} to predict the first-order flux $\hat T_{ij}$ between locations, as is the case for the gravity model.
Parameter $\gamma=1$ is a special choice that ensures that, when the embedding dimension is sufficiently large, there exists optimal in-vectors and out-vectors such that $\vect{v}_i = \vect{u}_i$ \cite{levy2014neural}.
By setting $\gamma=1$,  we have 
\begin{align}
	\label{eq:flow_w2v}
	\widehat{T}_{ij} \propto P(i) P^{\text{NS}}(j\given i) \propto \frac{P(i)P(j)\exp(\bm{u}_j \cdot \bm{u}_{i})}{Z_i}.
\end{align}
The flow $\widehat{T}_{ij}$ is symmetric (\ie $\widehat{T}_{ij}=\widehat{T}_{ji}$) because the skip-gram model neglects whether the context $j$ appears before or after the target $i$ in the trajectory which produces,
\begin{align}
	\label{eq:unbiasef2}
	T_{ij}=T_{ji} &\iff \frac{P(j) f(\vect{u}_j \cdot \vect{u}_{i}) }{Z_i} P(i) = \frac{P(i)  f(\vect{u}_i \cdot \vect{u}_{j}) }{Z_j} P(j) \nonumber \\
	&\iff \frac{1}{Z_i}= \frac{1}{Z_j}  \nonumber \\
	&\iff Z_i = Z_j
\end{align}
Taken together, the {\it word2vec} model with the negative sampling predicts a flow in the same form as the gravity model:
\begin{align}
	\hat{T}_{ij}=  C P(i) P(j)  \exp(\bm{v}_j \cdot \bm{v}_{i}). \label{eq:flow_w2v_ng0}
\end{align}

In other words, with large-enough dimensions and embedding optimally converges, \emph{word2vec with skip-gram negative sampling is mathematically equivalent to the gravity model}, with the mass given by the location's frequency $P(i)$, and the distance measured by their dot similarities.
While the gravity model describes migration flows from the given mass and locations, {\it word2vec}
estimates the positions in the vector space that best explain the given migration flow.

We further demonstrate \textit{word2vec}'s capacity to effectively represent gravity-like relationships with an synthetic benchmark. Namely, we train a \textit{word2vec} model using synthetic migration trajectories that strictly adhere to the gravity model (see Supporting Information for details). 
We find that distances in the embedding space strongly correlate with distances in the synthetic space which was explicitly structured according to the gravity model (Pearson correlation of 0.943 and 0.801, depending on the distance metric used, Fig. S4).
Our results align with a previous study about \textit{word2vec}' metric discovery capacity~\cite{hashimoto2016word}.

%
%
%
\subsection*{Embeddings provide functional distance between locations}

To ensure that \textit{word2vec} learns an accurate representation of migration that encodes meaningful functional distances, we devise an empirical validation task.
Here, we expect that an accurate representation of the migration data should provide a functional distance that better models the flux between institutions than does geographic distance and other representation methods.
We test this notion using three datasets representing different domains of human migration and mobility, showing that \textit{word2vec} consistently offers a better representation of actual migration flows than geographic distance, or alternative network and direct optimization approaches.

\begin{figure}[t]
	\centering
	\includegraphics[width=\linewidth]{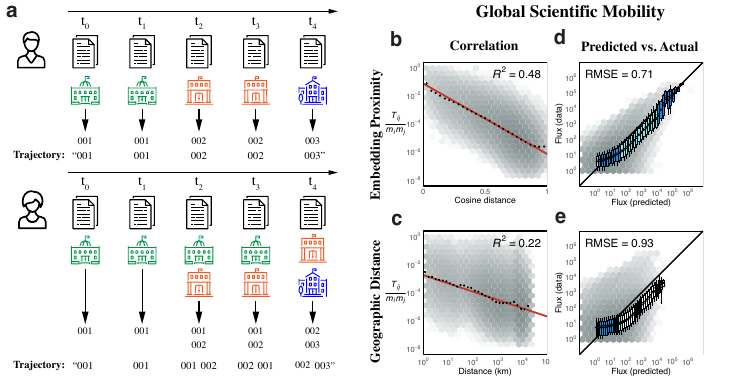}
	\caption{
		\textbf{Neural embedding provides functional distance that improves predictive power of the gravity model of migration best across three distinct human trajectory datasets. }
		\textbf{a.} A unique identifier is assigned to each organization and they are assembled into an affiliation trajectory ordered by year of publication (top). If an author lists multiple organization affiliations within the same year,  we shuffle the order within that year in each training iteration (bottom, see \SI). \textbf{b.} Embedding distance better explains the expected flux of global scientific migration than does geographic distance (\textbf{c}).
		The red line is the line of the best fit.
		Black dots are mean flux across binned distances.
		99\% confidence intervals are plotted for the mean flux in each bin.
		Correlation is calculated on the data in the log-log scale ($p < 0.0001$ across all fits).
		The lightness of each hex bin indicates the frequency of organization pairs within it.
		\textbf{d.} Predictions of flux between institutions made using embedding distance outperform those made using geographic distance (\textbf{e}).
		Box-plots show the distribution of actual flux for binned values of predicted flux.
		Box color corresponds to the degree to which the distribution overlaps with $y = x$.
		``RMSE'' is the root-mean-squared error between the actual and predicted values.
	Embedding distance consistently produces powerful functional distance for U.S. flight itineraries and Korean accommodation reservations (see \SI{}).}
	\label{fig:gravity_performance}
\end{figure}

In the case of scientific migration, the embedding distance explains more than twice the expected flux ($R^{2} = 0.48$, Fig.~\ref{fig:gravity_performance}b) than does the geographic distance ($R^{2} = 0.22$, Fig.~\ref{fig:gravity_performance}c), and predictions made using the embedding distance outperform those using the geographic distance (Fig.~\ref{fig:gravity_performance}d-e). These patterns hold for the subsets of only domestic (within-country organization pairs, Fig.~\ref{fig:supp:predict_geo_bymodel} and Fig.~\ref{fig:supp:gravity_correlation}c) and only international migration flows (across-country organization pairs, Fig.~\ref{fig:supp:gravity_correlation}d). 
We also find that the embedding distance outperforms a generalized version of the gravity model, which incorporates information on shared geography and language alongside geographic distance (see \SI).

Similarly, the embedding distance explains more than twice the expected flux between airports ($R^{2} = 0.51$, Fig.~\ref{fig:supp:gravity_performance_si}a) than does geographic distance ($R^{2} = 0.22$, Fig.~\ref{fig:supp:gravity_performance_si}b), which has traditionally been used to quantify distance for the gravity model. Also, the embedding distance produces better predictions of actual flux between airports than does the geographic distance Fig.~\ref{fig:supp:gravity_performance_si}c-d).
In the case of Korean accommodation reservations, embedding distance better explains the expected flux ($R^{2} = 0.57$, Fig.~\ref{fig:supp:gravity_performance_si}e) than does geographic distance ($R^{2} = 0.25$, Fig.~\ref{fig:supp:gravity_performance_si}f), and predictions made using the embedding distance outperform those made with geographic distance (Fig.~\ref{fig:supp:gravity_performance_si}g-h).

%
%
%
%

The embedding distance also out-performs alternative diffusion-based network distance measures including the personalized Page Rank scores calculated from the underlying migration network (Fig.~\ref{fig:supp:distancemetrics}, Fig.~\ref{fig:supp:gravity_pprcos}, Fig.~\ref{fig:supp:gravity_pprjsd}).
The embedding distance derived from neural embedding also explains more of the flux and better predicts migration flows than simpler embedding baselines, such as distances derived from a singular-value decomposition and a Laplacian Eigenmap embedding~\cite{belkin2003laplacian} of the underlying location co-occurrence matrix, Levy's symmetric \textit{word2vec}\cite{levy2014neural}, and even direct optimization of the gravity model (Fig.~\ref{fig:supp:distancemetrics} and Tables~\ref{supp:table:r2_table},~\ref{supp:table:rmse_exp_table},~\ref{supp:table:rmse_power_table}).

In sum, our results demonstrate that, consistently and efficiently, the embedding distance better captures patterns of actual migration than does the geographic distance.
The embedding distance also outperforms alternatives in terms of the common part of commuters measure~\cite{sorensen1948method} (Fig.~\ref{fig:supp:cpc-performance})

In practice, because of noise, limited amounts of data, and imperfect optimization, the equivalence may only approximately hold.
Indeed, we find that the in- and out-vectors tend to be different and that the cosine similarity tends to better capture real-world migration than the inner product similarity.
This result echos other applications of word embedding, such as word analogy testing~\cite{levy2015improving}, in which cosine distance also outperformed the inner product similarity.
Nevertheless, a model with the inner product similarity has the second-best performance after cosine similarity (Tables~\ref{supp:table:r2_table},~\ref{supp:table:rmse_exp_table},~\ref{supp:table:rmse_power_table}), and the embedding distance still outperforms all alternatives we considered.

\subsection*{Embeddings capture the global structure of migration}

\begin{figure*}
    \centering
    \includegraphics[width=\textwidth]{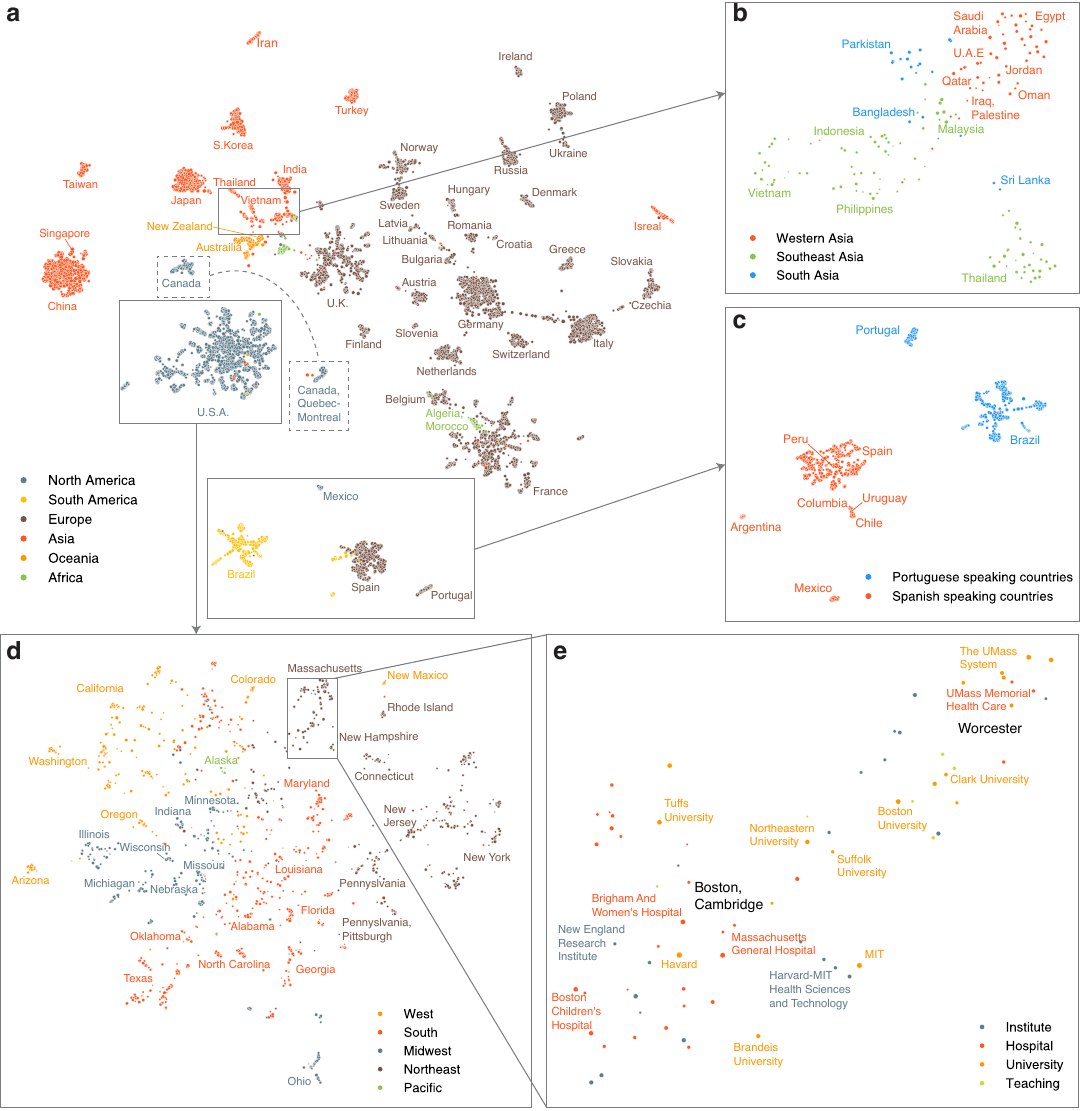}
    \caption{
		\textbf{Projection of embedding space reveals complex multi-scale structure of organizations.}
		\textbf{a.}
		UMAP projection \cite{mcinnes2018umap} of the embedding space reveals country-level clustering.
		Each point corresponds to an organization and its size indicates the average annual number of mobile and non-mobile authors affiliated with that organization from 2008 to 2019.
		Color indicates the region.
		The separation of organizations in Quebec and the rest of Canada is highlighted.
		\textbf{b.} Zooming into (re-projecting) the area containing countries in Western, South, and Southeast Asia shows a geographic and cultural gradient of country clusters.
		\textbf{c.} Similarly, zooming into the area containing organizations in Spain, Portugal, South, and Central America shows clustering by most widely-spoken majority language group: Spanish and Portuguese.
		\textbf{d.} Doing the same for organizations in the United States reveals geographic clustering based on state, roughly grouped by Census Bureau-designated regions,
		\textbf{e.} Zooming in further on Massachusetts reveals clusters based on urban center (Boston, Worcester), organizational sector (hospitals vs. university), and university systems and prestige (UMass system vs. Harvard, MIT).
	}
	\label{fig:projection}
\end{figure*}

In the remainder of the paper, we focus on scientific migration as a case study to interrogate the geometric space generated by the neural embedding.
In the process, we also study the multi-faceted relationships between scientific organizations.
To explore the topological structure of the embedding, we use a topology-based dimensionality reduction method (UMAP~\cite{mcinnes2018umap}) to obtain a two-dimensional representation of the embedding space (Fig.~\ref{fig:projection}a).
By leveraging the unique characteristics of representation learning approach, we are able to show the relationships between individual organizations, rather than aggregates such as nations or cities, producing the largest and highest resolution ``map'' of scientific migration to date.

Globally, the geographic constraints are conspicuous; organizations tend to form clusters based on their national affiliations and national clusters tend to be near their geographic neighbors.
At the same time, the embedding space also reflects a mix of geographic, historic, cultural, and linguistic relationships between regions much more clearly than alternative network representations (Fig.~\ref{fig:supp:network_vis}) that have been common in studies of scientific migration~\cite{chinchilla2018global, czaika2018globalisation}.

The embedding space also allows us to \textit{zoom in} on subsets and re-project them to reveal local relationships.
For example, re-projecting organizations located in Western, Southern, and Southeastern Asia with UMAP (Fig.~\ref{fig:projection}b) reveals a gradient of countries between Egypt and the Philippines that largely corresponds to geography, but with some exceptions seemingly stemming from cultural and religious similarity.
For example, Malaysia, with its official religion of Islam, is nearer to Middle Eastern countries in the embedding space than to many geographically-closer South Asian countries.
We validate this finding quantitatively with the cosine distance between nations (the centroids of organizations vectors belonging to a given country).
Malaysia is nearer to many Islamic countries such as Iraq ($d = 0.27$), Pakistan ($d = 0.32$), and Saudi Arabia ($d = 0.41$)  than neighboring but Buddhist Thailand ($d = 0.43$) and neighboring Singapore ($d = 0.48$).

Linguistic and historical ties also affect scientific migration.
We observe that Spanish-speaking Latin American nations are positioned near Spain (Fig.~\ref{fig:projection}c), rather than Portuguese-speaking Brazil ($d = 0.35$ vs. $d = 0.54$ for Mexico and $d = 0.39$ vs. $d = 0.49$ for Chile) reflecting linguistic and cultural ties.
Similarly, North-African countries that were once under French rule such as Morocco are closer to France ($d =0.32$) than to similarly geographically-distant European countries such as Spain ($d = 0.39$), Portugal ($d = 0.52$), and Italy ($d = 0.52$).
Comparable patterns exist even within a single country.
For example, organizations within Quebec in Canada are located nearer France ($d = 0.37$) than the United States ($d = 0.51$).

Mirroring the global pattern, organizations in the United States are largely arranged according to geography (Fig.~\ref{fig:projection}d).
Re-projecting organizations located in Massachusetts (Fig.~\ref{fig:projection}e) reveals structure based on urban centers (Boston vs. Worcester), organization type (e.g., hospitals vs. universities), and university systems (University of  Massachusetts system vs. Harvard \& MIT).
For example, even though UMass Boston is located in Boston, it clusters with other universities in the UMass System ($d = 0.29$) rather than the other typically more highly-ranked and research-focused organizations in Boston ($d = 0.39$), implying a relative lack of migration between the two systems.
Similar structures can be observed in other states such as among New York's CUNY and SUNY systems (Fig.~\ref{fig:supp:proj_newyork}), Pennsylvania's state system (Fig.~\ref{fig:supp:proj_pennsylvania}), Texas's Agricultural and Mechanical universities (Fig.~\ref{fig:supp:proj_texas}), and between the University of California and State University of California systems (Fig.~\ref{fig:supp:proj_california}).

%
%

\begin{figure}[tbhp]
	\centering
	\includegraphics[width=0.8\linewidth]{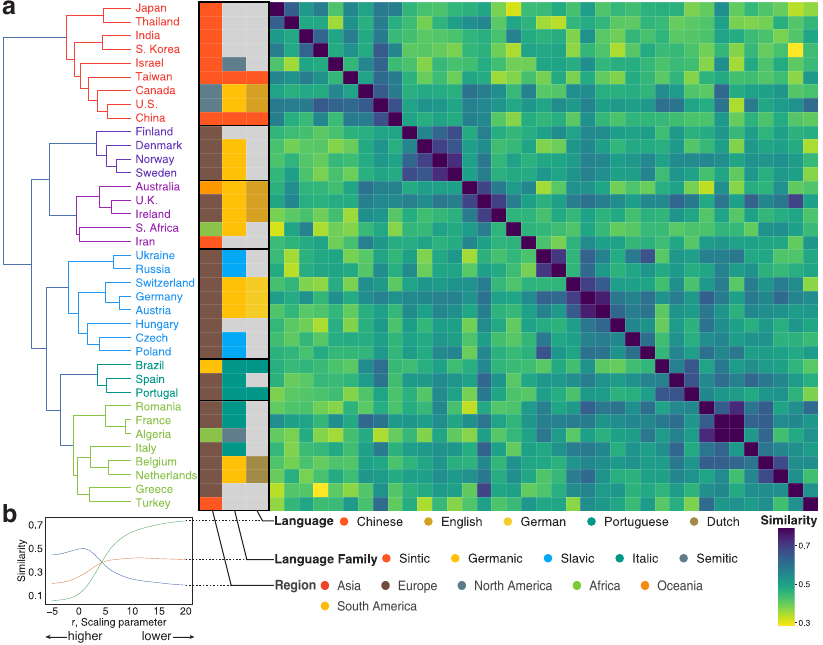}
	\caption{
		\textbf{Geography, then language, conditions international migration.}
		\textbf{a.} Hierarchically clustered similarity matrix of country vectors aggregated as the mean of all organization vectors within countries with at least 25 organizations.
		Color of matrix cells corresponds to the cosine similarity between country vectors.
		Color of country names corresponds to their cluster.
		Color of three cell columns separated from the matrix corresponds to, from left to right, the region of the country, the language family~\cite{ethnologue}, and the dominant language.
		\textbf{b.}	Element-centric cluster similarity~\cite{gates2019element} reveals the factors dictating hierarchical clustering (See methods).
		Region better explains the grouping of country vectors at higher levels of the clustering.
		Language family, and then the most widely-spoken language, better explain the fine-grained grouping of countries.
	}
	\label{fig:clustering}
\end{figure}

Just as the embedding space makes it possible to \textit{zoom in} on subsets of organizations, it is also possible to \textit{zoom out} by aggregating organizational vectors.
In doing so, we can examine the large-scale structure that governs scientific migration.
We define the representative vector of each country as the average of their organizational vectors and, using their cosine similarities, perform hierarchical clustering of nations that have at least 25 organizations represented in the embedding space (see Fig. \ref{fig:clustering}a).
The six identified clusters roughly correspond to countries in Asia and North America (orange), Northern Europe (dark blue), the British Commonwealth and Iran (purple), Central and Eastern Europe (light blue), South America and Iberia (dark green), and Western Europe and the Mediterranean (light green).
The cluster structure shows that not only geography but also linguistic and cultural ties between countries are related to scientific migration.

We quantify the relative importance of geography (by region), and language (by the most widely-spoken language of each country) using the element-centric clustering similarity~\cite{gates2019element}, a method that can compare hierarchical clustering and disjoint clustering such as geography or language at the different levels of hierarchy by explicitly adjusting a scaling parameter $r$, acting like a \textit{zooming lens} (See methods).
If $r$ is high, the similarity is based on the lower levels of the dendrogram, whereas when $r$ is low, the similarity is based on higher levels.
Fig. \ref{fig:clustering}b demonstrates that regional relationships play a major role at higher levels of the clustering process (low $r$), and language (family) explains the clustering more at the lower levels (high $r$).
This suggests that the embedding space captures the hierarchical structure of migration.

%
%

%
%
\subsection*{Embeddings capture latent prestige hierarchy}

The embedding space can also encode more fine-grained relationships between entities.
For example, prestige hierarchies, in which researchers tend to move to similar or less prestigious organizations~\cite{deville2014career, clauset2015hierarchy}, are known to underpin the dynamics of scientific migration.
Could the embedding space, to which no explicit prestige information is given, encode a prestige hierarchy?
This question is tested by exploiting the geometric properties of the embedding space with SemAxis~\cite{an2018semaxis}.
Here, we use SemAxis to operationalize the abstract notion of academic prestige, defining an axis in the embedding space using known high- and low-ranked universities as poles.
We use the Times Ranking of World Universities as an external proxy for prestige (we also use research impact from the Leiden Ranking~\cite{waltman2012leidenrankings}, see \SI),
The high-rank pole is defined as the average vector of the top five U.S. universities according to the rankings, whereas the low-rank pole is defined using the five bottom-ranked (geographically-matched by U.S. census region) universities.
We derive an embedding-based ranking for universities based on the geometrical spectrum from the high-ranked to low-ranked poles (see Materials and Methods).

\begin{figure}[hb!]
	\centering
	\includegraphics[width=0.6\textwidth]{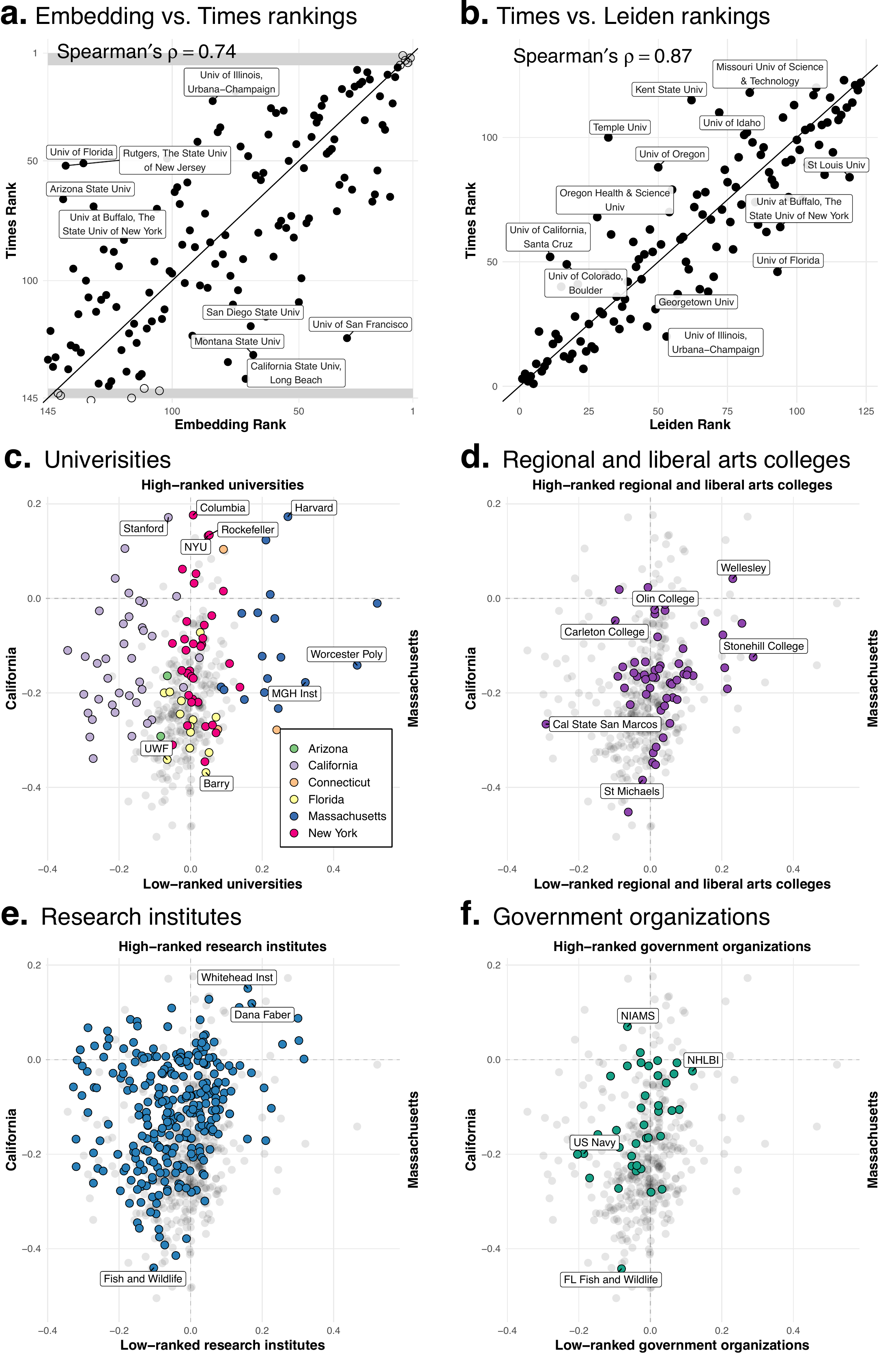}
	\caption{
		\textbf{Embedding captures latent geography and prestige hierarchy.}
		\textbf{a.} Comparison between the ranking of organizations in the Times ranking and the embedding ranking derived using SemAxis.
		Un-filled points are those top and bottom five universities used to span the axis.
		Even when considering only a total of ten organization vectors, the estimate of the Spearman's rank correlation between the embedding and Times ranking is $\rho = 0.73$ ($n = 145$, $p < 0.0001$), which increases when more top-and-bottom ranked universities are included (Fig.~\ref{fig:supp:semaxis_compare}).
		\textbf{b.} The Times ranking is correlated with Leiden Ranking of U.S. universities with Spearman's $\rho = 0.87$ and $p < 0.001$.
		\textbf{c-f.} Illustration of SemAxis projection along two axes;  the \textit{latent geographic axis}, from California to Massachusetts (left to right) and the \textit{prestige axis}.
		Shown for U.S. Universities (\textbf{c}), Regional and liberal arts colleges (\textbf{d}), Research institutes (\textbf{e}), and Government organizations (\textbf{f}).
		Full organization names are listed in Table~\ref{table:supp:orglabels}.
	}
	\label{fig:semaxis}
\end{figure}

The embedding space encodes the prestige hierarchy of U.S. universities that is coherent with real-world university rankings.
The embedding-based ranking is strongly correlated with the Times ranking (Spearman's $\rho = 0.73$, Fig.~\ref{fig:semaxis}a).
They are also strongly correlated with the mean normalized citation score of university's research output outlined in the Leiden rankings~\cite{waltman2012leidenrankings} (Spearman's $\rho = 0.77$, Fig.~\ref{fig:supp:semaxis_compare_leiden}b).
For reference, the correlation between the Times and the Leiden rankings is 0.87 (Spearman's $\rho$, Fig.~\ref{fig:semaxis}b).
The correlation between the embedding-based ranking and the Times ranking is robust regardless of the number of organizations used to define the axes (Fig.~\ref{fig:supp:semaxis_compare}), such that even using only the single top-ranked and bottom-ranked universities produces a ranking that is significantly correlated with the Times ranking (Spearman's $\rho = 0.46$, Fig.~\ref{fig:supp:semaxis_compare}a).
The correlation is also comparable to more direct measures such as node strength (sum of edge weights, Spearman's $\rho = 0.73$) and eigenvector centrality (Spearman's $\rho = 0.76$, see \SI) from the migration network.
The strongest outliers that were ranked more highly in the Times ranking than in the embedding-based ranking tend to be large state universities such as Arizona State University and the University of Florida.
The institution higher in the embedding-based ranking tend to be relatively-small universities near major urban areas such as the University of San Francisco and the University of Maryland Baltimore County, possibly reflecting exchanges of scholars with nearby highly-ranked institutions at these locations.
This analysis is not limited to the United States.
Among the ten countries with the most universities represented in the Leiden rankings, all except for China have a Spearman's $\rho \geq 0.5$ between their prestige axis and the relative rankings of their universities (see Table.~\ref{table:supp:semaxis_prestige_bycountry}).
In sum, our results suggest that the embedding space is capable of capturing information about academic prestige, even when the representation is learned using data without explicit information on the direction of migration (as in other formal models~\cite{clauset2015hierarchy}), or prestige.

The axes can be visualized to examine the relative position of organizations along the prestige axis, and along a geographic axis between California and Massachusetts.
Prestigious universities such as Columbia, Stanford, MIT, Harvard, and Rockefeller are positioned towards the top of the axis (Fig.~\ref{fig:semaxis}c).
Universities at the bottom of this axis tend to be regional universities with lower national profiles (yet still ranked by Times Higher Education) and with more emphasis on teaching, such as Barry University and California State University at Long Beach.
The Massachusetts-California axis, roughly corresponding to East-West, further demonstrates the ability of these embeddings to capture latent geography. 
Distance along this axis strongly correlates with the longitudes of U.S. organizations (Spearman's $\rho = 0.63$).

By projecting other types of organizations onto the prestige axis, SemAxis offers a new way of representing a continuous spectrum of organizational prestige for which rankings are often low-resolution, incomplete, or entirely absent, such as for regional and liberal arts universities (Fig.~\ref{fig:semaxis}d), research institutes (Fig.~\ref{fig:semaxis}e), and government organizations (Fig.~\ref{fig:semaxis}f).
Their estimated prestige is speculative, though we find that it significantly correlates with their citation impact (Fig.~\ref{fig:supp:sector_semaxis_impact}).
Correlation with the geographic axis is strongest for universities (Spearman's $\rho = 0.64$), followed by research institutes (Spearman's $\rho = 0.57$) regional and liberal arts colleges (Spearman's $\rho = 0.57$), and government organizations (Spearman's $\rho = 0.30$); 
the relatively low geographic correlation for government organizations may stem from them having only one set of coordinates even if offices are spread across the country.

SemAxis rankings can also be applied towards investigating how prestige drives patterns of individuals' migration (Fig.~\ref{fig:supp:semaxis-individual}).
In line with past findings, we observe that transitions tend to occur between universities of similar or lower prestige~\cite{clauset2015hierarchy}.
Additionally, we observe two clusters of internal migration at the top and bottom of the SemAxis hierarchy. 

%
%

\begin{figure}[h!]
	\centering
	\includegraphics[width=\linewidth]{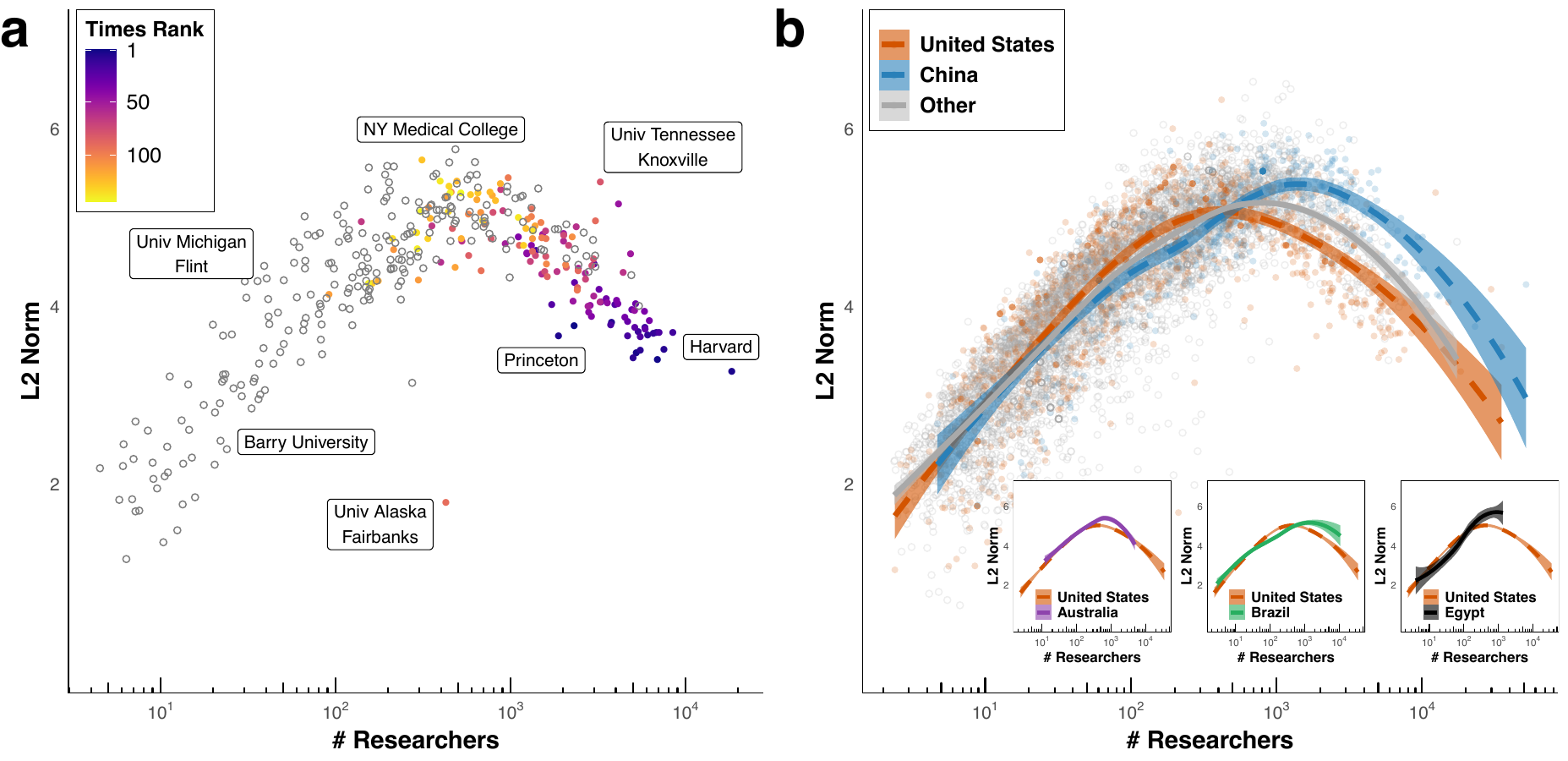}
	\caption{
		\textbf{Size of organization embedding vectors captures prestige and size of organizations.}
		\textbf{a.} Size (L2 norm) of organization embedding vectors compared to the number of researchers for U.S. universities.
		Color indicates the rank of the university from the Times ranking, with 1 being the highest ranked university.
		Uncolored points are universities not listed on the Times ranking.
		A concave-shape emerges, wherein larger universities tend to be more distant from the origin (large L2 norm); however, the more prestigious universities tend to have smaller L2 norms.
		\textbf{b.} We find a similar concave-curve pattern across many countries such as the United States, China, Australia, Brazil, and others (\textbf{inset}, and Fig.~\ref{fig:concave30}).
		Some countries exhibit variants of this pattern, such as Egypt, which is missing the right side of the curve.
		The loess regression lines are shown for each selected country, and for the aggregate of remaining countries, with ribbons mapping to the 99\% confidence intervals based on a normal distribution.
		Loess lines are also shown for organizations in Australia, Brazil, and Egypt (\textbf{inset}).
	}
	\label{fig:length}
\end{figure}

We also find that the size (L2 norm) of the organization embedding vectors provides insights into the characteristics of organizations (Fig.~\ref{fig:length}).
Up to a point (around 1,000 researchers), the size of U.S. organization's vectors tends to increase proportionally to the number of researchers (both mobile and non-mobile) with published work;
these organizations are primarily teaching-focused institutions, agencies, and hospitals that either are not ranked or have a low ranking.
However, at around 1,000 researchers, the size of the vector \emph{decreases} as the number of researchers increases.
These organizations are primarily research-intensive and prestigious universities with higher rank, research outputs, R\&D funding, and doctoral students (Fig.~\ref{fig:supp:length_vs_metaInfo}).
We report that this curve is almost universal across many countries.
For instance, China's curve closely mirrors that of the United States (Fig.~\ref{fig:length}b).
Smaller but scientifically advanced countries such as Australia and other populous countries such as Brazil also exhibit curves similar to the United States (Fig.~\ref{fig:length}b, inset).
Other nations exhibit different curves which lack the portions with decreasing norm, probably indicating the lack of internationally-prestigious institutions.
Similar patterns can be found across many of the 30 countries with the most total researchers (Fig.~\ref{fig:concave30}; see \SI{} for more discussion).

A similar pattern has been observed in applications of neural embedding to natural language, where it was proposed that a word vector's size represents its \emph{specificity}, i.e., the word associated with the vector frequently co-appears with particular context words~\cite{schakel2015measuring}.
If the word in question is universal, appearing frequently in many different contexts, it would not have a large norm due to a lack of strong association with a particular context.
Under this view, an organization with a small norm, such as Harvard, appears in many contexts alongside many different organizations in affiliation trajectories---it is well-connected.
We conduct simple empirical and model-based investigations to verify the underlying dynamics of this curve pattern.
However, in spite of theoretical support and an observed correlation between the vector size and the expected connectedness of the organization ($R^{2} = 0.17$), these experiments do not support ``specificity'' as the sole mechanism of the observed concavity (see \SI{}).
Another possibility is that the concave-curve is a result of distortion caused by representing hierarchy in an Euclidean space~\cite{chamberlain_neural_2017}, but this is also not supported by simulations. 
Instead, our findings emphasize that frequency and network connections constitute pivotal factors driving this pattern (Fig.~\ref{fig:supp:real_network} and ~\ref{fig:supp:model_network}, see \SI{} for more discussion).
Further work is necessary to determine the exact causes of this curve pattern in so many countries, whether the same pattern can be found in other domains of human migration, and if they suggest common structures between both migration and language.

%
%

\clearpage 

%
%
\section*{Conclusion}

Neural embedding approaches offer a novel, data-driven solution for efficiently learning an effective and robust representation of locations based on trajectory data, encoding the complex and multi-faceted nature of migration.
We discovered that the unique strength of \textit{word2vec} stems from its equivalence to a gravity model, making it a natural and theoretically-grounded tool for modeling migration.
By virtue of this equivalence, \textit{word2vec} learns accurate representations of migration across disparate domains as we demonstrated here.
Focusing on the case of scientific migration, we leverage the unique topological structure of the embedding space to reveal how it encodes nuanced aspects of migration, including global and regional geography, shared languages, and prestige hierarchies.

In revealing the correspondence between neural embeddings and the gravity model, the study of human migration can move beyond geographic and network-based models of migration, and instead leverage the high-order structure directly from individuals' migration trajectories using these robust and efficient methods.
This correspondence supplies a much-needed theoretical justification for the application of neural embedding techniques towards migration data, and contributes to a better understanding of neural embedding techniques. 
Moreover, our study offers a complementary approach to past applications of neural networks towards migration and mobility data~\cite{gao2018place2vec,bo2017place2vec}. 
Whereas most location-based embeddings highlight their predictive capability, we instead illustrate how an embedding \textit{word2vec} model creates accurate representations of migration data, a ``digital double'' that bundles many complex features of migration into a dense, continuous, and meaningful vector space representation.
Using this representation, functional distances can be derived at multiple scales, but it can also be interrogated to reveal fundamental insights about migration.
In addition to being intuitive, accessible, and theoretically grounded, the \textit{word2vec} approach outlined here also has the advantage of learning complex and implicit features of migration directly from raw trajectory data, rather than exploiting \textit{a priori} location features~\cite{simini2021deep}.

In conducting this analysis, we aim to offer a methodological framework for using \textit{word2vec} to study scientific migration, and migration more broadly, such as animal migration, immigration trends, transit-network mobility, discretized cell-phone location data, and international trade.
Once learned, functional distances between locations, such as countries, cities, or organizations, or the embedding model itself, can be published to facilitate re-use, and support reproducibility and transparency in cases when the underlying data is too sensitive to be made available.
Moreover, this approach can be used to learn a functional distance even between entities for which no geographic analog exists, such as between occupational categories based on individuals' career trajectories.
In addition to providing a functional distance that supports modeling and predicting migration patterns, we also demonstrate, through a variety of unique and power techniques, how the semantic topology of the embedding space can be leveraged to facilitate interpretation and application of the complex features of migration.
As we have shown, the embedding space allows the visualization of the complex structure of scientific migration at high resolution across multiple scales, providing a large and detailed map of the landscape of global scientific migration.
Other operations such as comparing entities or calculating aggregates, which could be complex and computationally expensive for other methods, are here reduced to simple vector arithmetic.
Embeddings also allow us to quantitatively explore abstract relationships between locations, such as academic prestige, and can potentially be generalized to other abstract axes.
Investigation of the structure of the embedding space, such as the vector norm, reveals universal patterns based on the organization's size and their vector norm that should be explored in future research.

This approach, and our study, also have several limitations.
First, the skip-gram \textit{word2vec} model assumes that migrations flows are symmetric, which is unlikely in real-world data. 
Breaking this assumption, however, also breaks the clear and simple mathematical equivalence between \textit{word2vec} and the gravity model of migration.
Future studies may consider directional embeddings to incorporate asymmetric nature of migration and mobility, such as the radiation model~\cite{simini2012universal}.
Second, the neural embedding approach is most useful in cases of migration between discrete units such as between countries, cities, and businesses; this approach is less useful in the case of mobility between locations represented using geographic coordinates, such as that sourced from cell phone tracking.
Third, neural embeddings are an inherently stochastic procedure, and so results may change across different iterations.
However, in this study we observe all results to be robust to stochasticity, likely emerging from the limited ``vocabulary'' of scientific mobility, airports, and accommodations (several thousand) and the relatively massive datasets used to learn representations (several million trajectories).
Applications of \textit{word2vec} to problem domains where the ratio of the vocabulary to data is smaller, however, should be implemented with caution to ensure that findings are not the result of random fluctuations.
Fourth, the case of scientific migration presents domain-specific limitations.
Reliance on bibliometric metadata means that we capture only long-term migration rather than the array of more frequent short-term mobility such as conference travel and temporary visits.
The kinds of migration we do capture---migration and co-affiliation---although conceptually different, are treated identically by our model.
Our data might further suffer from bias based on publication rates: researchers at prestigious organizations tend to have more publications, leading to these organizations appearing more frequently in affiliation trajectories.
Fifth, for simplicity, we ignore the role of specialities. In line with the concept of a ``persona''~\cite{yoon_persona2vec_2021}, it may be possible to create an interpretable embedding for each affiliation-subject pair, to understand the benefits associated with institutions that specialize in particular domains.
Finally, our data is limited to the period between 2008 and 2019, and so may not reflect current patterns of migration that were shaped by the COVID-19 pandemic. 

Migration and mobility are at the core of human nature and history, driving societal phenomena as diverse as epidemics~\cite{kraemer2020covid, truscott2012epidemicgravity} and innovation~\cite{kaiser2018innovation, sugimoto2017mostimpact, petersen2018multiscale, morgan2018prestige, rodrigues2016mobility}.
However, the paradigm of scientific migration may be changing.
Traditional hubs of migration have experienced many politically-motivated policy changes that affect scientific migration, such as travel restrictions in the U.S. and U.K.~\cite{chinchilla2018travelban}, whereas other countries, such as China, have risen as major attractors of international talent~\cite{cao2020returning}.
Unprecedented health crises such as the COVID-19 pandemic threaten to bring drastic global changes to migration by tightening borders and halting travel.
By revealing the correspondence between neural embedding and the gravity model and revealing their utility and efficacy, our study provides a theoretical foundation and methodological framework for a new approach that uses neural embeddings to study migration.

%
%
\section*{Methods}
\subsection*{Scientific migration data}
We source co-affiliation trajectories of authors from the Web of Science database hosted by the Center for Science and Technology Studies at Leiden University.
Trajectories are constructed from author affiliations listed on the byline of publications for an author.
Given the limitations of author-name disambiguation, we limit our analyses to papers published after 2008, when the Web of Science began providing full names and institutional affiliations~\cite{caron2014disambiguation} that improved disambiguation (see \SI).
This yields 33,934,672 author-affiliation combinations representing 12,963,792 authors.
Each author-affiliation combination is associated with the publication year and an ID linking it to one of 8,661 disambiguated organizational affiliations (see \SI{} for more detail).
Trajectories are represented as the list of author-affiliation combinations, ordered by year of publication, and randomly ordered for combinations within the same year. The most fine-grained geographic unit in this data is the organization, such as a university, research institute, business, or government agency.

Here, authors are classified as mobile when they have at least two distinct organization IDs in their trajectory, meaning that they have published using two or more distinct affiliations between 2008 and 2019.
Under this definition, mobile authors constitute 3,007,192 or 23.2\% of all authors and 17,700,095 author-affiliation combinations.
Mobile authors were associated with 2.5 distinct organizational affiliations on average.
Rates of migration differ across countries.
For example, France, Qatar, the USA, Iraq, and Luxembourg had the most mobile authors (Fig.~\ref{fig:supp:descriptives}c).
However, due to their size, the USA, accounted for nearly 40 \% of all mobile authors worldwide (Fig.~\ref{fig:supp:descriptives}a), with 10 countries accounting for 80 \% of all migration (Fig.~\ref{fig:supp:descriptives}b).
The countries with the highest proportion of mobile scientists are France, Qatar, the United States, and Iraq, whereas those with the lowest are Jamaica, Serbia, Bosnia \& Herzegovina, and North Macedonia (Fig.~\ref{fig:supp:descriptives}c).
In most cases, countries with a high degree of inter-organization migration also have a high degree of international migration, indicating that a high proportion of their total migration is international (Fig.~\ref{fig:supp:descriptives}d);
However, some countries such as France and the United States seem to have more domestic migration than international migration.
While the number of publications has increased year-to-year, the migration and disciplinary makeup of the dataset hvave not notably changed across the period of study (Fig.~\ref{fig:supp:pubs_over_time}).

\subsection*{U.S. flight itinerary data}
We source U.S. airport itinerary data from the Origin and Destination Survey (DB1B), provided by the Bureau of Transportation Statistics at the United States Department of Transportation.
DB1B is a sample of 10 percent of domestic airline tickets between 1993 and 2020, comprising 307,760,841 passenger itineraries between 828 U.S. airports.
A trajectory is constructed for each passenger flight itinerary, forming an ordered sequence of unique identifiers of the origin and destination airports.
Each itinerary is associated with a trajectory of airports including the origin, destination, and intermediary stops.
We use population $m_{i}$ as the total number of unique passengers who passed through each airport.

\subsection*{Korean accommodation reservation data}
We source Korean accommodation reservation data from collaboration with Goodchoice Company LTD.
The data contains customer-level reservation trajectories spanning the period of August 2018 through July 2020 and comprising 1,038 unique accommodation locations in Seoul, South Korea.
A trajectory is constructed for each customer, containing the ordered sequences of accommodations they reserved over time. We use the total number of unique customers who booked with each accommodation.

%
\subsection*{Embedding}

We embed trajectories by treating them analogously to sentences and locations analogously to words.
For U.S. airport itinerary data, trajectories are formed from the flight itineraries of individual passengers, in which airports correspond to unique identifiers.
In the case of Korean accommodation reservations, trajectories comprise a sequence of accommodations reserved over a customer's history.
For scientific migration, an ``affiliation trajectory'' is constructed for each mobile author, which is built by concatenating together their ordered list of unique organization identifiers, as demonstrated in Fig.~\ref{fig:gravity_performance}a (top).
In more complex cases, such as listing multiple affiliations on the same paper or publishing with different affiliations on multiple publications in the same year, the order is randomized within that year, as shown in Fig.~\ref{fig:gravity_performance}a (bottom).

These trajectories are used as input to the standard skip-gram negative sampling word embedding, commonly known as \textit{word2vec}~\cite{mikolov2013word2vec}.
\textit{word2vec} constructs dense and continuous vector representations of words and phrases, in which distance between words corresponds to a notion of semantic distance.
By embedding trajectories, we aim to learn a dense vector for every location, for which the distance between vectors relates to the tendency for two locations to occur in similar contexts.
Suppose a trajectory, denoted by ($a_{1}, a_{2}, \ldots, a_{T}$), where $a_{t}$ is the $t$th location in the trajectory. A location, $a_{t}$, is considered to have context locations, $a_{t-w}, \ldots, a_{t-1}, a_{t+1},\ldots, a_{t+w}$, that appear in the window surrounding $a_t$ up to a time lag of $w$, where $w$ is the window size parameter truncated at $t - w \geq 0$ and $t + w \leq T$. Then, the model learns probability $p(a_{t + \tau} \vert a_{t})$, where $-w\leq \tau\leq w$ and $\tau \neq 0$,  by maximizing its log likelihood given by
%
%
\begin{equation}
	{\cal J} = \frac{1}{T}\sum_{t = 1}^{T} \sum_{-w \leq \tau \leq w, \tau \neq 0} \log p(a_{t + \tau} \vert a_{t}),
\end{equation}
where,

\begin{equation}
	p(j \given i) = \frac{\exp(\bm{u}_j \cdot \bm{v}_{i})}{Z_i}, \label{eq:cond_prob_w2v}
\end{equation}
where $\bm{v}$ and $\bm{u}$ are the ``in-vector'' and ``out-vector'', respectively,  $Z_i=\sum_{j' \in \mathcal{A}} \exp(\bm{u}_{j'} \cdot \bm{v}_{i})$ is a normalization constant, and $\mathcal{A}$ is the set of all locations.
We follow the standard practice and only use the in-vector, $\bm{v}$,  which is known to be superior to the out-vector in link prediction benchmarks~\cite{linzhuo2020hyperbolic, tshitoyan2019mat2vec, garg2018gender, kozlowski2018geometry, hamilton2016diachronic, le2014doc2vec, nakandala2017gendered}.

We used the \textit{word2vec}  implementation in the python package \texttt{gensim}.
The skip-gram negative sampling  \textit{word2vec} model has several tunable hyper-parameters, including the embedding dimension $d$, the size of the context window $w$, the minimum frequency threshold $f_{\min}$, initial learning rate $\alpha$, shape of negative sampling distribution $\gamma$, the number of the negative samples should be drawn $k$, and the number of iterations.
For main results regarding scientific migration, we used $d=300$ and $w=1$, which were the parameters that best explained the flux between locations, though results were robust across different settings (Fig.~\ref{fig:supp:hyperparameter}).
Although the original \textit{word2vec } paper uses $\gamma = 0.75$\cite{mikolov2013word2vec}, here we set $\gamma = 1.0$, though results are only trivially different at different values of $\gamma$ (Fig.~\ref{fig:supp:distancemetrics}). We used $k=5$, which is suggested default of \textit{word2vec}.
We also use same setting for U.S. airport itinerary and Korean accommodation reservation data.

To mitigate the effect of less common locations, we set $f_{\min} = 50$, limiting to locations appearing at least 50 times across the training trajectories, resulting in embeddings reflecting 744 unique airports for the U.S. airport itinerary data, 1,004 unique accommodations for Korean accommodation reservation data, and 6,580 unique organizations for the scientific migration data.
We set $\alpha$ to its default value of 0.025 and iterate five times over all training trajectories.
For scientific migration, across each training iteration, the order of organizations within a single year is randomized to remove unclear sequential order.

\subsection*{Distance}
We calculate $T_{ij}$ as the total number of co-occurrences between two locations $i$ and $j$ across the data-set.
In scientific migration, $T_{ij} = 10$ indicates that the number of co-occurrences between both organization $i$ and $j$ between 2008 and 2019 is 10, as evidenced from their publications.
Here, we treat $T_{ij} = T_{ji}$ for the sake of simplicity and, in the case of scientific migration, because directionality cannot easily be derived from bibliometric records, or may not be particularly informative (see \SI).

We calculate two main forms of distance between locations.
The geographic distance, $g_{ij}$, is the pairwise geographic distance between locations.
Geographic distance is calculated as the great circle distance, in kilometers, between pairs of locations.
In the case of U.S. flight itinerary and scientific migration, we impute distance to 1 km when their distance is less than one kilometer.
In the case of Korean accommodation reservation data, because this data represents trajectories of intra-city mobility that occurs at a much smaller scale international migration, we impute distance to 0.01 km when their distance is less than 0.01 km.
The embedding distance with the cosine distance, $d_{ij}$, is calculated as $d_{ij} = 1 - \frac{\bm{v}_{i} \cdot \bm{v}_{j}}{\lVert \bm{v}_{i} \rVert \lVert \bm{v}_{j} \rVert}$,
where $v_i$ and $v_j$ are the embedding vectors for locations $i$ and $j$, respectively.
Note that $d_{ij}$ is not a formal metric because it does not satisfy the triangle inequality.
Nevertheless, cosine distance is often shown to be useful in practice~\cite{lerman2007embedding, brown1970migration, kim2018functional}.
We compare the performance of this cosine-based embedding distance against those derived using inner product similarity and Euclidean distance.

We compare the performance of the embedding distance to many baselines.
These include distances derived from simpler embedding approaches, such as Singular Value Decomposition (SVD) and a Laplacian Eigenmap embedding performed on the underlying location co-occurrence matrix.
We also use network-based distances, calculating vectors using a Personalized Page Rank approach and measuring the distance between them using cosine distance and Jensen-Shannon Divergence (see \SI).
Finally, we compare the embedding distance against embeddings calculated through direct matrix factorization, following the approach that \textit{word2vec} implicitley approximates~\cite{levy2014neural}.

%
%
\subsection*{Gravity Law}

We model co-occurences $T_{ij}$ for locations $i$ and $j$ (referred to as flux), using the gravity law of mobility~\cite{zipf1946gravity}.
The gravity law of mobility, which was inspired by Newton's law of gravity, postulates that attraction between two locations is a function of their population and the distance between them.
This formulation and variants have proven useful for modeling and predicting many kinds of migration and mobility~\cite{jung2008highwaygravity, curiel2018citygravity, truscott2012epidemicgravity, hong2016busgravity}.
In the gravity law of mobility, the \textit{expected flux}, $\hat{T}_{ij}$ between two locations $i$ and $j$ is defined as,

%
\begin{equation}
	\label{eq:gravity_basic}
	\hat{T}_{ij} = Cm_{i}m_{j}f(r_{ij}),
\end{equation}
where $m_{i}$ and $m_{j}$ are the population of locations, defined as the total number of passengers who passed through each airport for U.S. airport itineraries,  the total number of customers who booked with each accommodation for Korean accommodation reservations, and the yearly-average count of unique authors, both mobile and non-mobile, affiliated with each organization for scientific migration.
$f(r_{ij})$ is a decay function of distance $r_{ij}$ between locations $i$ and $j$.
Here, we used the most basic gravity model which assumes symmetry of the flow $\hat{T}_{ij} =\hat{T}_{ji}$ and distance $r_{ij} = r_{ji}$, while there are four proposed variants \cite{wilson2011entropy}.
There are two popular forms for the $f(r_{ij})$:
one is a power law function in the form $f(r_{ij})= r^{-\alpha}_{ij} \ (\alpha > 0)$, and the other is an exponential function in the form $f(r_{ij}) = e^{-\beta r_{ij}} \ (\beta > 0)$ \cite{chen2015distance}.
The parameters for $f(r_{ij})$ and $C$ are fit to given data using a log-linear regression~\cite{jung2008highwaygravity, curiel2018citygravity, truscott2012epidemicgravity, hong2016busgravity, simini2012universal}.

We consider separate variants of $f(r_{ij})$ for the geographic distance, $g_{ij}$, and the embedding distance, $d_{ij}$, and report the best-fit model of each distance.
For the geographic distance, we use the power-law function of the gravity law, $f(g_{ij})= g^{-\alpha}_{ij}$ (Eq.~\ref{eq:linear_flux_geo}).
For the embedding distance, we use the exponential function, with $f(d_{ij}) = e^{-\beta d_{ij}}$ (Eq.~\ref{eq:linear_flux_emb}).

\begin{equation}
	\label{eq:linear_flux_geo}
	\ln\frac{T_{ij}}{m_im_j} = \ln C  - \alpha \ln g_{ij},
\end{equation}

\begin{equation}
	\label{eq:linear_flux_emb}
	\ln\frac{T_{ij}}{m_im_j} = \ln C - \beta d_{ij},
\end{equation}
where $T_{ij}$ is the actual flow from the data.
The gravity law of mobility is sensitive to $T_{ij} = 0$, or zero movement between locations.
In our dataset, non-zero flows account for only 4.2\% of all possible pairs of the 6,580 organizations for scientific migration, 76.4\% of all possible pairs of the 744 airports for U.S. airport itinerary data, and 62.5\% of all possible pairs of the 1,004 accommodations for Korean accommodation reservation data.
This value is comparable to other common applications of the gravity law, such as phone calls, commuting, and migration \cite{simini2012universal}.
We follow standard practice and exclude zero flows from our analysis.

%
\subsection*{Element-centric clustering similarity}
Element-centric clustering similarity \cite{gates2019element} is a similarity measure that can produce disjoint, overlapping, and hierarchically-structured clusterings.
Element-centric clustering similarity captures cluster-induced relationships between elements through a cluster affiliation graph where one vertex set is the original element $V=\{v_1, ... v_N\}$ and the other corresponds to the cluster $C=\{c_1, ... c_M\}$ as a bipartite graph $\mathcal{B} (V \cup C, \mathcal{R})$.
An undirected edge $a_{i\beta} \in \mathcal{R}$ denotes element $v_i$ as a member of cluster $c_\beta$.
For hierarchically structured clustering, each cluster $c_\beta$ is assigned a hierarchical level $l_{\beta} \in [0,1]$ by re-scaling the dendrogram according to the maximum path length from its roots.
The weight of the cluster affiliation edge is given by the hierarchy weight function $a_{i\beta}  = e^{rl_{\beta}}$ with scaling parameter $r$ which determines the relative importance of membership at different levels of hierarchy.
In this context, smaller $r$ gives more importance to clusters that are closer to the root, prioritizing higher levels in the hierarchy. 
The lower levels are treated as a refinement of the higher level.
Conversely, larger $r$  places greater emphasis on the lower-level cluster structure, while viewing the higher levels of the hierarchy as an aggregation of the lower-level structure. 
When $r=0$, equal importance is assumed for every cluster.

The cluster affiliation graph is projected onto a cluster-induced element graph which is a weighted, directed graph summarizing the relationship induced by common cluster memberships.
In the cluster-induced element graph, each edge  between element $v_i$ and $v_j$ has weight $w_{ij} = \sum_{\gamma} \frac{a_{i\gamma}a_{j\gamma}}{\sum_{\kappa}a_{i\kappa} {\sum_{m}a_{m\gamma}}}$.
Given a cluster-induces element graph with weighted matrix $W$, the personalized PageRank vector $p_i$  is used as membership-aware similarity between element $i$ and other elements in the graph.
Then, the element-wise similarity of an element $v_i$ in two clusters  $\mathcal{A}$ and $\mathcal{B}$ is calculated with $S_i(\mathcal{A}, \mathcal{B}) = 1 - L_1(p_i^{\mathcal{A}},p_j^{\mathcal{B}})$, and the final element-centric similarity of two clustering $\mathcal{A}$ and $\mathcal{B}$ is found as the average of the element-wise similarities, $S(\mathcal{A}, \mathcal{B}) =\frac{1}{N}\sum_{i=1}^{N} S_i(\mathcal{A}, \mathcal{B})$.

%
\subsection*{SemAxis}
SemAxis and similar studies~\cite{an2018semaxis, nakandala2017gendered, kozlowski2018geometry} demonstrated that ``semantic axes'' can be found from an embedding space by defining the ``poles'' and that the latent semantic relationship along the semantic axis can be extracted with simple arithmetic.
In the case of natural language, the poles of the axis could be ``good'' and ``bad'', ``surprising'' and ``unsurprising'', or ``masculine'' and ``feminine''.
We can use SemAxis to leverage the semantic properties of the embedding vectors to operationalize abstract relationships between organizations.

Let $S^+ = \{\bm{v}_1^+, \bm{v}_2^+ \cdots \bm{v}_n^+\}$ and $S^- =\{\bm{v}_1^-, \bm{v}_2^- \cdots \bm{v}_n^-\}$ be the set of positive and negative pole organization vectors respectively.
Then, the average vectors of each set can be calculated as $\bm{V}^{+} = \frac{1}{n}\sum^n_{i=1} \bm{v}_i^+$ and $\bm{V}^{-} = \frac{1}{n}\sum^n_{i=1} \bm{v}_i^-$.
From these average vectors of each set of poles, the semantic axis is defined as $\bm{V}_{\text{axis}} = \bm{V}^{+} - \bm{V}^{-}$. Then,  a score of organization $a$ is calculated as the cosine similarity of the organization's vector with the axis,

\begin{equation}
	\frac{\bm{v}_{a} \cdot \bm{V}_{\text{axis}}}{\lVert \bm{v}_{a} \rVert \lVert \bm{V}_{\text{axis}} \rVert},
\end{equation}
where a higher score for organization $a$ indicates that $a$ is more closely aligned to $V^+$ than $V^-$.

We define two axes to capture geography and academic prestige, respectively.
The poles of the geographic axis are defined as the mean vector of all vectors corresponding to organizations in California, and then the mean of all vectors of organizations in Massachusetts.
For the prestige axis, we define a subset of top-ranked universities according to either the Times World University Ranking or based on the mean normalized research impact sourced from the Leiden Ranking.
The other end of the prestige axis is the geographically-matched (according to census region) set of universities ranked at the bottom of these rankings.
For example, if 20 top-ranked universities are selected and six of them are in the Northeastern U.S., then the bottom twenty will be chosen to also include six from the Northeastern U.S.
From the prestige axis, we derive a ranking of universities that we then compare to other formal university rankings using Spearman rank correlation.

\section*{Acknowledgement}
We thank the Center for Science and Technology Studies at Leiden University for managing and making available the dataset of scientific migration. We also thank the Goodchoice Company LTD. for making available the dataset of Korean accommodation reservation data.
For their comments, we thank Guillaume Cabanac, Cassidy R. Sugimoto, Vincent Lariviére, Alessandro Flammini, Filippo Menczer, Lili Miao, Xiaoran Yan, Inho Hong, and Esteban Moro Egido.
This material is based upon work supported by the Air Force Office of Scientific Research under award number FA9550-19-1-0391.
Jisung Yoon would like to acknowledge the support of the National Science Foundation Grant Award Number EF-2133863.
Rodrigo Costas is partially funded by the South African DST-NRF Centre of Excellence in Scientometrics and Science, Technology and Innovation Policy (SciSTIP).

\printbibliography

\newpage

\section*{Supporting Information: Unsupervised embedding of trajectories captures the latent structure of scientific migration} 

\beginsupplement

\makeatletter
\renewcommand \thesection{S\@arabic\c@section}
\renewcommand\thetable{S\@arabic\c@table}
\renewcommand \thefigure{S\@arabic\c@figure}
\makeatother

\paragraph*{S1 Text}
\label{si:text:mobility_science}
{\bf Mobility and science.}

As scholars move, they bring their knowledge, skills, and social connections with them---collectively the movements of researchers shape the structure and direction of the global scientific enterprise.
For example, prestige-driven mobility between doctoral-granting and employing institutions is highly unequal~\autocite{clauset2015hierarchy, deville2014career}, which affects the diffusion of ideas across academia~\autocite{morgan2018prestige}.
By placing researchers in new social settings, mobility can lead to the formation of new collaborative relationships~\autocite{rodrigues2016mobility}, which in turn spurs the further diffusion of knowledge and innovations~\autocite{braunerhjelm2020labor, azoulay2011diffusion, kaiser2018innovation, armano2017innovation}.
Perhaps resulting from the selection effects of who gets to move, or the reconfiguring of social and epistemic networks, movement is associated with increased scientific impact~\autocite{sugimoto2017mostimpact, petersen2018multiscale, jonkers2013return, franzoni2014advantage}.
At the national level, the understanding of mobility has progressed beyond simplistic narratives of brain drain and brain gain, and instead adopts a new perspective of \textit{flows} of talent~\autocite{meyer2001network, ioannidis2014braindrain, gaillard1998circulation}.
Under this flow model, a mobile researcher is viewed as contributing to both their origin and destination countries, a perspective that highlights the benefits of mobility to both the origin and destination countries. 
Perhaps because of these individual and national benefits, policy-makers have come to recognize the importance of global mobility~\autocite{box2008competition, oecd2010innovation}.
Movement is a key mechanism that has clear impact on the composition and direction of the global scientific workforce and our collective scientific understanding.
Understanding the structure and dynamics of mobility is thus essential for understanding global science.

%
\paragraph*{S2 Text}
\label{si:text:mobility_models}
{\bf Modeling scientific mobility.}

Modeling scientific mobility requires many considerations. 
First, the unit of analysis must be set. 
Most studies of mobility have focused on \textit{country-level} mobility--the flows of researchers across nations~\autocite{sugimoto2017mostimpact, scellato2015migrant, robinson-garcia2018indicators, franzoni2012foreign-born}.
Practically, country-level analyses benefit from higher reliability, such that idiosyncrasies and errors inherent to bibliographic databases are mitigated by this higher level of aggregation.
Epistemically, country-level analysis is useful for national science governance which aims to understand the status of their country in the global landscape and make informed policy decisions.
Analyses at lower units of analysis are far less common.
\textit{Regional}-level scientific mobility--the flow of researchers between regions or cities within or across countries has been only minimally studied~\autocite{vaccario2019mobility}, possibly due to lack of reliable long-term data and lack of policy relevance to national-level lawmakers.
\textit{Organization}-level mobility has the potential to inform institutional policy and to understand the composition of mobility within a single country or region, especially as it relates to organization performance, prestige, and inequality~\autocite{albarran2017topeconomic, deville2014career, morgan2018prestige, clauset2015hierarchy}.
However, affiliation disambiguation and noise in bibliographic data makes large-scale organization-level analysis challenging.
Here, we learn neural-network embeddings of scientific mobility at the level of organizations using a curated bibliographic database.
These embeddings are robust to noise, and so are capable of representing clear structure despite issues with organizational disambiguation.
In doing so, embeddings also capture a more detailed understanding of mobility than has been previously studied.

Another consideration when analyzing scientific mobility is what kinds of mobility to study.
Typical understandings of mobility are directional: movement is always \textit{from} one place and \textit{to} another.
However, scientific mobility is more complicated.
For example, scientists often hold multiple affiliations at a time~\autocite{markova2016synchronous}, listing them as co-affiliations on a single paper, or even choosing a subset of affiliations to use across multiple simultaneous projects~\autocite{robinson2019mobility}.
Even clearly-directional migration to another institution is complex--researchers may continue to publish with an old affiliation for projects that began before their move, and they may maintain social and organizational links to their old institution (e.g., collaborators, projects, graduate students) such that there is no clear breakage after migrating.
There is also a whole range of short-term scientific mobility, such as visiting scholarships and short-term visits that are only visible through intensive efforts such as manual extraction from CVs~\autocite{woolley2009cv, sandstrom2009cv, canibano2011temporary}.
Here, we focus on more long-term mobility that can be derived from bibliographic data.
Due to the complexity of scientific mobility, we make the simplifying assumption that all scientific mobility is \textit{symmetric} or without direction such that any move from an organization $A$ to organization $B$ is equivalent to a move from $B$ to $A$.
By assuming non-directional mobility, all mobility events are commensurate, meaning that they can be treated identically in our analysis--this allows us to represent the complexity of mobility without making decisions regarding the directionality of researchers' mobility or main affiliation.
Moreover, this assumption has the practical advantage of matching the data format expected by the \textit{word2vec} model, as well as the theoretical advantage of adhering to the symmetricity assumption of the gravity model of mobility.

%
\paragraph*{S3 Text}
\label{si:text:mobility_traj}
{\bf Building affiliation trajectories.}

For each mobile researcher who has at least two distinct affiliations, we construct an affiliation trajectory based on the affiliations listed on their published papers indexed in the Web of Science database between 2008 and 2019.
An author is considered mobile if they published with at least two distinct affiliations during the time period of study.
Affiliation names were manually disambiguated, and each was mapped to a unique organization identifier.
An affiliation trajectory for an individual researcher is a sequence of organizations in ascending order of year of publication.
If a researcher published papers with affiliation $A$ in year $t$, $B$ in $t+1$, $C$ in $t+2$ and $A$ again in $t+3$, then the affiliation trajectory is expressed as $(A, B, C, A)$.

In the case that an individual lists multiple affiliations in a single year, affiliations listed on publications published in that year are shuffled between each iteration of the \textit{word2vec} training process (each epoch).
For example, an author who published with affiliation $A$ in $t_{0}$, and affiliations $B$ and $C$ in $t_{1}$ could appear as one of $(A, B, C)$ or $(A, C, B)$ in each training iteration.
This effectively removes the effect of order within a year, as the order cannot be meaningfully established based on co-affiliations in a single paper, or on different affiliations listed on separate papers, for which the date of publication may not be representative of the actual completion of the project.

Other than restricting to only mobile researchers, we do not perform any filtering or reductions to affiliation trajectories.
In the case that an author publishes with organization $A$ four times in $t_{0}$, and affiliation $B$ two times in $t_{1}$,  their trajectory will be $(A, A, A, A, B, B)$.
Although mobile authors who publish more papers will have longer trajectories, \textit{word2vec} will skip duplicate consecutive organization IDs, mitigating the impact of long repetitive trajectories.

\paragraph*{S4 Text}
\label{si:text:synthetic_exp}
{\bf Demonstration on ability of \textit{word2vec} to effectively represent gravity-like relationships with a simulation experiment}

We conducted an additional experiment to test the ability of \textit{word2vec} to recover the metric between locations, and to further demonstrate its equivalency with the gravity model. Specifically, we perform a synthetic benchmark to test whether \textit{word2vec} can effectively recover gravity-like relationships in data that is known to contain them.

We start by generating synthetic mobility trajectories structured according to the gravity model. For this purpose, we first generate an artificial ``geographic'' space. 1,000 points are generated in a 2-dimensional space, with their coordinates distributed evenly between -1 and 1. These points act as the ``locations'' in our simulation. Each point is also assigned a ``mass'' randomly drawn from a power-law distribution $p(x) \propto x^{-k}$ (we use $k=2$) that is meant to represent the ``population'' or ``size'' of each location. 

With this simulated geography in place, we calculate the pairwise distance matrix $r_{ij}^{synthetic}$, which is then used to calculate a transition matrix between points, defined as $T_{ij} = M_i M_j f(r_{ij})$, with transition probabilities following the gravity law. The purpose of this transition matrix is to define the likelihood of movement between points (or \textit{locations}) in our simulated geography based on their mass and pairwise distances. We conduct this experiment with two different transition matrices, one based on cosine distance where $f(r_{ij})=e^{-\beta r_{ij}}$ and another with Euclidean distance where $f(r_{ij})=r_{ij}^{-\alpha}$. For simplicity, we use $\beta=2$ and $\alpha=2$. 

Finally, synthetic trajectories are generated based on the probabilities in the transition matrix.
Specifically, we use a random walker based approach, where random walkers are initialized in locations with a probability proportional to their mass and transition between points according to the transition matrix. 
These synthetic trajectories are then used to train the skip-gram \textit{word2vec} model. 
For parameters, we set window size to $w=1$ and the embedding dimension to $d=100$. 

From the embedding trained with synthetic trajectories, we calculate the pairwise embedding distances between points, $r_{ij}^{embedding}$.
To evaluate the ability of \textit{word2vec} to recover the known gravity-like structure of the synthetic data, we compare the embedding distance to the distances between the 1,000 points in the synthetic space, $r_{ij}^{synthetic}$ using the Pearson correlation coefficient. 
This result is repeated 100 times, and the correlation coefficients are aggregated across iterations. 

We find that \textit{word2vec} effectively recovers the gravity-like relationships in the synthetic data. 
Based on cosine distance, the Pearson correlation between the embedding and synthetic distances is 0.943, with a standard error of 0.01. 
Based on Euclidean distance, the correlation was 0.801, with a standard error of 0.01.
To better illustrate this point, Fig. \ref{fig:supp:synthetic_exp} shows the correlation for one representative iteration of this experiment.

In order to evaluate the robustness of the outcome across hyperparameter settings, we repeat our synthetic benchmark (using cosine distance) varying values for the mass exponent, $k$ and distance exponent $\beta$. 
Fig. \ref{fig:supp:synthetic_robust} presents the average Pearson correlation between the synthetic and embedding distance over ten iterations with for different values of $k$ and $\beta$. 
This shows our findings to be robust. 
Neural embedding consistently recovers the distance from the synthetic trajectories across a broad range of hyperparameter values. 
A low correlation occurs only when the mass distribution is too heterogeneous ($k < 1.9$).
In a heterogeneous mass distribution, most of the trajectories will concentrate on just a few high-mass locations, regardless of distance, which makes learning a distance challenging.  Meanwhile, the exponent of the city population usually ranges between 2 and 3~\autocite{clauset2009power}.

\

%
\paragraph*{S5 Text}
\label{si:text:ppr_dist}
{\bf Network-based personalized Page Rank distances.}

We examine the gravity model on the Personalized Page Rank (PPR)\autocite{jeh2003scaling} as a benchmark on the network. We construct the co-occurrence network of $N$ organizations, in which each edge between organizations $i$ and $j$ represents a co-occurence of $i$ and $j$ in the same affiliation trajectory, with weight given by the sum of the co-occurences over all researchers. and edges are co-occurrence between two organizations. The Personalized Page Rank is a ranking algorithm for nodes based on a random walk process on networks.
At each step, the walker visiting a node moves to a neighboring node chosen randomly with a probability proportional to the weight of the edge. Furthermore, with probability $\alpha$, the walker is teleported back to the starting node. The rank of a node is determined by the probability that the walker visits the node in the stationary state. The stationary distribution of the random walker starting from node $i$, denoted by $p_i=(p_{ik})$, is given by

\begin{equation}
	\label{eq:ppr}
	p_i = (1 - \alpha) W p_i +  \alpha v_i 
\end{equation}
where $v_i$ is a column vector of length $N$ with entries that are all zero except the $i$th entry that equals one, $W = (W_{ij})$ is the weighted transition matrix of the unbiased random walker. We used $\alpha=0.9$ here.

We can think of $p_i$ as a representation vector of the organization $i$, and calculate the distance between organizations $i$ and $j$, $d_{ij}$, with measuring distance between $p_i$ and $p_j$ to examine the gravity law. We consider two distance measures in this analysis. The first one is cosine distance which is used for our embedding method, $d_{ij} = 1 - \frac{\bm{p}_{i} \cdot \bm{p}_{j}}{\lVert \bm{p}_{i} \rVert \lVert \bm{p}_{j} \rVert}$. 
Also, if we think of $p_i$ as a discrete probability distribution, then we can consider Jensen–Shannon divergence (JSD), which can be written as,

\begin{equation}
	\label{eq:JSD}
	d_{ij} = JSD(p_i||p_j) = \frac{1}{2}D_{KL}(p_i||m) + \frac{1}{2}D_{KL}(p_j||m),
\end{equation}

\begin{equation}
	\label{eq:KL}
	D_{KL}(p_i||m) = \sum^x p_{ix}\log\frac{p_{ix}}{m_x},
\end{equation}
where $m=\frac{1}{2}(p_i+p_j)$. We report the result with cosine distance ($R^2=0.14$, Fig. \ref{fig:supp:gravity_pprcos}) and  Jensen–Shannon divergence ($R^2=0.19$, Fig. \ref{fig:supp:gravity_pprjsd}). In both cases, the performance is under the performance of the model with geographical distance. 
Even though the length of the PPR vectors is much larger than the length of our embedding vectors, the result with the embedding distance outperforms both of them.

%
\paragraph*{S6 Text}
\label{si:text:Raw SVD}
{\bf Singular value decomposition distance.}

We use the truncated singular value decomposition (SVD) on the underlying mobility co-occurrence matrix as a baseline embedding.
In short, truncated SVD performs linear low-rank approximation of the matrix with given dimensions, $d$.
First, we construct the co-occurrence matrix of $N$ organizations, $A_{ij}$ given by the co-occurrence of organizations $i$ and $j$ in the same affiliation trajectory.
Then, we apply truncated singular value decomposition with $d=300$ on the flow matrix $A$ directly.

We calculate the distance between organizations in the SVD embedding space using cosine distance, finding that it explains slighly more of the flux between organizations than geographic distance ($R^2=0.247$, Table~\ref{supp:table:r2_table}).
When used as an input to the gravity model, this distance produces better predictions than geographic distance using both the exponential (RMSE $= 0.859$, Table~\ref{supp:table:rmse_exp_table}) and power-law models (RMSE $= 0.839$, Table~\ref{supp:table:rmse_exp_table}), performing slightly better with the power-law formulation.

%
\paragraph*{S7 Text}
\label{si:text:Laplacian}
{\bf Laplacian Eigenmap distance.}

We also consider Laplacian Eigenmap embeddings~\autocite{belkin2003laplacian} as a baseline, which is one of the most fundamental approaches for graph embedding.
First, we construct the co-occurrence matrix of $N$ organizations, $A_{ij}$ given by the co-occurrence of organizations $i$ and $j$ in the same affiliation trajectory and degree matrix $D$ which is the diagonal matrix for which $D_{ii}=\sum_j {A_{ij}}$. Then we construct graph Lapalcian matrix $L = D-A$  and apply truncated singular value decomposition in the matrix $L$ with $d=300$.

We only report results based on the cosine distance between Laplacian embedding vectors, finding that it explains less of the total flux than geographic distance ($R^{2} = 0.212$, Table~\ref{supp:table:r2_table}).
When used as an input to the gravity model, the Laplacian cosine distance produces marginally-better predictions than geographic distance using both the exponential (RMSE $= 0.878$, Table~\ref{supp:table:rmse_exp_table}) and power-law models (RMSE = $0.87$, Table~\ref{supp:table:rmse_power_table}), performing slightly better with the power-law formulation.

%
\paragraph*{S8 Text}
\label{si:text:levy}
{\bf Levy's  symmetric SVD \textit{word2vec} distance}

We also compare the \textit{word2vec} embedding distance against a baseline of direct matrix factorization approach, using the symmetric SVD  \textit{word2vec} method~\autocite{levy2014neural}.
Based on notion that \textit{word2vec} is equivalent to implicit matrix factorization, Levy proposed symmetric SVD \textit{word2vec} embedding, which should directly compute the embedding that \textit{word2vec} only attempts to efficiently approximate.
First, we construct the matrix of $N$ organizations

\begin{equation}
 M_{ij} = \log\left( \frac{N(i,j) |D|}{N(i) N(j)} \right)- \log k,
\end{equation}
where $N(i,j)$ is the number of times the location pair $(i,j)$ appears given the window size $w$ in the total corpus $D$, $N(i)=\sum_{j=N}^i N(i,j)$ as the number of items $i$ occurred given the window size $w$ in $D$, and $k$ is the number of negative samples. We used $w=1$ and $k=5$ which is the same setting as in our main result. Then, we factorized matrix $M$ with truncated singular value decomposition in the matrix with $d=300$ into $U_d \Sigma_d V_d$, and used the embedding vector as $U_d\sqrt{\Sigma_d}$.

For this baseline, we report results using the cosine distance, Euclidean distance, and dot product between the embedding vectors.
We find that the dot product performs by far the worst, worse than any other baseline considered $R^{2} = 0.004$, Table~\ref{supp:table:r2_table}).
The cosine distance performs better, but worse than geographic distance ($R^{2} = 0.212$, Table~\ref{supp:table:r2_table}).
The Euclidean distance performs best, explaining more of the flux than geographic distance, and only being below the embedding distance ($R^{2} = 0.341$, Table~\ref{supp:table:r2_table}).
Focusing on the Euclidean distance, we find that using it as input to the gravity model results in better predictions than geographic distance using both the exponential model (RMSE $= 0.803$, Table~\ref{supp:table:rmse_exp_table}) and power law models, (RMSE $= 0.78$, Table~\ref{supp:table:rmse_power_table}), though it performs slightly better with the power law formulation.

%
\paragraph*{S9 Text}
\label{si:text:direct}
{\bf Direct optimization of gravity model}

Finally, we create an embedding by optimizing for the gravity law directly. 
Specifically, we construct a gravity matrix of $N$ organizations, where each cell of the matrix is calculated as $T_{ij}m_i m_j$, where $T_{ij}$ is co-occurrence of organizations $i$ and $j$ in the same affiliation trajectory, and $m_i$ and $m_j$ are the organizations' populations, defined here as the mean annual number of unique mobile and non-mobile authors affiliated.

We then embed the gravity matrix using two approaches: a truncated singular value decomposition (SVD) and multidimensional scaling (MDS), which embeds each location in a N-dimensional space such that pairwise distances are preserved as well as possible. 
Typically, MDS uses Euclidean distance to measure distance between vectors. 

For this analysis, we report results using the cosine distance between embedding vectors for the SVD embedding, and the logged Euclidean distance for the MDS embedding. 
We observe that the gravity-optimized SVD cosine distance has poor performance, more weakly correlated with actual flux than geographic distance ($R^{2} = 0.122$, Table~\ref{supp:table:r2_table}), and similarly poor prediction error when used as input to the gravity mode.
In contrast, the cosine distance between MDS vectors has the second-highest correlation with actual flux, after only the embedding distance ($R^{2} = 0.355$, Table~\ref{supp:table:r2_table}), and third best for predicting actual flux using both the power-law version of the gravity model (RMSE = $0.795$, Table~\ref{supp:table:rmse_power_table}) and the exponential form (RMSE = $0.904$, Table~\ref{supp:table:rmse_exp_table}). 
We note that the MDS embedding actually has the lowest error when organizations' populations are defined as the raw frequency when using the power-law model (RMSE = $0.691$, Table~\ref{supp:table:rmse_power_table}).
However, we note that the MDS distance is defined with a population of ``all'' mobile and non-mobile scholars, whereas the predictions made in Table~\ref{supp:table:rmse_power_table} use the raw frequency, which likely confounds this result. 
MDS is also computationally-intensive, requiring upwards of 6 times more time to compute (on the machine used in this analysis) than the more computationally efficient \textit{word2vec} model.

%
\paragraph*{S10 Text}
\label{si:text:whynueralbetter}
{\bf Why do neural embeddings outperform direct matrix factorization?}

Why do neural-embedding approaches, which rely on stochastic gradient descent, outperform Levy's direct matrix factorization~\autocite{levy2014neural}, especially given that \textit{word2vec} \emph{is} implicitly approximating factorization?
We speculate that is stems, in part, from the sensitivity of matrix factorization to small flows between locations.  
Levy's matrix factorization embeds affiliations $i$ and $j$ such that their dot similarity is as close as possible to $\log (T_{ij} / P(i) P(j))$. 
If the flow $T_{ij}$ is considerably small or zero, the dot similarity goes to $-\infty$, pushing $i$ and $j$ very far from other affiliations in the embedding space~\autocite{levy2014neural, Qui2018}. 
This is particularly problematic when the window size is small because most affiliation pairs would have no flow, which is indeed the case in our experiments. 
To circumvent this problem, previous studies~\autocite{levy2014neural, Qui2018} added a constant flow between the affiliation pairs with no flow. 
However, in addition to altering the underlying data, these small flows can still have a strong impact on the embedding.

It is also well known that singular value decomposition (SVD) is vulnerable to outliers~\autocite{xu2012robust, huber1981robust, xu1995robust, chandrasekaran2011rank, candes2011robust}.
The stochastic gradient descent algorithm, which is employed in SGNS \textit{word2vec}, is more robust than SVD and can enhance generalization and effectiveness of {\it word2vec} model \autocite{ma2018power, smith2020generalization, zhang2019algorithmic}.

%
\paragraph*{S11 Text}
\label{si:text:organizations}
{\bf Organization disambiguation and metadata.}

Affiliations are mapped to one of 8,661 organizations and disambiguated following the procedure originally designed for the Leiden Rankings of World Universities~\autocite{waltman2012leidenrankings}.
Organizational records were associated with a full name, a type indicating the sector (e.g., University, Government, Industry), and an identifier for the country and city of the organization.
Sixteen different sector types were included in the analysis, which we aggregated to four high-level codes: \textit{University}, \textit{Hospital}, \textit{Government}, and \textit{Other}.
Each record was also associated with a latitude and longitude.
However, for many organizations, these geographic coordinates were missing or incorrect.
We manually updated the coordinates of 2,267 organizations by searching the institution name and city on Google Maps;
in cases where a precise location of the organization could not be identified, we used the coordinates returned when searching the name of the city.
The data was further enriched with country-level information, including region, most widely-spoken language, and its language family (e.g., the language family of \textit{Spanish} is \textit{Italic}).
State/province-level information was added using the reverse geocoding service LocationIQ using each organization's latitude and longitude as input.
Regional census classifications were added for states in the United States.
For each organization, we calculated size as the average number of unique authors (mobile and non-mobile) who published with that organization across each year of our dataset;
in the case that authors publish with multiple affiliations in a single year, they are counted towards each.

As a result of our disambiguation procedure, some affiliations are mapped to two organizations, one specific, and one more general.
For example, any author affiliated with ``Indiana University Bloomington'' will also be listed as being affiliated with the ``Indiana University System'', a more general designation for all public universities in Indiana.
However, a more general organization may not always occur alongside the more specific one.
For example, a researcher affiliated with the smaller regional school ``Indiana University South Bend'' will be listed as affiliated with only the ``Indiana University System''.
We identify all specific organizations that always co-occur along with a more general one.
For every career trajectory that includes one of these specific organizations, we remove all occurrences of the more general organization;
trajectories containing only a general designation are not altered.

%
\paragraph*{S12 Text}
\label{si:text:disambiguation}
{\bf Author name disambiguation.}

Author-name disambiguation, the problem of associating names on papers with individuals authors, remains difficult for the use of bibliographic data~\autocite{dangelo2020disambiguation}.
Authors in our dataset have been disambiguated using a rule-based algorithm that makes use of author and paper metadata, such as physical addresses, co-authors, and journal, to score papers on the likelihood of belonging to an author cluster---a cluster of publications believed to have been authored by the same individual~\autocite{caron2014disambiguation}.
We limit our analysis to the period from 2008 to 2019, as in 2008 the Web of Science began indexing additional author-level metadata such as full names and email addresses.
The disambiguation algorithm is conservative, favoring splitting clusters over merging.
Past studies have validated this data and shown that the disambiguated authors are comparable to ground-truth records such as those from ORCID and useful for a wide range of bibliometric studies~\autocite{sugimoto2017mostimpact, robinson2019mobility, chinchilla2018global, chinchilla2018travelban}.

\paragraph*{S13 Text}
\label{si:text:nce}
{\bf Derivation of Eq. 5 - Noise Contrastive Estimation}

The noise contrastive estimation (NCE)~\autocite{Chia2010,Dyer2014}.
NCE is an unbiased estimator for a probability model $P_m$ of the form:
\begin{align}
	\label{eq:nce-model}
	P_m(x) = \frac{ f(x) }{\sum_{ x' \in {\cal X}} f(x') },
\end{align}
where $f$ is a non-negative likelihood function of data $x$, and ${\cal X}$ is the set of all data.
This general form includes the \textit{word2vec} model (Eq. 2), where $f(x) = \exp(x)$ and $x = \vect{u}_j \cdot \vect{v}_{i}$.
NCE fits the probability model using a binary classification task in the same way as in negative sampling but using a Bayesian formalism for logistic regression.
Specifically, before the training, we know that $1$ in $1+k$ words is sampled from the given data, which can be modeled as prior probabilities
\begin{align}
	\label{eq:prior}
	P(Y_{j} = 1) = \frac{1}{k + 1},\quad P(Y_{j} = 0) = \frac{k}{k + 1}.
\end{align}
Using the Bayes rule, the posterior probability for $Y_{j}$ given word $j$ is given by
\begin{align}
	\label{eq:posterior}
	P\left(Y_{j} \vert j\right) = \frac{
		P\left(j \vert Y_{j}\right)P(Y_{j})
	}{
		P\left(j \vert Y_{j} = 0\right)P(Y_{j} = 0)
		+ P\left(j \vert Y_{j} = 1\right)P(Y_{j} = 1)
	}.
\end{align}
Bearing in mind that word $j$ is sampled from the given data if $Y_{j}=1$ and from the noise distribution $p_0$ if $Y_j = 0$.
Assuming that the given data is generated from the probability model to fit, the class-conditional probability, $P\left(j \vert Y_{j}\right)$, is given by
\begin{align}
	\label{eq:class-cond}
	P(j \vert Y_{j} = 1) & = P_m(\vect{u}_j \cdot \vect{v}_{i}),
	\quad
	P(j \vert Y_{j} = 0)  = p_0 (j).
\end{align}
Putting Eqs.~\eqref{eq:prior}, \eqref{eq:posterior} and \eqref{eq:class-cond} together, the posterior probability for $Y_j$ is given by
\begin{align}
	\label{eq:nce}
	P\left(Y_{j} = 1 \vert j\right) & =
	\frac{
		P_m(\vect{u}_j \cdot \vect{v}_{i}) / (k + 1)
	}{
		P_m(\vect{u}_j \cdot \vect{v}_{i}) / (k + 1)  + kp_0(j) / (k + 1)
	}                                          \\
	                                & = \frac{
		P_m(\vect{u}_j \cdot \vect{v}_{i})
	}{
		P_m(\vect{u}_j \cdot \vect{v}_{i})  + kp_0(j)
	},
\end{align}
which can be rewritten in form of sigmoid function:
\begin{align}
	\label{eq:nce}
	P^{\text{NCE}}\left(Y_{j}=1 \vert j\right) =
	 & = \frac{
		1
	}{
		1 + kp_0(j) / P_m(\vect{u}_j \cdot \vect{v}_{i})
	}           \\
	 & = \frac{
		1
	}{
		1 + \exp\left[ \ln kp_0(j) - \ln P_m(\vect{u}_j \cdot \vect{v}_{i}) \right]
	}           \\
	 & = \frac{
		1
	}{
		1 + \exp\left[ - \ln f(\vect{u}_j \cdot \vect{v}_{i})  + \ln p_0(j) + c \right]
	},
\end{align}
where $c = \ln k + \ln\sum_{ x' \in {\cal X}} f(x') $ is a constant.
NCE maximizes the log-likelihood
\begin{align}
	\label{eq:log-likelihood-logistic-regress}
	{\cal J}^{\text{NCE}} = \sum_{i \in {\cal A}}\sum_{j \in {\cal D}} \left[ Y_{j} \log P^{\text{NCE}}(Y_{j} = 1\vert j) + (1-Y_{j}) \log P^{\text{NCE}}(Y_{j} = 0 \vert j)\right].
\end{align}
by calculating the gradients for embedding vectors $\vect{u}_j$, $\vect{v}_{i}$ and iteratively updating them.
An important consequence of this framework is that NCE is an unbiased estimator that has convergence to the optimal embedding in terms of the original \textit{word2vec's} objective function, ${\cal J}$~ if we increase the number of words to sample and the training iterations \autocite{Chia2010,Dyer2014}.

%
\paragraph*{S14 Text}
\label{si:text:network_ranking}
{\bf Reconstructing Times ranking with network measure.}

The performance of the embedding ranking in reconstructing the Times ranking is comparable to that of network-derived measures such as degree strength (Spearman's $\rho = 0.73$, Fig.~\ref{fig:supp:centrality_times_compare}a) and eigenvector centrality (Spearman's $\rho = 0.76$, Fig.~\ref{fig:supp:centrality_times_compare}b).
However, while both embedding- and network-based measures relate to university prestige, they are qualitatively and quantitatively different.
The embedding-ranking of U.S. universities is less correlated with degree strength (Spearman's $\rho = 0.45$, Fig.~\ref{fig:supp:centrality_semaxis_compare}a) and eigenvector  centrality (Spearman's $\rho = 0.55$) than with the Times ranking itself (Spearman's $\rho = 0.73$, Fig.~\ref{fig:supp:centrality_semaxis_compare}b).
The embedding ranking over-ranks large research-intensive universities such as North Carolina State University, University of Florida, and Texas A\&M University, whereas the network-derived ranking over-ranks smaller, more specialized universities such as Brandeis University, Yeshiva University, and University of San Francisco.
This suggests that the embedding encodes information on prestige hierarchy at least as well as a network representation, with some noticeable qualitative differences.

%
\paragraph*{S15 Text}
\label{si:text:boomerang_speculation}
{\bf Speculation on variations of the concave-curve pattern.}

The concave-curve pattern observed in Fig.5 repeats across many countries, with variations.
We attempt to explain these variations through the lens of the ``specificity'' theory introduced in previous research~\autocite{schakel2015measuring};
that is, more ``specific'' terms appear in a limited contexts, resulting in a longer vector, whereas more ``universal'' terms appear across many contexts, resulting in a short vector.
We note that our own simulations (S16 Text) do not necessarily support this as the sole mechanism driving the concave-curve pattern, and so more work is needed to confirm and understand the factors underlying this pattern.
Still, differences in the curve by country are consistent with the view that the length of a vector roughly corresponds to an organization's degree of specificity or universality. 

To illustrate the country-level differences in average organizational vector length, the representative vector of Chinese organizations has a larger norm than that of the U.S. ($\bar{l} = 2.97$ vs $\bar{l} = 2.39$, Table~\ref{table:supp:norm}), causing its curve to be shifted upwards with a larger peak vector norm;
this may reflect a tendency for organizations in the U.S. to appear more frequently in different contexts than Chinese organizations.
Other nations such as Poland, Iran, and Turkey show a linear relationship between an organization's number of researchers and the vector norm, indicating that their largest organizations belong to very specific contexts (Fig.~\ref{fig:concave30}).
The organization-level distribution of vector norms reveals deeper heterogeneity.
The distribution of the vector norms for the U.S. is relatively skewed, suggesting their large norm is driven by a small and tight community of organizations (\textit{skew}$ = -0.82$, Fig.~\ref{fig:norm_dist}).
Germany and the U.K. have comparable representative vector norms to the U.S. ($\bar{l} = 2.6$ and $\bar{l} = 2.61$, respectively), with lower skewness (\textit{skew}$ = -0.63$ and \textit{skew}$ = -0.55$), suggesting more tight community of organizations.
The vector norms of organizations in some countries are even more skewed, such as in Iran ($\bar{l} = 3.57$, \textit{skew}$ = -2.13$) and China ($\bar{l} = 2.97$, \textit{skew}$ = -1.08$), indicating the strong difference between their most- and least-connected organizations.
For some countries, their organizations are positively-skewed, though seemingly for different reasons.
For example, Austria has a balanced distribution of organization vector norms, suggesting a diverse range of organizations with most being well connected ($\bar{l} =2.64$, $s = 0.18$);
Russia, in contrast, has a number of organization vectors of moderate norms, but also several isolated organizations with large vector norms ($\bar{l} = 3.08$, $s = 0.67$).

\paragraph*{S16 Text}
\label{si:text:boomerang_explain}
{\bf Experiment on the origin of the concave-curve pattern.}
We conduct a series of experiments to further investigate the concave-curve pattern between an organization's size (in total number of researchers) and the L2 norm of its embedding vector.
We divide our explanation according to the two halves of the concave-curve: the ``rising'' portion, such that the vector L2 norm grows proportional to the size of an organization, and the ``falling'' portion, where the L2 norm begins to decline for larger organizations at a certain critical size. 

We propose that the ``rising'' portion of the curve is directly related to the frequency with which an organization appears in the trajectory data. 
This phenomenon results from the \textit{word2vec} training procedure. 
Before training begins, terms in the vocabulary (here, organization identifiers) are assigned random embedding vectors near the origin of the embedding space.
During training, these vectors are iteratively updated based on information from each trajectory.
Organizations that appear more often will be updated more frequently, stretching their embedding vector and thus their L2 norm. 
Organizations that appear rarely, however, will rarely be updated, and by the end of training will remain where they were instantiated near the origin of the embedding space, with short vectors.
Thus, barring other circumstances, we should expect the size of an organization's L2 norm to correlate with their frequency in the trajectories.

We support this argument with simulations based on synthetic trajectories which vary the prevalence of rare institutions.
We start with the co-occurence network used in our manuscript, where nodes represent organizations and edges are weighted by the frequency of co-occurence of two organizations in the same trajectory. 
Synthetic trajectories are generated from this network by initializing random walkers at a node, and moving five steps, the average length in the real trajectory data, with each edge chosen with a probability proportional to its weight relative to all options. 
We generate two sets of synthetic data using this process, with the only difference being the initial placement of random walkers.
The first set is generated by aligning the initial placement of walkers with that of individuals in the original data.
The second set of synthetic data is generated by initializing random walkers as uniformly random across all organizations.

We anticipate that synthetic trajectories following the \textit{first} process (random walkers initialized at each node proportional to their occurence in real trajectories) should produce the inverted ``U'' curvilinear pattern that we observe in our manuscript.
Indeed, this is what we observe (Fig.~\ref{fig:supp:real_network}a), implying that the process for generating synthetic data is producing trajectories that are an effective stand-in for real-world trajectory data. 
If our proposed mechanism for the rising trend is true, then we would expect the second set of synthetic trajectories (in which random walkers initialized at each node with uniform probability) to produce far more uniform L2 norms across organizations. 
Indeed, this is what we observe (Fig.~\ref{fig:supp:real_network}b, note that the x-axis corresponds to size in the real data, \textbf{\textit{not}} the synthetic data).
Still, though, in Fig.~\ref{fig:supp:real_network}b we observe a ``falling'' trend among larger institutions, which the existing mechanism does not account for, which we explore next. 

Next we examine the ``falling'' portion of the curve.
Previous research in natural language embedding [48] posits that groups of words confined to specific contexts often exhibit longer vectors, due to their continuously being updated in the same direction.
Conversely, when a word is pervasive and appears frequently across diverse contexts, its vector norm is smaller, because each update pulls the vector in a different direction, leaving it near the origin of the embedding space. 
This phenomenon has also been observed in the field of chemistry [49] and has been recently explored in studies concerning the concept of specificity using L2-norm [50-52].
Alternative explanations are that the falling portion is a result of distortion caused by representing hierarchical structure in an Euclidean space, or that it is the result of some other process not outlined in previous literature. 

We conduct a series of simulations in an attempt to obtain evidence in support or against these mechanisms. 
Synthetic trajectories are generated and embedded for a series of artificial networks, each network generated in an attempt to highlight a particular structural characteristic. 
The modeled networks are undirected, unweighted, and all consist of the same number of nodes as in the real co-occurence network of organizations used in our manuscript. 
Trajectories are generated on the network following the same random walker process as before, with starting positions sampled with a probability proportional to the node's degree. 
Below, we outline each generated network,

\begin{itemize}
  \item Configuration model [53]: Random graph with the given degree sequence. We generate a configuration model network from the degree sequence of the mobility network. We define the size of each node as node degree.
  \item Regular tree: Regular tree is a representative model network with a strict hierarchical structure, characterized by $m$, the number of child nodes for each node, and $d$, the height of the tree. We generated the network with $m=3$ and $d=8$ consisting of $n=9841$ nodes.
  Unlike the configuration model, the degree distribution of the tree network is nearly regular, since every node except the leaves and root have a comparable number of edges.
  In order to maintain the heterogeneity in the node sizes (i.e., number of scientists/authors in the nodes), we assign individual nodes the node sizes randomly sampled from the node size distribution for the original network. 
  \item  Barabási-Albert Model [54]: Model network with heterogeneous degree distributions using a preferential attachment mechanism. We set $m=3$, the number of edges to attach from a new node to existing nodes. We define the size of each node as node degree.

  \item Perfectly nested network [55]: Simple nested model network without self-loop whose adjacency matrix is 
\begin{equation}
	\label{eq:adj_nestdness}
	  A = \begin{array}{l}\begin{bmatrix} 0 & 1 & 1 & ….& 1 \\ 1 & 0 & 1 & …. & 0 \\ . & . & . & …. & 0 \\ 1 & 1 & 0 & …. & 0 \\ 1 & 0 & 0 & …. & 0 \end{bmatrix}\end{array}.
\end{equation}
  In this network, a node with the largest degree connects to every node in the network, while a node with the smallest network only has a single connection. We define the size of each node as node degree.
\end{itemize}

The results derived from the model network are shown in Fig.S32. Remarkably similar concave shapes also manifest in the configuration model (Fig. S32a), the  Barabási-Albert model (Fig. S32c), and the perfectly nested network (Fig. S32d). This strongly implies that the heterogeneous connection plays a crucial role in shaping these intriguing concave forms. In contrast, the regular tree (Fig. S32b), which solely embodies hierarchy, fails to yield such concave shapes. This observation underscores that the concave pattern doesn't emerge solely from distortion caused by embedding hierarchy into Euclidean spaces.

%
\paragraph*{S17 Text}
\label{si:text:generalized_gravity_model}
{\bf Generalized gravity model.}

Generalized gravity models use external information about each location to inform mobility predictions. 
    Here, we test whether a generalized gravity model that includes information about shared city, region, nation, continent, and language alongside geographic distance offers better predictions than embedding distance. 
    We specify the generalized model as follows,
    \begin{equation}
      \log{\frac{T_{ij}}{M_iM_j}} =  \gamma \log{d_{ij}} +  \sum_{x} \alpha^{x}*{g_{ij}^x} +  \sum_{x} \beta^{x}*{l_{ij}^x} 
   \end{equation}
    where $g_{ij}^{x}, x \in [\text{city, region, nation, and continent}]$ is a dummy variable that is given a value of 1 when two organizations are located in the same geographic city, region, nation, or continent, respectively; 
    otherwise, dummy variable is zero.
    Similarly, $l_{ij}^x,  x \in [\text{language, language family}]$ is a dummy variable that is set to 1 if two organizations are in countries that share the same language or language family, respectively. 
    
    Better results are obtained using the generalized gravity model ($R^2=0.242$, $RSME=0.861$, and $CPC=0.383$) compared to standard gravity model using only geographical distance ($R^2=0.219$, $RSME=0.874$, and $CPC=0.373$).
    However, the gravity model using embedding distance results in the best performance by a wide margin ($R^2=0.482$, $RSME=0.711$, and $CPC=0.459$). 
    These findings demonstrate how the embedding space can encode the complex and multi-faceted nature of mobility into a single robust vector-space representation, even without any external information provided.

\newpage

\begin{table}[ht]
\centering
\caption{\textbf{Full organization names}}
\label{table:supp:orglabels}
\begin{adjustwidth}{0cm}{}
\begingroup\tiny
\begin{tabular}{llll}
  \hline
Short & Full & Short & Full \\
  \hline
Stanford & Stanford Univ & Northwestern & Northwestern Univ \\
  Columbia & Columbia Univ & Ball State & Ball State Univ \\
  Harvard & Harvard Univ & IU Bloomington & Indiana Univ, Bloomington \\
  UCLA & Univ of California, Los Angeles & Stevens Institute & Stevens Institute of Technology \\
  Cal State Long Beach & California State Univ, Long Beach & NJIT & New Jersey Institute of Technology \\
  Wright State & Wright State Univ & NYU & New York Univ \\
  U Toledo & Univ of Toledo & SUNY Albany & Univ at Albany, The State Univ of New York \\
  Boston U & Boston Univ & NY Medical College & New York Medical College \\
  Suffolk & Suffolk Univ & Miami University & Miami Univ \\
  CUNY & City Univ of New York (CUNY) & IU Pennsylvania & Indiana Univ of Pennsylvania \\
  U Arizona & Univ of Arizona & Baylor & Baylor College of Medicine \\
  OSU & Ohio State Univ & UT Health Center & Univ of Texas Health Science Center \\
  MIT & Massachusetts Institute of Technology & Bard College & Bard College \\
  Princeton & Princeton Univ & Stonehill College & Stonehill College \\
  GCU & Grand Canyon Univ & Carleton College & Carleton College \\
  Northcentral & Northcentral Univ & Hanover College & Hanover College \\
  UCSF & Univ of California, San Francisco & Queens College & Queens College \\
  Fielding & Fielding Graduate Univ & DePauw & DePauw College \\
  Pepperdine & Pepperdine Univ & Naval Academy & United States Naval Academy \\
  Argosy & Argosy Univ & Cal State San Marcos & California State Univ San Marcos \\
  Yale & Yale Univ & Broad Inst & Broad Institute \\
  U Hartford & Univ of Hartford & Forsyth Inst & Forsyth Institute \\
  FAU & Florida Atlantic Univ & U Alaska Museum & Univ of Alaska Museum of the North \\
  U Miami & Univ of Miami & Lawrence Berkeley & Lawrence Berkeley Natl Laboratory \\
  UWF & The Univ of West Florida & Allen Institute & Allen Institute for Brain Science \\
  FIT & Florida Institute of Technology & RTI International & RTI InterNatl \\
  Purdue & Purdue Univ, West Lafayette & Fermilab & Fermilab \\
  Notre Dame & Univ of Notre Dame & State of NY & State of New York \\
  Indiana State & Indiana State Univ & Mayo Clinic & Mayo Clinic \\
  Saint Mary's & Saint Mary's College & Fish and Wildlife & Fish and Wildlife Research Institute \\
  Tufts & Tufts Univ & EPA & United States Environmental Protection Agency \\
  Mattel & Mattel Children's Hospital & US Army & United States Army \\
  Clark & Clark Univ & NSF & Natl Science Foundation \\
  UMass Amherst & Univ of Massachusetts Amherst & US Navy & United States Navy \\
  Montclair & Montclair State Univ & US Air Force & United States Air Force \\
  Farleigh Dickinson & Fairleigh Dickinson Univ-Metro Campus & Ames Laboratory & Ames Laboratory \\
  Rockefeller & Rockefeller Univ & Olin College & Oin College of Engineering \\
  Adelphi & Adelphi Univ & Scrips Institute & Scrips Institute \\
  Barnard & Barnard College & Idaho Natl Lab & Idaho Natl Laboratory \\
  Saint John Fisher & Saint John Fisher College & Dana Faber & Dana Faber Cancer Institute \\
  U Penn & Univ of Pennsylvania & Dept of Agriculture & United States Department of Agriculture \\
  Villanova & Villanova Univ & DOE & United States Department of Energy \\
  Widener & Widener Univ-Main Campus & NIAMS & Natl Institute of Arthritis, Skin Diseases \\
  Robert Morris & Robert Morris Univ & JMI Labs & JMI Laboratories \\
  U Cincinnati & Univ of Cincinnati & Whitehead Inst & Whitehead Institute of Biomedical Research \\
  Case Western & Case Western Reserve Univ & Wellesley & Wellesley Univ\\
  Ashland & Ashland Univ & UT Health, San Antonio & Univ of Texas Health Science Center, San Antonio \\
  Texas A\&M & Texas A\&M Univ-Commerce & UNT & Univ of North Texas \\
  Texas Southern & Texas Southern Univ & UT Southwestern Med & Univ of Texas Southwestern Medical Center \\
  Baylor & Univ of Mary Hardin-Baylor & UT El Paso & Univ of Texas, El Paso \\
  U Washington & Univ of Washington - Seattle & USF & Univ of South Florida, Tampa \\
  Washington State & Washington State Univ & Florida A\&M & Florida Agricultural and Mechanical Univ \\
  Seattle Pacific & Seattle Pacific Univ & Barry & Barry Univ \\
  Cal State Fresno & California State Univ-Fresno & UMass Dartmouth & Univ of Massachusetts Dartmouth \\
  Northern Arizona & Northern Arizona Univ & Worcester Poly & Worcester Polytechnic Institute \\
  IUPUI & Indiana Univ - Purdue Univ Indianapolis & Umass Boston & Univ of Massachusetts Boston \\
  U Dayton & Univ of Dayton & MGH Inst & MGH Institute of Health Professions \\
  U Conn & Univ of Connecticut & Joseph W. Jones Center & Joseph W. Jones Ecological Research Center \\
  ASU & Arizona State Univ & Vaccine Research Center & Vaccine Research Center, San Diego \\
  U Florida & Univ of Florida & LA Ag Center & Lousianna Agricultural Center \\
  Northern Illinois & Northern Illinois Univ & FL Fish and Wildlife & Florida Fish and Wildlife Conservation Commission \\
  Concordia Chicago & Concordia Univ-Chicago & NHLBI & Natl Heart, Lung, and Blood Institute \\
  U Chicago & Univ of Chicago & NY Dept. of Health & New York Department of Health \\
  SIU Edwardsville & Southern Illinois Univ, Edwardsville & St Michaels & Saint Michaels College \\
  SIU Carbondale & Southern Illinois Univ, Carbondale &  &  \\
   \hline
\end{tabular}
\endgroup
\end{adjustwidth}
\end{table}

\newpage

\begin{table}[]
\centering
\caption{\textbf{L2 Norm of country's representative vectors}.
			 Shown for top 30 countries with the most unique mobile and non-mobile researchers}
\label{table:supp:norm}
\begin{tabular}{lrr}
\textbf{Country} & \textbf{L2 Norm} & \textbf{\# Organizations} \\ \hline
United States    & 2.39             & 1281                      \\
Germany          & 2.6              & 485                       \\
United Kingdom   & 2.61             & 514                       \\
Austria          & 2.64             & 74                        \\
France           & 2.83             & 688                       \\
Belgium          & 2.84             & 84                        \\
Switzerland      & 2.85             & 66                        \\
Spain            & 2.94             & 322                       \\
China            & 2.97             & 497                       \\
India            & 2.99             & 114                       \\
Poland           & 3.02             & 145                       \\
Canada           & 3.02             & 147                       \\
Italy            & 3.04             & 386                       \\
Russia           & 3.08             & 187                       \\
Norway           & 3.1              & 122                       \\
Netherlands      & 3.11             & 136                       \\
Sweden           & 3.16             & 75                        \\
Brazil           & 3.16             & 286                       \\
Finland          & 3.17             & 66                        \\
Denmark          & 3.21             & 54                        \\
Czech Republic   & 3.23             & 97                        \\
Greece           & 3.24             & 62                        \\
Australia        & 3.24             & 90                        \\
Turkey           & 3.28             & 99                        \\
South Korea      & 3.28             & 156                       \\
Israel           & 3.32             & 71                        \\
Portugal         & 3.33             & 57                        \\
Japan            & 3.35             & 465                       \\
Iran             & 3.57             & 68                        \\
Taiwan           & 3.67             & 72
\end{tabular}
\end{table}

%
%
\clearpage
\begin{landscape}

  \providecommand{\huxb}[2]{\arrayrulecolor[RGB]{#1}\global\arrayrulewidth=#2pt}
  \providecommand{\huxvb}[2]{\color[RGB]{#1}\vrule width #2pt}
  \providecommand{\huxtpad}[1]{\rule{0pt}{#1}}
  \providecommand{\huxbpad}[1]{\rule[-#1]{0pt}{#1}}

\begin{table}[ht]
\caption{
	\textbf{Correlation between flux and distance over metrics, experimental parameters}.
			Each cell corresponds to the correlation between the real-world flux between scientific organizations (measured with $R^{2}$) and baseline metrics, shown by subsets of mobility data and by definitions of organization population.
			The asterisk denotes the top-performing distance metric by column.
			Distance metrics are ordered from highest $R^{2}$ to lowest, based on global mobility with organization population defined using all mobile and non-mobile authors.
			``All'' means that population is defined as the average yearly number of unique mobile and non-mobile scholars who published with the organizations' affiliation;
			population is defined in the same way for ``Mobile only'', except only using unique mobile researchers;
			``Raw freq'' means that organization populations are defined as their frequency across all the trajectories, similar to word frequency in language embedding.
			Embedding distance, measured as the cosine distance between embedding vectors, explains more of the flux than baselines in nearly every case, except using raw frequency population and domestic and international mobility, where direct optimization of the gravity model works better, as well as Levy's factorization~\autocite{levy2014neural} for domestic and international only mobility.
}
\label{supp:table:r2_table}
\begin{centerbox}
\begin{threeparttable}
\scriptsize
 \setlength{\tabcolsep}{-1pt}
\begin{tabular}{l l l l l l l l l l}

\multicolumn{1}{!{\huxvb{0, 0, 0}{0}}c!{\huxvb{0, 0, 0}{0.5}}}{\huxtpad{0pt + 1em}\centering \hspace{6pt} \textbf{} \hspace{6pt}\huxbpad{0pt}} &
\multicolumn{3}{c!{\huxvb{0, 0, 0}{0.5}}}{\huxtpad{0pt + 1em}\centering \hspace{6pt} \textbf{All} \hspace{6pt}\huxbpad{0pt}} &
\multicolumn{3}{c!{\huxvb{0, 0, 0}{0.5}}}{\huxtpad{0pt + 1em}\centering \hspace{6pt} \textbf{Mobile only} \hspace{6pt}\huxbpad{0pt}} &
\multicolumn{3}{c!{\huxvb{0, 0, 0}{0}}}{\huxtpad{0pt + 1em}\centering \hspace{6pt} \textbf{Raw freq} \hspace{6pt}\huxbpad{0pt}} \tabularnewline[-0.5pt]

\hhline{>{\huxb{0, 0, 0}{0.5}}|>{\huxb{0, 0, 0}{0.5}}|>{\huxb{0, 0, 0}{0.5}}|}
\arrayrulecolor{black}

\multicolumn{1}{!{\huxvb{0, 0, 0}{0}}l!{\huxvb{0, 0, 0}{0.5}}}{\huxtpad{6pt + 1em}\raggedright \hspace{6pt} \textbf{ } \hspace{6pt}\huxbpad{6pt}} &
\multicolumn{1}{r!{\huxvb{0, 0, 0}{0}}}{\huxtpad{6pt + 1em}\raggedleft \hspace{6pt} \textbf{All} \hspace{6pt}\huxbpad{6pt}} &
\multicolumn{1}{r!{\huxvb{0, 0, 0}{0}}}{\huxtpad{6pt + 1em}\raggedleft \hspace{6pt} \textbf{Domestic} \hspace{6pt}\huxbpad{6pt}} &
\multicolumn{1}{r!{\huxvb{0, 0, 0}{0.5}}}{\huxtpad{6pt + 1em}\raggedleft \hspace{6pt} \textbf{Inter.} \hspace{6pt}\huxbpad{6pt}} &
\multicolumn{1}{r!{\huxvb{0, 0, 0}{0}}}{\huxtpad{6pt + 1em}\raggedleft \hspace{6pt} \textbf{All} \hspace{6pt}\huxbpad{6pt}} &
\multicolumn{1}{r!{\huxvb{0, 0, 0}{0}}}{\huxtpad{6pt + 1em}\raggedleft \hspace{6pt} \textbf{Domestic} \hspace{6pt}\huxbpad{6pt}} &
\multicolumn{1}{r!{\huxvb{0, 0, 0}{0.5}}}{\huxtpad{6pt + 1em}\raggedleft \hspace{6pt} \textbf{Inter.} \hspace{6pt}\huxbpad{6pt}} &
\multicolumn{1}{r!{\huxvb{0, 0, 0}{0}}}{\huxtpad{6pt + 1em}\raggedleft \hspace{6pt} \textbf{All} \hspace{6pt}\huxbpad{6pt}} &
\multicolumn{1}{r!{\huxvb{0, 0, 0}{0}}}{\huxtpad{6pt + 1em}\raggedleft \hspace{6pt} \textbf{Domestic} \hspace{6pt}\huxbpad{6pt}} &
\multicolumn{1}{r!{\huxvb{0, 0, 0}{0}}}{\huxtpad{6pt + 1em}\raggedleft \hspace{6pt} \textbf{Inter.} \hspace{6pt}\huxbpad{6pt}} \tabularnewline[-0.5pt]

\hhline{>{\huxb{0, 0, 0}{0.5}}->{\huxb{0, 0, 0}{0.5}}->{\huxb{0, 0, 0}{0.5}}->{\huxb{0, 0, 0}{0.5}}->{\huxb{0, 0, 0}{0.5}}->{\huxb{0, 0, 0}{0.5}}->{\huxb{0, 0, 0}{0.5}}->{\huxb{0, 0, 0}{0.5}}->{\huxb{0, 0, 0}{0.5}}->{\huxb{0, 0, 0}{0.5}}-}
\arrayrulecolor{black}

\multicolumn{1}{!{\huxvb{0, 0, 0}{0}}l!{\huxvb{0, 0, 0}{0.5}}}{\huxtpad{6pt + 1em}\raggedright \hspace{6pt} \textbf{Embedding cosine} \hspace{6pt}\huxbpad{6pt}} &
\multicolumn{1}{r!{\huxvb{0, 0, 0}{0}}}{\huxtpad{6pt + 1em}\raggedleft \hspace{6pt} *0.481 \hspace{6pt}\huxbpad{6pt}} &
\multicolumn{1}{r!{\huxvb{0, 0, 0}{0}}}{\huxtpad{6pt + 1em}\raggedleft \hspace{6pt} *0.418 \hspace{6pt}\huxbpad{6pt}} &
\multicolumn{1}{r!{\huxvb{0, 0, 0}{0.5}}}{\huxtpad{6pt + 1em}\raggedleft \hspace{6pt} *0.435 \hspace{6pt}\huxbpad{6pt}} &
\multicolumn{1}{r!{\huxvb{0, 0, 0}{0}}}{\huxtpad{6pt + 1em}\raggedleft \hspace{6pt} *0.492 \hspace{6pt}\huxbpad{6pt}} &
\multicolumn{1}{r!{\huxvb{0, 0, 0}{0}}}{\huxtpad{6pt + 1em}\raggedleft \hspace{6pt} *0.456 \hspace{6pt}\huxbpad{6pt}} &
\multicolumn{1}{r!{\huxvb{0, 0, 0}{0.5}}}{\huxtpad{6pt + 1em}\raggedleft \hspace{6pt} *0.489 \hspace{6pt}\huxbpad{6pt}} &
\multicolumn{1}{r!{\huxvb{0, 0, 0}{0}}}{\huxtpad{6pt + 1em}\raggedleft \hspace{6pt} 0.325 \hspace{6pt}\huxbpad{6pt}} &
\multicolumn{1}{r!{\huxvb{0, 0, 0}{0}}}{\huxtpad{6pt + 1em}\raggedleft \hspace{6pt} 0.251 \hspace{6pt}\huxbpad{6pt}} &
\multicolumn{1}{r!{\huxvb{0, 0, 0}{0}}}{\huxtpad{6pt + 1em}\raggedleft \hspace{6pt} 0.252 \hspace{6pt}\huxbpad{6pt}} \tabularnewline[-0.5pt]

\hhline{>{\huxb{0, 0, 0}{0.5}}|>{\huxb{0, 0, 0}{0.5}}|>{\huxb{0, 0, 0}{0.5}}|}
\arrayrulecolor{black}

\multicolumn{1}{!{\huxvb{0, 0, 0}{0}}l!{\huxvb{0, 0, 0}{0.5}}}{\huxtpad{6pt + 1em}\raggedright \hspace{6pt} \textbf{Gravity MDS Euclidean} \hspace{6pt}\huxbpad{6pt}} &
\multicolumn{1}{r!{\huxvb{0, 0, 0}{0}}}{\huxtpad{6pt + 1em}\raggedleft \hspace{6pt} 0.355 \hspace{6pt}\huxbpad{6pt}} &
\multicolumn{1}{r!{\huxvb{0, 0, 0}{0}}}{\huxtpad{6pt + 1em}\raggedleft \hspace{6pt} 0.165 \hspace{6pt}\huxbpad{6pt}} &
\multicolumn{1}{r!{\huxvb{0, 0, 0}{0.5}}}{\huxtpad{6pt + 1em}\raggedleft \hspace{6pt} 0.161 \hspace{6pt}\huxbpad{6pt}} &
\multicolumn{1}{r!{\huxvb{0, 0, 0}{0}}}{\huxtpad{6pt + 1em}\raggedleft \hspace{6pt} 0.328 \hspace{6pt}\huxbpad{6pt}} &
\multicolumn{1}{r!{\huxvb{0, 0, 0}{0}}}{\huxtpad{6pt + 1em}\raggedleft \hspace{6pt} 0.112 \hspace{6pt}\huxbpad{6pt}} &
\multicolumn{1}{r!{\huxvb{0, 0, 0}{0.5}}}{\huxtpad{6pt + 1em}\raggedleft \hspace{6pt} 0.115 \hspace{6pt}\huxbpad{6pt}} &
\multicolumn{1}{r!{\huxvb{0, 0, 0}{0}}}{\huxtpad{6pt + 1em}\raggedleft \hspace{6pt} *0.369 \hspace{6pt}\huxbpad{6pt}} &
\multicolumn{1}{r!{\huxvb{0, 0, 0}{0}}}{\huxtpad{6pt + 1em}\raggedleft \hspace{6pt} 0.174 \hspace{6pt}\huxbpad{6pt}} &
\multicolumn{1}{r!{\huxvb{0, 0, 0}{0}}}{\huxtpad{6pt + 1em}\raggedleft \hspace{6pt} 0.164 \hspace{6pt}\huxbpad{6pt}} \tabularnewline[-0.5pt]

\hhline{>{\huxb{0, 0, 0}{0.5}}|>{\huxb{0, 0, 0}{0.5}}|>{\huxb{0, 0, 0}{0.5}}|}
\arrayrulecolor{black}

\multicolumn{1}{!{\huxvb{0, 0, 0}{0}}l!{\huxvb{0, 0, 0}{0.5}}}{\huxtpad{6pt + 1em}\raggedright \hspace{6pt} \textbf{Levy's Euclidean} \hspace{6pt}\huxbpad{6pt}} &
\multicolumn{1}{r!{\huxvb{0, 0, 0}{0}}}{\huxtpad{6pt + 1em}\raggedleft \hspace{6pt} 0.341 \hspace{6pt}\huxbpad{6pt}} &
\multicolumn{1}{r!{\huxvb{0, 0, 0}{0}}}{\huxtpad{6pt + 1em}\raggedleft \hspace{6pt} 0.369 \hspace{6pt}\huxbpad{6pt}} &
\multicolumn{1}{r!{\huxvb{0, 0, 0}{0.5}}}{\huxtpad{6pt + 1em}\raggedleft \hspace{6pt} 0.382 \hspace{6pt}\huxbpad{6pt}} &
\multicolumn{1}{r!{\huxvb{0, 0, 0}{0}}}{\huxtpad{6pt + 1em}\raggedleft \hspace{6pt} 0.213 \hspace{6pt}\huxbpad{6pt}} &
\multicolumn{1}{r!{\huxvb{0, 0, 0}{0}}}{\huxtpad{6pt + 1em}\raggedleft \hspace{6pt} 0.271 \hspace{6pt}\huxbpad{6pt}} &
\multicolumn{1}{r!{\huxvb{0, 0, 0}{0.5}}}{\huxtpad{6pt + 1em}\raggedleft \hspace{6pt} 0.284 \hspace{6pt}\huxbpad{6pt}} &
\multicolumn{1}{r!{\huxvb{0, 0, 0}{0}}}{\huxtpad{6pt + 1em}\raggedleft \hspace{6pt} 0.305 \hspace{6pt}\huxbpad{6pt}} &
\multicolumn{1}{r!{\huxvb{0, 0, 0}{0}}}{\huxtpad{6pt + 1em}\raggedleft \hspace{6pt} *0.323 \hspace{6pt}\huxbpad{6pt}} &
\multicolumn{1}{r!{\huxvb{0, 0, 0}{0}}}{\huxtpad{6pt + 1em}\raggedleft \hspace{6pt} *0.334 \hspace{6pt}\huxbpad{6pt}} \tabularnewline[-0.5pt]

\hhline{>{\huxb{0, 0, 0}{0.5}}|>{\huxb{0, 0, 0}{0.5}}|>{\huxb{0, 0, 0}{0.5}}|}
\arrayrulecolor{black}

\multicolumn{1}{!{\huxvb{0, 0, 0}{0}}l!{\huxvb{0, 0, 0}{0.5}}}{\huxtpad{6pt + 1em}\raggedright \hspace{6pt} \textbf{Embedding dot} \hspace{6pt}\huxbpad{6pt}} &
\multicolumn{1}{r!{\huxvb{0, 0, 0}{0}}}{\huxtpad{6pt + 1em}\raggedleft \hspace{6pt} 0.341 \hspace{6pt}\huxbpad{6pt}} &
\multicolumn{1}{r!{\huxvb{0, 0, 0}{0}}}{\huxtpad{6pt + 1em}\raggedleft \hspace{6pt} 0.313 \hspace{6pt}\huxbpad{6pt}} &
\multicolumn{1}{r!{\huxvb{0, 0, 0}{0.5}}}{\huxtpad{6pt + 1em}\raggedleft \hspace{6pt} 0.316 \hspace{6pt}\huxbpad{6pt}} &
\multicolumn{1}{r!{\huxvb{0, 0, 0}{0}}}{\huxtpad{6pt + 1em}\raggedleft \hspace{6pt} 0.254 \hspace{6pt}\huxbpad{6pt}} &
\multicolumn{1}{r!{\huxvb{0, 0, 0}{0}}}{\huxtpad{6pt + 1em}\raggedleft \hspace{6pt} 0.265 \hspace{6pt}\huxbpad{6pt}} &
\multicolumn{1}{r!{\huxvb{0, 0, 0}{0.5}}}{\huxtpad{6pt + 1em}\raggedleft \hspace{6pt} 0.267 \hspace{6pt}\huxbpad{6pt}} &
\multicolumn{1}{r!{\huxvb{0, 0, 0}{0}}}{\huxtpad{6pt + 1em}\raggedleft \hspace{6pt} 0.218 \hspace{6pt}\huxbpad{6pt}} &
\multicolumn{1}{r!{\huxvb{0, 0, 0}{0}}}{\huxtpad{6pt + 1em}\raggedleft \hspace{6pt} 0.181 \hspace{6pt}\huxbpad{6pt}} &
\multicolumn{1}{r!{\huxvb{0, 0, 0}{0}}}{\huxtpad{6pt + 1em}\raggedleft \hspace{6pt} 0.177 \hspace{6pt}\huxbpad{6pt}} \tabularnewline[-0.5pt]

\hhline{>{\huxb{0, 0, 0}{0.5}}|>{\huxb{0, 0, 0}{0.5}}|>{\huxb{0, 0, 0}{0.5}}|}
\arrayrulecolor{black}

\multicolumn{1}{!{\huxvb{0, 0, 0}{0}}l!{\huxvb{0, 0, 0}{0.5}}}{\huxtpad{6pt + 1em}\raggedright \hspace{6pt} \textbf{SVD cosine} \hspace{6pt}\huxbpad{6pt}} &
\multicolumn{1}{r!{\huxvb{0, 0, 0}{0}}}{\huxtpad{6pt + 1em}\raggedleft \hspace{6pt} 0.247 \hspace{6pt}\huxbpad{6pt}} &
\multicolumn{1}{r!{\huxvb{0, 0, 0}{0}}}{\huxtpad{6pt + 1em}\raggedleft \hspace{6pt} 0.297 \hspace{6pt}\huxbpad{6pt}} &
\multicolumn{1}{r!{\huxvb{0, 0, 0}{0.5}}}{\huxtpad{6pt + 1em}\raggedleft \hspace{6pt} 0.309 \hspace{6pt}\huxbpad{6pt}} &
\multicolumn{1}{r!{\huxvb{0, 0, 0}{0}}}{\huxtpad{6pt + 1em}\raggedleft \hspace{6pt} 0.213 \hspace{6pt}\huxbpad{6pt}} &
\multicolumn{1}{r!{\huxvb{0, 0, 0}{0}}}{\huxtpad{6pt + 1em}\raggedleft \hspace{6pt} 0.314 \hspace{6pt}\huxbpad{6pt}} &
\multicolumn{1}{r!{\huxvb{0, 0, 0}{0.5}}}{\huxtpad{6pt + 1em}\raggedleft \hspace{6pt} 0.325 \hspace{6pt}\huxbpad{6pt}} &
\multicolumn{1}{r!{\huxvb{0, 0, 0}{0}}}{\huxtpad{6pt + 1em}\raggedleft \hspace{6pt} 0.111 \hspace{6pt}\huxbpad{6pt}} &
\multicolumn{1}{r!{\huxvb{0, 0, 0}{0}}}{\huxtpad{6pt + 1em}\raggedleft \hspace{6pt} 0.152 \hspace{6pt}\huxbpad{6pt}} &
\multicolumn{1}{r!{\huxvb{0, 0, 0}{0}}}{\huxtpad{6pt + 1em}\raggedleft \hspace{6pt} 0.16 \hspace{6pt}\huxbpad{6pt}} \tabularnewline[-0.5pt]

\hhline{>{\huxb{0, 0, 0}{0.5}}|>{\huxb{0, 0, 0}{0.5}}|>{\huxb{0, 0, 0}{0.5}}|}
\arrayrulecolor{black}

\multicolumn{1}{!{\huxvb{0, 0, 0}{0}}l!{\huxvb{0, 0, 0}{0.5}}}{\huxtpad{6pt + 1em}\raggedright \hspace{6pt} \textbf{Geographic} \hspace{6pt}\huxbpad{6pt}} &
\multicolumn{1}{r!{\huxvb{0, 0, 0}{0}}}{\huxtpad{6pt + 1em}\raggedleft \hspace{6pt} 0.219 \hspace{6pt}\huxbpad{6pt}} &
\multicolumn{1}{r!{\huxvb{0, 0, 0}{0}}}{\huxtpad{6pt + 1em}\raggedleft \hspace{6pt} 0.174 \hspace{6pt}\huxbpad{6pt}} &
\multicolumn{1}{r!{\huxvb{0, 0, 0}{0.5}}}{\huxtpad{6pt + 1em}\raggedleft \hspace{6pt} 0.197 \hspace{6pt}\huxbpad{6pt}} &
\multicolumn{1}{r!{\huxvb{0, 0, 0}{0}}}{\huxtpad{6pt + 1em}\raggedleft \hspace{6pt} 0.188 \hspace{6pt}\huxbpad{6pt}} &
\multicolumn{1}{r!{\huxvb{0, 0, 0}{0}}}{\huxtpad{6pt + 1em}\raggedleft \hspace{6pt} 0.157 \hspace{6pt}\huxbpad{6pt}} &
\multicolumn{1}{r!{\huxvb{0, 0, 0}{0.5}}}{\huxtpad{6pt + 1em}\raggedleft \hspace{6pt} 0.176 \hspace{6pt}\huxbpad{6pt}} &
\multicolumn{1}{r!{\huxvb{0, 0, 0}{0}}}{\huxtpad{6pt + 1em}\raggedleft \hspace{6pt} 0.04 \hspace{6pt}\huxbpad{6pt}} &
\multicolumn{1}{r!{\huxvb{0, 0, 0}{0}}}{\huxtpad{6pt + 1em}\raggedleft \hspace{6pt} 0.019 \hspace{6pt}\huxbpad{6pt}} &
\multicolumn{1}{r!{\huxvb{0, 0, 0}{0}}}{\huxtpad{6pt + 1em}\raggedleft \hspace{6pt} 0.03 \hspace{6pt}\huxbpad{6pt}} \tabularnewline[-0.5pt]

\hhline{>{\huxb{0, 0, 0}{0.5}}|>{\huxb{0, 0, 0}{0.5}}|>{\huxb{0, 0, 0}{0.5}}|}
\arrayrulecolor{black}

\multicolumn{1}{!{\huxvb{0, 0, 0}{0}}l!{\huxvb{0, 0, 0}{0.5}}}{\huxtpad{6pt + 1em}\raggedright \hspace{6pt} \textbf{Laplacian cosine} \hspace{6pt}\huxbpad{6pt}} &
\multicolumn{1}{r!{\huxvb{0, 0, 0}{0}}}{\huxtpad{6pt + 1em}\raggedleft \hspace{6pt} 0.212 \hspace{6pt}\huxbpad{6pt}} &
\multicolumn{1}{r!{\huxvb{0, 0, 0}{0}}}{\huxtpad{6pt + 1em}\raggedleft \hspace{6pt} 0.199 \hspace{6pt}\huxbpad{6pt}} &
\multicolumn{1}{r!{\huxvb{0, 0, 0}{0.5}}}{\huxtpad{6pt + 1em}\raggedleft \hspace{6pt} 0.218 \hspace{6pt}\huxbpad{6pt}} &
\multicolumn{1}{r!{\huxvb{0, 0, 0}{0}}}{\huxtpad{6pt + 1em}\raggedleft \hspace{6pt} 0.176 \hspace{6pt}\huxbpad{6pt}} &
\multicolumn{1}{r!{\huxvb{0, 0, 0}{0}}}{\huxtpad{6pt + 1em}\raggedleft \hspace{6pt} 0.157 \hspace{6pt}\huxbpad{6pt}} &
\multicolumn{1}{r!{\huxvb{0, 0, 0}{0.5}}}{\huxtpad{6pt + 1em}\raggedleft \hspace{6pt} 0.18 \hspace{6pt}\huxbpad{6pt}} &
\multicolumn{1}{r!{\huxvb{0, 0, 0}{0}}}{\huxtpad{6pt + 1em}\raggedleft \hspace{6pt} 0.079 \hspace{6pt}\huxbpad{6pt}} &
\multicolumn{1}{r!{\huxvb{0, 0, 0}{0}}}{\huxtpad{6pt + 1em}\raggedleft \hspace{6pt} 0.1 \hspace{6pt}\huxbpad{6pt}} &
\multicolumn{1}{r!{\huxvb{0, 0, 0}{0}}}{\huxtpad{6pt + 1em}\raggedleft \hspace{6pt} 0.111 \hspace{6pt}\huxbpad{6pt}} \tabularnewline[-0.5pt]

\hhline{>{\huxb{0, 0, 0}{0.5}}|>{\huxb{0, 0, 0}{0.5}}|>{\huxb{0, 0, 0}{0.5}}|}
\arrayrulecolor{black}

\multicolumn{1}{!{\huxvb{0, 0, 0}{0}}l!{\huxvb{0, 0, 0}{0.5}}}{\huxtpad{6pt + 1em}\raggedright \hspace{6pt} \textbf{Levy's cosine} \hspace{6pt}\huxbpad{6pt}} &
\multicolumn{1}{r!{\huxvb{0, 0, 0}{0}}}{\huxtpad{6pt + 1em}\raggedleft \hspace{6pt} 0.208 \hspace{6pt}\huxbpad{6pt}} &
\multicolumn{1}{r!{\huxvb{0, 0, 0}{0}}}{\huxtpad{6pt + 1em}\raggedleft \hspace{6pt} 0.227 \hspace{6pt}\huxbpad{6pt}} &
\multicolumn{1}{r!{\huxvb{0, 0, 0}{0.5}}}{\huxtpad{6pt + 1em}\raggedleft \hspace{6pt} 0.231 \hspace{6pt}\huxbpad{6pt}} &
\multicolumn{1}{r!{\huxvb{0, 0, 0}{0}}}{\huxtpad{6pt + 1em}\raggedleft \hspace{6pt} 0.169 \hspace{6pt}\huxbpad{6pt}} &
\multicolumn{1}{r!{\huxvb{0, 0, 0}{0}}}{\huxtpad{6pt + 1em}\raggedleft \hspace{6pt} 0.246 \hspace{6pt}\huxbpad{6pt}} &
\multicolumn{1}{r!{\huxvb{0, 0, 0}{0.5}}}{\huxtpad{6pt + 1em}\raggedleft \hspace{6pt} 0.246 \hspace{6pt}\huxbpad{6pt}} &
\multicolumn{1}{r!{\huxvb{0, 0, 0}{0}}}{\huxtpad{6pt + 1em}\raggedleft \hspace{6pt} 0.057 \hspace{6pt}\huxbpad{6pt}} &
\multicolumn{1}{r!{\huxvb{0, 0, 0}{0}}}{\huxtpad{6pt + 1em}\raggedleft \hspace{6pt} 0.054 \hspace{6pt}\huxbpad{6pt}} &
\multicolumn{1}{r!{\huxvb{0, 0, 0}{0}}}{\huxtpad{6pt + 1em}\raggedleft \hspace{6pt} 0.053 \hspace{6pt}\huxbpad{6pt}} \tabularnewline[-0.5pt]

\hhline{>{\huxb{0, 0, 0}{0.5}}|>{\huxb{0, 0, 0}{0.5}}|>{\huxb{0, 0, 0}{0.5}}|}
\arrayrulecolor{black}

\multicolumn{1}{!{\huxvb{0, 0, 0}{0}}l!{\huxvb{0, 0, 0}{0.5}}}{\huxtpad{6pt + 1em}\raggedright \hspace{6pt} \textbf{PPR JSD} \hspace{6pt}\huxbpad{6pt}} &
\multicolumn{1}{r!{\huxvb{0, 0, 0}{0}}}{\huxtpad{6pt + 1em}\raggedleft \hspace{6pt} 0.194 \hspace{6pt}\huxbpad{6pt}} &
\multicolumn{1}{r!{\huxvb{0, 0, 0}{0}}}{\huxtpad{6pt + 1em}\raggedleft \hspace{6pt} 0.276 \hspace{6pt}\huxbpad{6pt}} &
\multicolumn{1}{r!{\huxvb{0, 0, 0}{0.5}}}{\huxtpad{6pt + 1em}\raggedleft \hspace{6pt} 0.276 \hspace{6pt}\huxbpad{6pt}} &
\multicolumn{1}{r!{\huxvb{0, 0, 0}{0}}}{\huxtpad{6pt + 1em}\raggedleft \hspace{6pt} 0.218 \hspace{6pt}\huxbpad{6pt}} &
\multicolumn{1}{r!{\huxvb{0, 0, 0}{0}}}{\huxtpad{6pt + 1em}\raggedleft \hspace{6pt} 0.335 \hspace{6pt}\huxbpad{6pt}} &
\multicolumn{1}{r!{\huxvb{0, 0, 0}{0.5}}}{\huxtpad{6pt + 1em}\raggedleft \hspace{6pt} 0.335 \hspace{6pt}\huxbpad{6pt}} &
\multicolumn{1}{r!{\huxvb{0, 0, 0}{0}}}{\huxtpad{6pt + 1em}\raggedleft \hspace{6pt} 0.012 \hspace{6pt}\huxbpad{6pt}} &
\multicolumn{1}{r!{\huxvb{0, 0, 0}{0}}}{\huxtpad{6pt + 1em}\raggedleft \hspace{6pt} 0.077 \hspace{6pt}\huxbpad{6pt}} &
\multicolumn{1}{r!{\huxvb{0, 0, 0}{0}}}{\huxtpad{6pt + 1em}\raggedleft \hspace{6pt} 0.069 \hspace{6pt}\huxbpad{6pt}} \tabularnewline[-0.5pt]

\hhline{>{\huxb{0, 0, 0}{0.5}}|>{\huxb{0, 0, 0}{0.5}}|>{\huxb{0, 0, 0}{0.5}}|}
\arrayrulecolor{black}

\multicolumn{1}{!{\huxvb{0, 0, 0}{0}}l!{\huxvb{0, 0, 0}{0.5}}}{\huxtpad{6pt + 1em}\raggedright \hspace{6pt} \textbf{PPR cosine} \hspace{6pt}\huxbpad{6pt}} &
\multicolumn{1}{r!{\huxvb{0, 0, 0}{0}}}{\huxtpad{6pt + 1em}\raggedleft \hspace{6pt} 0.138 \hspace{6pt}\huxbpad{6pt}} &
\multicolumn{1}{r!{\huxvb{0, 0, 0}{0}}}{\huxtpad{6pt + 1em}\raggedleft \hspace{6pt} 0.136 \hspace{6pt}\huxbpad{6pt}} &
\multicolumn{1}{r!{\huxvb{0, 0, 0}{0.5}}}{\huxtpad{6pt + 1em}\raggedleft \hspace{6pt} 0.143 \hspace{6pt}\huxbpad{6pt}} &
\multicolumn{1}{r!{\huxvb{0, 0, 0}{0}}}{\huxtpad{6pt + 1em}\raggedleft \hspace{6pt} 0.196 \hspace{6pt}\huxbpad{6pt}} &
\multicolumn{1}{r!{\huxvb{0, 0, 0}{0}}}{\huxtpad{6pt + 1em}\raggedleft \hspace{6pt} 0.186 \hspace{6pt}\huxbpad{6pt}} &
\multicolumn{1}{r!{\huxvb{0, 0, 0}{0.5}}}{\huxtpad{6pt + 1em}\raggedleft \hspace{6pt} 0.197 \hspace{6pt}\huxbpad{6pt}} &
\multicolumn{1}{r!{\huxvb{0, 0, 0}{0}}}{\huxtpad{6pt + 1em}\raggedleft \hspace{6pt} 0.13 \hspace{6pt}\huxbpad{6pt}} &
\multicolumn{1}{r!{\huxvb{0, 0, 0}{0}}}{\huxtpad{6pt + 1em}\raggedleft \hspace{6pt} 0.149 \hspace{6pt}\huxbpad{6pt}} &
\multicolumn{1}{r!{\huxvb{0, 0, 0}{0}}}{\huxtpad{6pt + 1em}\raggedleft \hspace{6pt} 0.152 \hspace{6pt}\huxbpad{6pt}} \tabularnewline[-0.5pt]

\hhline{>{\huxb{0, 0, 0}{0.5}}|>{\huxb{0, 0, 0}{0.5}}|>{\huxb{0, 0, 0}{0.5}}|}
\arrayrulecolor{black}

\multicolumn{1}{!{\huxvb{0, 0, 0}{0}}l!{\huxvb{0, 0, 0}{0.5}}}{\huxtpad{6pt + 1em}\raggedright \hspace{6pt} \textbf{Gravity SVD cosine} \hspace{6pt}\huxbpad{6pt}} &
\multicolumn{1}{r!{\huxvb{0, 0, 0}{0}}}{\huxtpad{6pt + 1em}\raggedleft \hspace{6pt} 0.122 \hspace{6pt}\huxbpad{6pt}} &
\multicolumn{1}{r!{\huxvb{0, 0, 0}{0}}}{\huxtpad{6pt + 1em}\raggedleft \hspace{6pt} 0.118 \hspace{6pt}\huxbpad{6pt}} &
\multicolumn{1}{r!{\huxvb{0, 0, 0}{0.5}}}{\huxtpad{6pt + 1em}\raggedleft \hspace{6pt} 0.122 \hspace{6pt}\huxbpad{6pt}} &
\multicolumn{1}{r!{\huxvb{0, 0, 0}{0}}}{\huxtpad{6pt + 1em}\raggedleft \hspace{6pt} 0.118 \hspace{6pt}\huxbpad{6pt}} &
\multicolumn{1}{r!{\huxvb{0, 0, 0}{0}}}{\huxtpad{6pt + 1em}\raggedleft \hspace{6pt} 0.133 \hspace{6pt}\huxbpad{6pt}} &
\multicolumn{1}{r!{\huxvb{0, 0, 0}{0.5}}}{\huxtpad{6pt + 1em}\raggedleft \hspace{6pt} 0.138 \hspace{6pt}\huxbpad{6pt}} &
\multicolumn{1}{r!{\huxvb{0, 0, 0}{0}}}{\huxtpad{6pt + 1em}\raggedleft \hspace{6pt} 0.056 \hspace{6pt}\huxbpad{6pt}} &
\multicolumn{1}{r!{\huxvb{0, 0, 0}{0}}}{\huxtpad{6pt + 1em}\raggedleft \hspace{6pt} 0.047 \hspace{6pt}\huxbpad{6pt}} &
\multicolumn{1}{r!{\huxvb{0, 0, 0}{0}}}{\huxtpad{6pt + 1em}\raggedleft \hspace{6pt} 0.05 \hspace{6pt}\huxbpad{6pt}} \tabularnewline[-0.5pt]

\hhline{>{\huxb{0, 0, 0}{0.5}}|>{\huxb{0, 0, 0}{0.5}}|>{\huxb{0, 0, 0}{0.5}}|}
\arrayrulecolor{black}

\multicolumn{1}{!{\huxvb{0, 0, 0}{0}}l!{\huxvb{0, 0, 0}{0.5}}}{\huxtpad{6pt + 1em}\raggedright \hspace{6pt} \textbf{Levy's dot} \hspace{6pt}\huxbpad{6pt}} &
\multicolumn{1}{r!{\huxvb{0, 0, 0}{0}}}{\huxtpad{6pt + 1em}\raggedleft \hspace{6pt} 0.004 \hspace{6pt}\huxbpad{6pt}} &
\multicolumn{1}{r!{\huxvb{0, 0, 0}{0}}}{\huxtpad{6pt + 1em}\raggedleft \hspace{6pt} 0.002 \hspace{6pt}\huxbpad{6pt}} &
\multicolumn{1}{r!{\huxvb{0, 0, 0}{0.5}}}{\huxtpad{6pt + 1em}\raggedleft \hspace{6pt} 0.002 \hspace{6pt}\huxbpad{6pt}} &
\multicolumn{1}{r!{\huxvb{0, 0, 0}{0}}}{\huxtpad{6pt + 1em}\raggedleft \hspace{6pt} 0.013 \hspace{6pt}\huxbpad{6pt}} &
\multicolumn{1}{r!{\huxvb{0, 0, 0}{0}}}{\huxtpad{6pt + 1em}\raggedleft \hspace{6pt} 0.002 \hspace{6pt}\huxbpad{6pt}} &
\multicolumn{1}{r!{\huxvb{0, 0, 0}{0.5}}}{\huxtpad{6pt + 1em}\raggedleft \hspace{6pt} 0.004 \hspace{6pt}\huxbpad{6pt}} &
\multicolumn{1}{r!{\huxvb{0, 0, 0}{0}}}{\huxtpad{6pt + 1em}\raggedleft \hspace{6pt} 0.039 \hspace{6pt}\huxbpad{6pt}} &
\multicolumn{1}{r!{\huxvb{0, 0, 0}{0}}}{\huxtpad{6pt + 1em}\raggedleft \hspace{6pt} 0.034 \hspace{6pt}\huxbpad{6pt}} &
\multicolumn{1}{r!{\huxvb{0, 0, 0}{0}}}{\huxtpad{6pt + 1em}\raggedleft \hspace{6pt} 0.035 \hspace{6pt}\huxbpad{6pt}} \tabularnewline[-0.5pt]

\hhline{>{\huxb{0, 0, 0}{0.5}}|>{\huxb{0, 0, 0}{0.5}}|>{\huxb{0, 0, 0}{0.5}}|}
\arrayrulecolor{black}
\end{tabular}
\end{threeparttable}\par\end{centerbox}

\end{table}

\end{landscape}

%
%
\clearpage
\begin{landscape}

  \providecommand{\huxb}[2]{\arrayrulecolor[RGB]{#1}\global\arrayrulewidth=#2pt}
  \providecommand{\huxvb}[2]{\color[RGB]{#1}\vrule width #2pt}
  \providecommand{\huxtpad}[1]{\rule{0pt}{#1}}
  \providecommand{\huxbpad}[1]{\rule[-#1]{0pt}{#1}}

\begin{table}[ht]
\caption{
	\textbf{Prediction error between actual and predicted mobility with exponential gravity model, by metrics and experimental parameters}.
	Each cell corresponds to the prediction error (measured with root mean squared error) when using each distance as input to the exponential form of the gravity model of mobility to predict the flux between organizations, shown by subsets of mobility data, and by definitions of organization population.
	The asterisk denotes the top-performing distance metric by column (lowest prediction error).
	Distance metrics are ordered from lowest prediction error to highest, based on global mobility with organization population defined using all mobile and non-mobile authors.
	``All'' means that population is defined as the average yearly number of unique mobile and non-mobile scholars who published with the organizations' affiliation;
		population is defined in the same way for ``Mobile only'', except only using unique mobile researchers;
		``Raw freq'' means that organization populations are defined as their frequency across all the trajectories, similar to word frequency in language embedding.
		Embedding distance, measured as the cosine distance between embedding vectors, results in better predictions of mobility than baselines in nearly every case, however Levy's 'factorization~\autocite{levy2014neural} perform better in the case of international and domestic only mobility when using raw frequency populations.
}
\label{supp:table:rmse_exp_table}
\begin{centerbox}
\begin{threeparttable}
\scriptsize
 \setlength{\tabcolsep}{0pt}
\begin{tabular}{l l l l l l l l l l}

\hhline{>{\huxb{0, 0, 0}{0.5}}|>{\huxb{0, 0, 0}{0.5}}|>{\huxb{0, 0, 0}{0.5}}|}
\arrayrulecolor{black}

\multicolumn{1}{!{\huxvb{0, 0, 0}{0}}c!{\huxvb{0, 0, 0}{0.5}}}{\huxtpad{0pt + 1em}\centering \hspace{6pt} \textbf{} \hspace{6pt}\huxbpad{0pt}} &
\multicolumn{3}{c!{\huxvb{0, 0, 0}{0.5}}}{\huxtpad{0pt + 1em}\centering \hspace{6pt} \textbf{All} \hspace{6pt}\huxbpad{0pt}} &
\multicolumn{3}{c!{\huxvb{0, 0, 0}{0.5}}}{\huxtpad{0pt + 1em}\centering \hspace{6pt} \textbf{Mobile only} \hspace{6pt}\huxbpad{0pt}} &
\multicolumn{3}{c!{\huxvb{0, 0, 0}{0}}}{\huxtpad{0pt + 1em}\centering \hspace{6pt} \textbf{Raw freq} \hspace{6pt}\huxbpad{0pt}} \tabularnewline[-0.5pt]

\hhline{>{\huxb{0, 0, 0}{0.5}}|>{\huxb{0, 0, 0}{0.5}}|>{\huxb{0, 0, 0}{0.5}}|}
\arrayrulecolor{black}

\multicolumn{1}{!{\huxvb{0, 0, 0}{0}}l!{\huxvb{0, 0, 0}{0.5}}}{\huxtpad{6pt + 1em}\raggedright \hspace{6pt} \textbf{ } \hspace{6pt}\huxbpad{6pt}} &
\multicolumn{1}{r!{\huxvb{0, 0, 0}{0}}}{\huxtpad{6pt + 1em}\raggedleft \hspace{6pt} \textbf{All} \hspace{6pt}\huxbpad{6pt}} &
\multicolumn{1}{r!{\huxvb{0, 0, 0}{0}}}{\huxtpad{6pt + 1em}\raggedleft \hspace{6pt} \textbf{Domestic} \hspace{6pt}\huxbpad{6pt}} &
\multicolumn{1}{r!{\huxvb{0, 0, 0}{0.5}}}{\huxtpad{6pt + 1em}\raggedleft \hspace{6pt} \textbf{Inter.} \hspace{6pt}\huxbpad{6pt}} &
\multicolumn{1}{r!{\huxvb{0, 0, 0}{0}}}{\huxtpad{6pt + 1em}\raggedleft \hspace{6pt} \textbf{All} \hspace{6pt}\huxbpad{6pt}} &
\multicolumn{1}{r!{\huxvb{0, 0, 0}{0}}}{\huxtpad{6pt + 1em}\raggedleft \hspace{6pt} \textbf{Domestic} \hspace{6pt}\huxbpad{6pt}} &
\multicolumn{1}{r!{\huxvb{0, 0, 0}{0.5}}}{\huxtpad{6pt + 1em}\raggedleft \hspace{6pt} \textbf{Inter.} \hspace{6pt}\huxbpad{6pt}} &
\multicolumn{1}{r!{\huxvb{0, 0, 0}{0}}}{\huxtpad{6pt + 1em}\raggedleft \hspace{6pt} \textbf{All} \hspace{6pt}\huxbpad{6pt}} &
\multicolumn{1}{r!{\huxvb{0, 0, 0}{0}}}{\huxtpad{6pt + 1em}\raggedleft \hspace{6pt} \textbf{Domestic} \hspace{6pt}\huxbpad{6pt}} &
\multicolumn{1}{r!{\huxvb{0, 0, 0}{0}}}{\huxtpad{6pt + 1em}\raggedleft \hspace{6pt} \textbf{Inter.} \hspace{6pt}\huxbpad{6pt}} \tabularnewline[-0.5pt]

\hhline{>{\huxb{0, 0, 0}{0.5}}->{\huxb{0, 0, 0}{0.5}}->{\huxb{0, 0, 0}{0.5}}->{\huxb{0, 0, 0}{0.5}}->{\huxb{0, 0, 0}{0.5}}->{\huxb{0, 0, 0}{0.5}}->{\huxb{0, 0, 0}{0.5}}->{\huxb{0, 0, 0}{0.5}}->{\huxb{0, 0, 0}{0.5}}->{\huxb{0, 0, 0}{0.5}}-}
\arrayrulecolor{black}

\multicolumn{1}{!{\huxvb{0, 0, 0}{0}}l!{\huxvb{0, 0, 0}{0.5}}}{\huxtpad{6pt + 1em}\raggedright \hspace{6pt} \textbf{Embedding cosine} \hspace{6pt}\huxbpad{6pt}} &
\multicolumn{1}{r!{\huxvb{0, 0, 0}{0}}}{\huxtpad{6pt + 1em}\raggedleft \hspace{6pt} *0.713 \hspace{6pt}\huxbpad{6pt}} &
\multicolumn{1}{r!{\huxvb{0, 0, 0}{0}}}{\huxtpad{6pt + 1em}\raggedleft \hspace{6pt} *0.76 \hspace{6pt}\huxbpad{6pt}} &
\multicolumn{1}{r!{\huxvb{0, 0, 0}{0.5}}}{\huxtpad{6pt + 1em}\raggedleft \hspace{6pt} *0.737 \hspace{6pt}\huxbpad{6pt}} &
\multicolumn{1}{r!{\huxvb{0, 0, 0}{0}}}{\huxtpad{6pt + 1em}\raggedleft \hspace{6pt} *0.702 \hspace{6pt}\huxbpad{6pt}} &
\multicolumn{1}{r!{\huxvb{0, 0, 0}{0}}}{\huxtpad{6pt + 1em}\raggedleft \hspace{6pt} *0.749 \hspace{6pt}\huxbpad{6pt}} &
\multicolumn{1}{r!{\huxvb{0, 0, 0}{0.5}}}{\huxtpad{6pt + 1em}\raggedleft \hspace{6pt} *0.713 \hspace{6pt}\huxbpad{6pt}} &
\multicolumn{1}{r!{\huxvb{0, 0, 0}{0}}}{\huxtpad{6pt + 1em}\raggedleft \hspace{6pt} *0.715 \hspace{6pt}\huxbpad{6pt}} &
\multicolumn{1}{r!{\huxvb{0, 0, 0}{0}}}{\huxtpad{6pt + 1em}\raggedleft \hspace{6pt} 0.764 \hspace{6pt}\huxbpad{6pt}} &
\multicolumn{1}{r!{\huxvb{0, 0, 0}{0}}}{\huxtpad{6pt + 1em}\raggedleft \hspace{6pt} 0.748 \hspace{6pt}\huxbpad{6pt}} \tabularnewline[-0.5pt]

\hhline{>{\huxb{0, 0, 0}{0.5}}|>{\huxb{0, 0, 0}{0.5}}|>{\huxb{0, 0, 0}{0.5}}|}
\arrayrulecolor{black}

\multicolumn{1}{!{\huxvb{0, 0, 0}{0}}l!{\huxvb{0, 0, 0}{0.5}}}{\huxtpad{6pt + 1em}\raggedright \hspace{6pt} \textbf{Levy's Euclidean} \hspace{6pt}\huxbpad{6pt}} &
\multicolumn{1}{r!{\huxvb{0, 0, 0}{0}}}{\huxtpad{6pt + 1em}\raggedleft \hspace{6pt} 0.803 \hspace{6pt}\huxbpad{6pt}} &
\multicolumn{1}{r!{\huxvb{0, 0, 0}{0}}}{\huxtpad{6pt + 1em}\raggedleft \hspace{6pt} 0.791 \hspace{6pt}\huxbpad{6pt}} &
\multicolumn{1}{r!{\huxvb{0, 0, 0}{0.5}}}{\huxtpad{6pt + 1em}\raggedleft \hspace{6pt} 0.771 \hspace{6pt}\huxbpad{6pt}} &
\multicolumn{1}{r!{\huxvb{0, 0, 0}{0}}}{\huxtpad{6pt + 1em}\raggedleft \hspace{6pt} 0.874 \hspace{6pt}\huxbpad{6pt}} &
\multicolumn{1}{r!{\huxvb{0, 0, 0}{0}}}{\huxtpad{6pt + 1em}\raggedleft \hspace{6pt} 0.867 \hspace{6pt}\huxbpad{6pt}} &
\multicolumn{1}{r!{\huxvb{0, 0, 0}{0.5}}}{\huxtpad{6pt + 1em}\raggedleft \hspace{6pt} 0.844 \hspace{6pt}\huxbpad{6pt}} &
\multicolumn{1}{r!{\huxvb{0, 0, 0}{0}}}{\huxtpad{6pt + 1em}\raggedleft \hspace{6pt} 0.725 \hspace{6pt}\huxbpad{6pt}} &
\multicolumn{1}{r!{\huxvb{0, 0, 0}{0}}}{\huxtpad{6pt + 1em}\raggedleft \hspace{6pt} *0.726 \hspace{6pt}\huxbpad{6pt}} &
\multicolumn{1}{r!{\huxvb{0, 0, 0}{0}}}{\huxtpad{6pt + 1em}\raggedleft \hspace{6pt} *0.705 \hspace{6pt}\huxbpad{6pt}} \tabularnewline[-0.5pt]

\hhline{>{\huxb{0, 0, 0}{0.5}}|>{\huxb{0, 0, 0}{0.5}}|>{\huxb{0, 0, 0}{0.5}}|}
\arrayrulecolor{black}

\multicolumn{1}{!{\huxvb{0, 0, 0}{0}}l!{\huxvb{0, 0, 0}{0.5}}}{\huxtpad{6pt + 1em}\raggedright \hspace{6pt} \textbf{Embedding dot} \hspace{6pt}\huxbpad{6pt}} &
\multicolumn{1}{r!{\huxvb{0, 0, 0}{0}}}{\huxtpad{6pt + 1em}\raggedleft \hspace{6pt} 0.803 \hspace{6pt}\huxbpad{6pt}} &
\multicolumn{1}{r!{\huxvb{0, 0, 0}{0}}}{\huxtpad{6pt + 1em}\raggedleft \hspace{6pt} 0.825 \hspace{6pt}\huxbpad{6pt}} &
\multicolumn{1}{r!{\huxvb{0, 0, 0}{0.5}}}{\huxtpad{6pt + 1em}\raggedleft \hspace{6pt} 0.811 \hspace{6pt}\huxbpad{6pt}} &
\multicolumn{1}{r!{\huxvb{0, 0, 0}{0}}}{\huxtpad{6pt + 1em}\raggedleft \hspace{6pt} 0.851 \hspace{6pt}\huxbpad{6pt}} &
\multicolumn{1}{r!{\huxvb{0, 0, 0}{0}}}{\huxtpad{6pt + 1em}\raggedleft \hspace{6pt} 0.87 \hspace{6pt}\huxbpad{6pt}} &
\multicolumn{1}{r!{\huxvb{0, 0, 0}{0.5}}}{\huxtpad{6pt + 1em}\raggedleft \hspace{6pt} 0.854 \hspace{6pt}\huxbpad{6pt}} &
\multicolumn{1}{r!{\huxvb{0, 0, 0}{0}}}{\huxtpad{6pt + 1em}\raggedleft \hspace{6pt} 0.769 \hspace{6pt}\huxbpad{6pt}} &
\multicolumn{1}{r!{\huxvb{0, 0, 0}{0}}}{\huxtpad{6pt + 1em}\raggedleft \hspace{6pt} 0.798 \hspace{6pt}\huxbpad{6pt}} &
\multicolumn{1}{r!{\huxvb{0, 0, 0}{0}}}{\huxtpad{6pt + 1em}\raggedleft \hspace{6pt} 0.784 \hspace{6pt}\huxbpad{6pt}} \tabularnewline[-0.5pt]

\hhline{>{\huxb{0, 0, 0}{0.5}}|>{\huxb{0, 0, 0}{0.5}}|>{\huxb{0, 0, 0}{0.5}}|}
\arrayrulecolor{black}

\multicolumn{1}{!{\huxvb{0, 0, 0}{0}}l!{\huxvb{0, 0, 0}{0.5}}}{\huxtpad{6pt + 1em}\raggedright \hspace{6pt} \textbf{SVD cosine} \hspace{6pt}\huxbpad{6pt}} &
\multicolumn{1}{r!{\huxvb{0, 0, 0}{0}}}{\huxtpad{6pt + 1em}\raggedleft \hspace{6pt} 0.859 \hspace{6pt}\huxbpad{6pt}} &
\multicolumn{1}{r!{\huxvb{0, 0, 0}{0}}}{\huxtpad{6pt + 1em}\raggedleft \hspace{6pt} 0.835 \hspace{6pt}\huxbpad{6pt}} &
\multicolumn{1}{r!{\huxvb{0, 0, 0}{0.5}}}{\huxtpad{6pt + 1em}\raggedleft \hspace{6pt} 0.815 \hspace{6pt}\huxbpad{6pt}} &
\multicolumn{1}{r!{\huxvb{0, 0, 0}{0}}}{\huxtpad{6pt + 1em}\raggedleft \hspace{6pt} 0.874 \hspace{6pt}\huxbpad{6pt}} &
\multicolumn{1}{r!{\huxvb{0, 0, 0}{0}}}{\huxtpad{6pt + 1em}\raggedleft \hspace{6pt} 0.841 \hspace{6pt}\huxbpad{6pt}} &
\multicolumn{1}{r!{\huxvb{0, 0, 0}{0.5}}}{\huxtpad{6pt + 1em}\raggedleft \hspace{6pt} 0.82 \hspace{6pt}\huxbpad{6pt}} &
\multicolumn{1}{r!{\huxvb{0, 0, 0}{0}}}{\huxtpad{6pt + 1em}\raggedleft \hspace{6pt} 0.82 \hspace{6pt}\huxbpad{6pt}} &
\multicolumn{1}{r!{\huxvb{0, 0, 0}{0}}}{\huxtpad{6pt + 1em}\raggedleft \hspace{6pt} 0.812 \hspace{6pt}\huxbpad{6pt}} &
\multicolumn{1}{r!{\huxvb{0, 0, 0}{0}}}{\huxtpad{6pt + 1em}\raggedleft \hspace{6pt} 0.792 \hspace{6pt}\huxbpad{6pt}} \tabularnewline[-0.5pt]

\hhline{>{\huxb{0, 0, 0}{0.5}}|>{\huxb{0, 0, 0}{0.5}}|>{\huxb{0, 0, 0}{0.5}}|}
\arrayrulecolor{black}

\multicolumn{1}{!{\huxvb{0, 0, 0}{0}}l!{\huxvb{0, 0, 0}{0.5}}}{\huxtpad{6pt + 1em}\raggedright \hspace{6pt} \textbf{Laplacian cosine} \hspace{6pt}\huxbpad{6pt}} &
\multicolumn{1}{r!{\huxvb{0, 0, 0}{0}}}{\huxtpad{6pt + 1em}\raggedleft \hspace{6pt} 0.878 \hspace{6pt}\huxbpad{6pt}} &
\multicolumn{1}{r!{\huxvb{0, 0, 0}{0}}}{\huxtpad{6pt + 1em}\raggedleft \hspace{6pt} 0.891 \hspace{6pt}\huxbpad{6pt}} &
\multicolumn{1}{r!{\huxvb{0, 0, 0}{0.5}}}{\huxtpad{6pt + 1em}\raggedleft \hspace{6pt} 0.867 \hspace{6pt}\huxbpad{6pt}} &
\multicolumn{1}{r!{\huxvb{0, 0, 0}{0}}}{\huxtpad{6pt + 1em}\raggedleft \hspace{6pt} 0.894 \hspace{6pt}\huxbpad{6pt}} &
\multicolumn{1}{r!{\huxvb{0, 0, 0}{0}}}{\huxtpad{6pt + 1em}\raggedleft \hspace{6pt} 0.933 \hspace{6pt}\huxbpad{6pt}} &
\multicolumn{1}{r!{\huxvb{0, 0, 0}{0.5}}}{\huxtpad{6pt + 1em}\raggedleft \hspace{6pt} 0.904 \hspace{6pt}\huxbpad{6pt}} &
\multicolumn{1}{r!{\huxvb{0, 0, 0}{0}}}{\huxtpad{6pt + 1em}\raggedleft \hspace{6pt} 0.835 \hspace{6pt}\huxbpad{6pt}} &
\multicolumn{1}{r!{\huxvb{0, 0, 0}{0}}}{\huxtpad{6pt + 1em}\raggedleft \hspace{6pt} 0.837 \hspace{6pt}\huxbpad{6pt}} &
\multicolumn{1}{r!{\huxvb{0, 0, 0}{0}}}{\huxtpad{6pt + 1em}\raggedleft \hspace{6pt} 0.815 \hspace{6pt}\huxbpad{6pt}} \tabularnewline[-0.5pt]

\hhline{>{\huxb{0, 0, 0}{0.5}}|>{\huxb{0, 0, 0}{0.5}}|>{\huxb{0, 0, 0}{0.5}}|}
\arrayrulecolor{black}

\multicolumn{1}{!{\huxvb{0, 0, 0}{0}}l!{\huxvb{0, 0, 0}{0.5}}}{\huxtpad{6pt + 1em}\raggedright \hspace{6pt} \textbf{Levy's cosine} \hspace{6pt}\huxbpad{6pt}} &
\multicolumn{1}{r!{\huxvb{0, 0, 0}{0}}}{\huxtpad{6pt + 1em}\raggedleft \hspace{6pt} 0.881 \hspace{6pt}\huxbpad{6pt}} &
\multicolumn{1}{r!{\huxvb{0, 0, 0}{0}}}{\huxtpad{6pt + 1em}\raggedleft \hspace{6pt} 0.875 \hspace{6pt}\huxbpad{6pt}} &
\multicolumn{1}{r!{\huxvb{0, 0, 0}{0.5}}}{\huxtpad{6pt + 1em}\raggedleft \hspace{6pt} 0.86 \hspace{6pt}\huxbpad{6pt}} &
\multicolumn{1}{r!{\huxvb{0, 0, 0}{0}}}{\huxtpad{6pt + 1em}\raggedleft \hspace{6pt} 0.898 \hspace{6pt}\huxbpad{6pt}} &
\multicolumn{1}{r!{\huxvb{0, 0, 0}{0}}}{\huxtpad{6pt + 1em}\raggedleft \hspace{6pt} 0.882 \hspace{6pt}\huxbpad{6pt}} &
\multicolumn{1}{r!{\huxvb{0, 0, 0}{0.5}}}{\huxtpad{6pt + 1em}\raggedleft \hspace{6pt} 0.866 \hspace{6pt}\huxbpad{6pt}} &
\multicolumn{1}{r!{\huxvb{0, 0, 0}{0}}}{\huxtpad{6pt + 1em}\raggedleft \hspace{6pt} 0.844 \hspace{6pt}\huxbpad{6pt}} &
\multicolumn{1}{r!{\huxvb{0, 0, 0}{0}}}{\huxtpad{6pt + 1em}\raggedleft \hspace{6pt} 0.858 \hspace{6pt}\huxbpad{6pt}} &
\multicolumn{1}{r!{\huxvb{0, 0, 0}{0}}}{\huxtpad{6pt + 1em}\raggedleft \hspace{6pt} 0.841 \hspace{6pt}\huxbpad{6pt}} \tabularnewline[-0.5pt]

\hhline{>{\huxb{0, 0, 0}{0.5}}|>{\huxb{0, 0, 0}{0.5}}|>{\huxb{0, 0, 0}{0.5}}|}
\arrayrulecolor{black}

\multicolumn{1}{!{\huxvb{0, 0, 0}{0}}l!{\huxvb{0, 0, 0}{0.5}}}{\huxtpad{6pt + 1em}\raggedright \hspace{6pt} \textbf{PPR JSD} \hspace{6pt}\huxbpad{6pt}} &
\multicolumn{1}{r!{\huxvb{0, 0, 0}{0}}}{\huxtpad{6pt + 1em}\raggedleft \hspace{6pt} 0.888 \hspace{6pt}\huxbpad{6pt}} &
\multicolumn{1}{r!{\huxvb{0, 0, 0}{0}}}{\huxtpad{6pt + 1em}\raggedleft \hspace{6pt} 
7 \hspace{6pt}\huxbpad{6pt}} &
\multicolumn{1}{r!{\huxvb{0, 0, 0}{0.5}}}{\huxtpad{6pt + 1em}\raggedleft \hspace{6pt} 0.835 \hspace{6pt}\huxbpad{6pt}} &
\multicolumn{1}{r!{\huxvb{0, 0, 0}{0}}}{\huxtpad{6pt + 1em}\raggedleft \hspace{6pt} 0.871 \hspace{6pt}\huxbpad{6pt}} &
\multicolumn{1}{r!{\huxvb{0, 0, 0}{0}}}{\huxtpad{6pt + 1em}\raggedleft \hspace{6pt} 0.828 \hspace{6pt}\huxbpad{6pt}} &
\multicolumn{1}{r!{\huxvb{0, 0, 0}{0.5}}}{\huxtpad{6pt + 1em}\raggedleft \hspace{6pt} 0.814 \hspace{6pt}\huxbpad{6pt}} &
\multicolumn{1}{r!{\huxvb{0, 0, 0}{0}}}{\huxtpad{6pt + 1em}\raggedleft \hspace{6pt} 0.865 \hspace{6pt}\huxbpad{6pt}} &
\multicolumn{1}{r!{\huxvb{0, 0, 0}{0}}}{\huxtpad{6pt + 1em}\raggedleft \hspace{6pt} 0.847 \hspace{6pt}\huxbpad{6pt}} &
\multicolumn{1}{r!{\huxvb{0, 0, 0}{0}}}{\huxtpad{6pt + 1em}\raggedleft \hspace{6pt} 0.834 \hspace{6pt}\huxbpad{6pt}} \tabularnewline[-0.5pt]

\hhline{>{\huxb{0, 0, 0}{0.5}}|>{\huxb{0, 0, 0}{0.5}}|>{\huxb{0, 0, 0}{0.5}}|}
\arrayrulecolor{black}

\multicolumn{1}{!{\huxvb{0, 0, 0}{0}}l!{\huxvb{0, 0, 0}{0.5}}}{\huxtpad{6pt + 1em}\raggedright \hspace{6pt} \textbf{Gravity MDS Euclidean} \hspace{6pt}\huxbpad{6pt}} &
\multicolumn{1}{r!{\huxvb{0, 0, 0}{0}}}{\huxtpad{6pt + 1em}\raggedleft \hspace{6pt} 0.904 \hspace{6pt}\huxbpad{6pt}} &
\multicolumn{1}{r!{\huxvb{0, 0, 0}{0}}}{\huxtpad{6pt + 1em}\raggedleft \hspace{6pt} 0.944 \hspace{6pt}\huxbpad{6pt}} &
\multicolumn{1}{r!{\huxvb{0, 0, 0}{0.5}}}{\huxtpad{6pt + 1em}\raggedleft \hspace{6pt} 0.93 \hspace{6pt}\huxbpad{6pt}} &
\multicolumn{1}{r!{\huxvb{0, 0, 0}{0}}}{\huxtpad{6pt + 1em}\raggedleft \hspace{6pt} 0.907 \hspace{6pt}\huxbpad{6pt}} &
\multicolumn{1}{r!{\huxvb{0, 0, 0}{0}}}{\huxtpad{6pt + 1em}\raggedleft \hspace{6pt} 0.972 \hspace{6pt}\huxbpad{6pt}} &
\multicolumn{1}{r!{\huxvb{0, 0, 0}{0.5}}}{\huxtpad{6pt + 1em}\raggedleft \hspace{6pt} 0.955 \hspace{6pt}\huxbpad{6pt}} &
\multicolumn{1}{r!{\huxvb{0, 0, 0}{0}}}{\huxtpad{6pt + 1em}\raggedleft \hspace{6pt} 0.793 \hspace{6pt}\huxbpad{6pt}} &
\multicolumn{1}{r!{\huxvb{0, 0, 0}{0}}}{\huxtpad{6pt + 1em}\raggedleft \hspace{6pt} 0.838 \hspace{6pt}\huxbpad{6pt}} &
\multicolumn{1}{r!{\huxvb{0, 0, 0}{0}}}{\huxtpad{6pt + 1em}\raggedleft \hspace{6pt} 0.821 \hspace{6pt}\huxbpad{6pt}} \tabularnewline[-0.5pt]

\hhline{>{\huxb{0, 0, 0}{0.5}}|>{\huxb{0, 0, 0}{0.5}}|>{\huxb{0, 0, 0}{0.5}}|}
\arrayrulecolor{black}

\multicolumn{1}{!{\huxvb{0, 0, 0}{0}}l!{\huxvb{0, 0, 0}{0.5}}}{\huxtpad{6pt + 1em}\raggedright \hspace{6pt} \textbf{PPR cosine} \hspace{6pt}\huxbpad{6pt}} &
\multicolumn{1}{r!{\huxvb{0, 0, 0}{0}}}{\huxtpad{6pt + 1em}\raggedleft \hspace{6pt} 0.918 \hspace{6pt}\huxbpad{6pt}} &
\multicolumn{1}{r!{\huxvb{0, 0, 0}{0}}}{\huxtpad{6pt + 1em}\raggedleft \hspace{6pt} 0.925 \hspace{6pt}\huxbpad{6pt}} &
\multicolumn{1}{r!{\huxvb{0, 0, 0}{0.5}}}{\huxtpad{6pt + 1em}\raggedleft \hspace{6pt} 0.908 \hspace{6pt}\huxbpad{6pt}} &
\multicolumn{1}{r!{\huxvb{0, 0, 0}{0}}}{\huxtpad{6pt + 1em}\raggedleft \hspace{6pt} 0.884 \hspace{6pt}\huxbpad{6pt}} &
\multicolumn{1}{r!{\huxvb{0, 0, 0}{0}}}{\huxtpad{6pt + 1em}\raggedleft \hspace{6pt} 0.916 \hspace{6pt}\huxbpad{6pt}} &
\multicolumn{1}{r!{\huxvb{0, 0, 0}{0.5}}}{\huxtpad{6pt + 1em}\raggedleft \hspace{6pt} 0.894 \hspace{6pt}\huxbpad{6pt}} &
\multicolumn{1}{r!{\huxvb{0, 0, 0}{0}}}{\huxtpad{6pt + 1em}\raggedleft \hspace{6pt} 0.812 \hspace{6pt}\huxbpad{6pt}} &
\multicolumn{1}{r!{\huxvb{0, 0, 0}{0}}}{\huxtpad{6pt + 1em}\raggedleft \hspace{6pt} 0.814 \hspace{6pt}\huxbpad{6pt}} &
\multicolumn{1}{r!{\huxvb{0, 0, 0}{0}}}{\huxtpad{6pt + 1em}\raggedleft \hspace{6pt} 0.796 \hspace{6pt}\huxbpad{6pt}} \tabularnewline[-0.5pt]

\hhline{>{\huxb{0, 0, 0}{0.5}}|>{\huxb{0, 0, 0}{0.5}}|>{\huxb{0, 0, 0}{0.5}}|}
\arrayrulecolor{black}

\multicolumn{1}{!{\huxvb{0, 0, 0}{0}}l!{\huxvb{0, 0, 0}{0.5}}}{\huxtpad{6pt + 1em}\raggedright \hspace{6pt} \textbf{Geographic} \hspace{6pt}\huxbpad{6pt}} &
\multicolumn{1}{r!{\huxvb{0, 0, 0}{0}}}{\huxtpad{6pt + 1em}\raggedleft \hspace{6pt} 0.929 \hspace{6pt}\huxbpad{6pt}} &
\multicolumn{1}{r!{\huxvb{0, 0, 0}{0}}}{\huxtpad{6pt + 1em}\raggedleft \hspace{6pt} 0.951 \hspace{6pt}\huxbpad{6pt}} &
\multicolumn{1}{r!{\huxvb{0, 0, 0}{0.5}}}{\huxtpad{6pt + 1em}\raggedleft \hspace{6pt} 0.928 \hspace{6pt}\huxbpad{6pt}} &
\multicolumn{1}{r!{\huxvb{0, 0, 0}{0}}}{\huxtpad{6pt + 1em}\raggedleft \hspace{6pt} 0.973 \hspace{6pt}\huxbpad{6pt}} &
\multicolumn{1}{r!{\huxvb{0, 0, 0}{0}}}{\huxtpad{6pt + 1em}\raggedleft \hspace{6pt} 1.002 \hspace{6pt}\huxbpad{6pt}} &
\multicolumn{1}{r!{\huxvb{0, 0, 0}{0.5}}}{\huxtpad{6pt + 1em}\raggedleft \hspace{6pt} 0.983 \hspace{6pt}\huxbpad{6pt}} &
\multicolumn{1}{r!{\huxvb{0, 0, 0}{0}}}{\huxtpad{6pt + 1em}\raggedleft \hspace{6pt} 0.856 \hspace{6pt}\huxbpad{6pt}} &
\multicolumn{1}{r!{\huxvb{0, 0, 0}{0}}}{\huxtpad{6pt + 1em}\raggedleft \hspace{6pt} 0.875 \hspace{6pt}\huxbpad{6pt}} &
\multicolumn{1}{r!{\huxvb{0, 0, 0}{0}}}{\huxtpad{6pt + 1em}\raggedleft \hspace{6pt} 0.853 \hspace{6pt}\huxbpad{6pt}} \tabularnewline[-0.5pt]

\hhline{>{\huxb{0, 0, 0}{0.5}}|>{\huxb{0, 0, 0}{0.5}}|>{\huxb{0, 0, 0}{0.5}}|}
\arrayrulecolor{black}

\multicolumn{1}{!{\huxvb{0, 0, 0}{0}}l!{\huxvb{0, 0, 0}{0.5}}}{\huxtpad{6pt + 1em}\raggedright \hspace{6pt} \textbf{Gravity SVD cosine} \hspace{6pt}\huxbpad{6pt}} &
\multicolumn{1}{r!{\huxvb{0, 0, 0}{0}}}{\huxtpad{6pt + 1em}\raggedleft \hspace{6pt} 0.933 \hspace{6pt}\huxbpad{6pt}} &
\multicolumn{1}{r!{\huxvb{0, 0, 0}{0}}}{\huxtpad{6pt + 1em}\raggedleft \hspace{6pt} 0.939 \hspace{6pt}\huxbpad{6pt}} &
\multicolumn{1}{r!{\huxvb{0, 0, 0}{0.5}}}{\huxtpad{6pt + 1em}\raggedleft \hspace{6pt} 0.923 \hspace{6pt}\huxbpad{6pt}} &
\multicolumn{1}{r!{\huxvb{0, 0, 0}{0}}}{\huxtpad{6pt + 1em}\raggedleft \hspace{6pt} 0.92 \hspace{6pt}\huxbpad{6pt}} &
\multicolumn{1}{r!{\huxvb{0, 0, 0}{0}}}{\huxtpad{6pt + 1em}\raggedleft \hspace{6pt} 0.946 \hspace{6pt}\huxbpad{6pt}} &
\multicolumn{1}{r!{\huxvb{0, 0, 0}{0.5}}}{\huxtpad{6pt + 1em}\raggedleft \hspace{6pt} 0.927 \hspace{6pt}\huxbpad{6pt}} &
\multicolumn{1}{r!{\huxvb{0, 0, 0}{0}}}{\huxtpad{6pt + 1em}\raggedleft \hspace{6pt} 0.853 \hspace{6pt}\huxbpad{6pt}} &
\multicolumn{1}{r!{\huxvb{0, 0, 0}{0}}}{\huxtpad{6pt + 1em}\raggedleft \hspace{6pt} 0.865 \hspace{6pt}\huxbpad{6pt}} &
\multicolumn{1}{r!{\huxvb{0, 0, 0}{0}}}{\huxtpad{6pt + 1em}\raggedleft \hspace{6pt} 0.847 \hspace{6pt}\huxbpad{6pt}} \tabularnewline[-0.5pt]

\hhline{>{\huxb{0, 0, 0}{0.5}}|>{\huxb{0, 0, 0}{0.5}}|>{\huxb{0, 0, 0}{0.5}}|}
\arrayrulecolor{black}

\multicolumn{1}{!{\huxvb{0, 0, 0}{0}}l!{\huxvb{0, 0, 0}{0.5}}}{\huxtpad{6pt + 1em}\raggedright \hspace{6pt} \textbf{Levy's dot} \hspace{6pt}\huxbpad{6pt}} &
\multicolumn{1}{r!{\huxvb{0, 0, 0}{0}}}{\huxtpad{6pt + 1em}\raggedleft \hspace{6pt} 0.987 \hspace{6pt}\huxbpad{6pt}} &
\multicolumn{1}{r!{\huxvb{0, 0, 0}{0}}}{\huxtpad{6pt + 1em}\raggedleft \hspace{6pt} 0.995 \hspace{6pt}\huxbpad{6pt}} &
\multicolumn{1}{r!{\huxvb{0, 0, 0}{0.5}}}{\huxtpad{6pt + 1em}\raggedleft \hspace{6pt} 0.98 \hspace{6pt}\huxbpad{6pt}} &
\multicolumn{1}{r!{\huxvb{0, 0, 0}{0}}}{\huxtpad{6pt + 1em}\raggedleft \hspace{6pt} 0.979 \hspace{6pt}\huxbpad{6pt}} &
\multicolumn{1}{r!{\huxvb{0, 0, 0}{0}}}{\huxtpad{6pt + 1em}\raggedleft \hspace{6pt} 1.014 \hspace{6pt}\huxbpad{6pt}} &
\multicolumn{1}{r!{\huxvb{0, 0, 0}{0.5}}}{\huxtpad{6pt + 1em}\raggedleft \hspace{6pt} 0.996 \hspace{6pt}\huxbpad{6pt}} &
\multicolumn{1}{r!{\huxvb{0, 0, 0}{0}}}{\huxtpad{6pt + 1em}\raggedleft \hspace{6pt} 0.853 \hspace{6pt}\huxbpad{6pt}} &
\multicolumn{1}{r!{\huxvb{0, 0, 0}{0}}}{\huxtpad{6pt + 1em}\raggedleft \hspace{6pt} 0.867 \hspace{6pt}\huxbpad{6pt}} &
\multicolumn{1}{r!{\huxvb{0, 0, 0}{0}}}{\huxtpad{6pt + 1em}\raggedleft \hspace{6pt} 0.849 \hspace{6pt}\huxbpad{6pt}} \tabularnewline[-0.5pt]

\hhline{>{\huxb{0, 0, 0}{0.5}}|>{\huxb{0, 0, 0}{0.5}}|>{\huxb{0, 0, 0}{0.5}}|}
\arrayrulecolor{black}
\end{tabular}
\end{threeparttable}\par\end{centerbox}

\end{table}

\end{landscape} 

%
%
\clearpage
\begin{landscape}

  \providecommand{\huxb}[2]{\arrayrulecolor[RGB]{#1}\global\arrayrulewidth=#2pt}
  \providecommand{\huxvb}[2]{\color[RGB]{#1}\vrule width #2pt}
  \providecommand{\huxtpad}[1]{\rule{0pt}{#1}}
  \providecommand{\huxbpad}[1]{\rule[-#1]{0pt}{#1}}

\begin{table}[ht]
\caption{
	\textbf{Prediction error between actual and predicted mobility with power-law gravity model, by metrics and experimental parameters}.
	Each cell corresponds to the prediction error (measured with root mean squared error) when using each distance as input to the power-law form of the gravity model of mobility to predict the flux between organizations, shown by subsets of mobility data, and by definitions of organization population.
	The asterisk denotes the top-performing distance metric by column (lowest prediction error).
	Distance metrics are ordered from lowest prediction error to highest, based on global mobility with organization population defined using all mobile and non-mobile authors.
	``All'' means that population is defined as the average yearly number of unique mobile and non-mobile scholars who published with the organizations' affiliation;
		population is defined in the same way for ``Mobile only'', except only using unique mobile researchers;
		``Raw freq'' means that organization populations are defined as their frequency across all the trajectories, similar to word frequency in language embedding.
		Embedding distance, measured as the cosine distance between embedding vectors, results in better predictions for global mobility with ``All'; and ``Mobile only'' definitions of population, though direvt gravity-law opimization with MDS performs better when using raw frequencies to measure organization population, and Levy's factorization~\autocite{levy2014neural} perform better here with domestic and international only mobility. 
}
\label{supp:table:rmse_power_table}
\begin{centerbox}
\begin{threeparttable}
\scriptsize
 \setlength{\tabcolsep}{0pt}
\begin{tabular}{l l l l l l l l l l}

\hhline{>{\huxb{0, 0, 0}{0.5}}|>{\huxb{0, 0, 0}{0.5}}|>{\huxb{0, 0, 0}{0.5}}|}
\arrayrulecolor{black}

\multicolumn{1}{!{\huxvb{0, 0, 0}{0}}c!{\huxvb{0, 0, 0}{0.5}}}{\huxtpad{0pt + 1em}\centering \hspace{6pt} \textbf{} \hspace{6pt}\huxbpad{0pt}} &
\multicolumn{3}{c!{\huxvb{0, 0, 0}{0.5}}}{\huxtpad{0pt + 1em}\centering \hspace{6pt} \textbf{All} \hspace{6pt}\huxbpad{0pt}} &
\multicolumn{3}{c!{\huxvb{0, 0, 0}{0.5}}}{\huxtpad{0pt + 1em}\centering \hspace{6pt} \textbf{Mobile only} \hspace{6pt}\huxbpad{0pt}} &
\multicolumn{3}{c!{\huxvb{0, 0, 0}{0}}}{\huxtpad{0pt + 1em}\centering \hspace{6pt} \textbf{Raw freq} \hspace{6pt}\huxbpad{0pt}} \tabularnewline[-0.5pt]

\hhline{>{\huxb{0, 0, 0}{0.5}}|>{\huxb{0, 0, 0}{0.5}}|>{\huxb{0, 0, 0}{0.5}}|}
\arrayrulecolor{black}

\multicolumn{1}{!{\huxvb{0, 0, 0}{0}}l!{\huxvb{0, 0, 0}{0.5}}}{\huxtpad{6pt + 1em}\raggedright \hspace{6pt} \textbf{ } \hspace{6pt}\huxbpad{6pt}} &
\multicolumn{1}{r!{\huxvb{0, 0, 0}{0}}}{\huxtpad{6pt + 1em}\raggedleft \hspace{6pt} \textbf{All} \hspace{6pt}\huxbpad{6pt}} &
\multicolumn{1}{r!{\huxvb{0, 0, 0}{0}}}{\huxtpad{6pt + 1em}\raggedleft \hspace{6pt} \textbf{Domestic} \hspace{6pt}\huxbpad{6pt}} &
\multicolumn{1}{r!{\huxvb{0, 0, 0}{0.5}}}{\huxtpad{6pt + 1em}\raggedleft \hspace{6pt} \textbf{Inter.} \hspace{6pt}\huxbpad{6pt}} &
\multicolumn{1}{r!{\huxvb{0, 0, 0}{0}}}{\huxtpad{6pt + 1em}\raggedleft \hspace{6pt} \textbf{All} \hspace{6pt}\huxbpad{6pt}} &
\multicolumn{1}{r!{\huxvb{0, 0, 0}{0}}}{\huxtpad{6pt + 1em}\raggedleft \hspace{6pt} \textbf{Domestic} \hspace{6pt}\huxbpad{6pt}} &
\multicolumn{1}{r!{\huxvb{0, 0, 0}{0.5}}}{\huxtpad{6pt + 1em}\raggedleft \hspace{6pt} \textbf{Inter.} \hspace{6pt}\huxbpad{6pt}} &
\multicolumn{1}{r!{\huxvb{0, 0, 0}{0}}}{\huxtpad{6pt + 1em}\raggedleft \hspace{6pt} \textbf{All} \hspace{6pt}\huxbpad{6pt}} &
\multicolumn{1}{r!{\huxvb{0, 0, 0}{0}}}{\huxtpad{6pt + 1em}\raggedleft \hspace{6pt} \textbf{Domestic} \hspace{6pt}\huxbpad{6pt}} &
\multicolumn{1}{r!{\huxvb{0, 0, 0}{0}}}{\huxtpad{6pt + 1em}\raggedleft \hspace{6pt} \textbf{Inter.} \hspace{6pt}\huxbpad{6pt}} \tabularnewline[-0.5pt]

\hhline{>{\huxb{0, 0, 0}{0.5}}->{\huxb{0, 0, 0}{0.5}}->{\huxb{0, 0, 0}{0.5}}->{\huxb{0, 0, 0}{0.5}}->{\huxb{0, 0, 0}{0.5}}->{\huxb{0, 0, 0}{0.5}}->{\huxb{0, 0, 0}{0.5}}->{\huxb{0, 0, 0}{0.5}}->{\huxb{0, 0, 0}{0.5}}->{\huxb{0, 0, 0}{0.5}}-}
\arrayrulecolor{black}

\multicolumn{1}{!{\huxvb{0, 0, 0}{0}}l!{\huxvb{0, 0, 0}{0.5}}}{\huxtpad{6pt + 1em}\raggedright \hspace{6pt} \textbf{Embedding cosine} \hspace{6pt}\huxbpad{6pt}} &
\multicolumn{1}{r!{\huxvb{0, 0, 0}{0}}}{\huxtpad{6pt + 1em}\raggedleft \hspace{6pt} *0.743 \hspace{6pt}\huxbpad{6pt}} &
\multicolumn{1}{r!{\huxvb{0, 0, 0}{0}}}{\huxtpad{6pt + 1em}\raggedleft \hspace{6pt} 0.784 \hspace{6pt}\huxbpad{6pt}} &
\multicolumn{1}{r!{\huxvb{0, 0, 0}{0.5}}}{\huxtpad{6pt + 1em}\raggedleft \hspace{6pt} 0.76 \hspace{6pt}\huxbpad{6pt}} &
\multicolumn{1}{r!{\huxvb{0, 0, 0}{0}}}{\huxtpad{6pt + 1em}\raggedleft \hspace{6pt} *0.73 \hspace{6pt}\huxbpad{6pt}} &
\multicolumn{1}{r!{\huxvb{0, 0, 0}{0}}}{\huxtpad{6pt + 1em}\raggedleft \hspace{6pt} *0.784 \hspace{6pt}\huxbpad{6pt}} &
\multicolumn{1}{r!{\huxvb{0, 0, 0}{0.5}}}{\huxtpad{6pt + 1em}\raggedleft \hspace{6pt} *0.749 \hspace{6pt}\huxbpad{6pt}} &
\multicolumn{1}{r!{\huxvb{0, 0, 0}{0}}}{\huxtpad{6pt + 1em}\raggedleft \hspace{6pt} 0.714 \hspace{6pt}\huxbpad{6pt}} &
\multicolumn{1}{r!{\huxvb{0, 0, 0}{0}}}{\huxtpad{6pt + 1em}\raggedleft \hspace{6pt} 0.762 \hspace{6pt}\huxbpad{6pt}} &
\multicolumn{1}{r!{\huxvb{0, 0, 0}{0}}}{\huxtpad{6pt + 1em}\raggedleft \hspace{6pt} 0.745 \hspace{6pt}\huxbpad{6pt}} \tabularnewline[-0.5pt]

\hhline{>{\huxb{0, 0, 0}{0.5}}|>{\huxb{0, 0, 0}{0.5}}|>{\huxb{0, 0, 0}{0.5}}|}
\arrayrulecolor{black}

\multicolumn{1}{!{\huxvb{0, 0, 0}{0}}l!{\huxvb{0, 0, 0}{0.5}}}{\huxtpad{6pt + 1em}\raggedright \hspace{6pt} \textbf{Levy's Euclidean} \hspace{6pt}\huxbpad{6pt}} &
\multicolumn{1}{r!{\huxvb{0, 0, 0}{0}}}{\huxtpad{6pt + 1em}\raggedleft \hspace{6pt} 0.78 \hspace{6pt}\huxbpad{6pt}} &
\multicolumn{1}{r!{\huxvb{0, 0, 0}{0}}}{\huxtpad{6pt + 1em}\raggedleft \hspace{6pt} *0.776 \hspace{6pt}\huxbpad{6pt}} &
\multicolumn{1}{r!{\huxvb{0, 0, 0}{0.5}}}{\huxtpad{6pt + 1em}\raggedleft \hspace{6pt} *0.755 \hspace{6pt}\huxbpad{6pt}} &
\multicolumn{1}{r!{\huxvb{0, 0, 0}{0}}}{\huxtpad{6pt + 1em}\raggedleft \hspace{6pt} 0.844 \hspace{6pt}\huxbpad{6pt}} &
\multicolumn{1}{r!{\huxvb{0, 0, 0}{0}}}{\huxtpad{6pt + 1em}\raggedleft \hspace{6pt} 0.847 \hspace{6pt}\huxbpad{6pt}} &
\multicolumn{1}{r!{\huxvb{0, 0, 0}{0.5}}}{\huxtpad{6pt + 1em}\raggedleft \hspace{6pt} 0.822 \hspace{6pt}\huxbpad{6pt}} &
\multicolumn{1}{r!{\huxvb{0, 0, 0}{0}}}{\huxtpad{6pt + 1em}\raggedleft \hspace{6pt} 0.703 \hspace{6pt}\huxbpad{6pt}} &
\multicolumn{1}{r!{\huxvb{0, 0, 0}{0}}}{\huxtpad{6pt + 1em}\raggedleft \hspace{6pt} *0.711 \hspace{6pt}\huxbpad{6pt}} &
\multicolumn{1}{r!{\huxvb{0, 0, 0}{0}}}{\huxtpad{6pt + 1em}\raggedleft \hspace{6pt} *0.691 \hspace{6pt}\huxbpad{6pt}} \tabularnewline[-0.5pt]

\hhline{>{\huxb{0, 0, 0}{0.5}}|>{\huxb{0, 0, 0}{0.5}}|>{\huxb{0, 0, 0}{0.5}}|}
\arrayrulecolor{black}

\multicolumn{1}{!{\huxvb{0, 0, 0}{0}}l!{\huxvb{0, 0, 0}{0.5}}}{\huxtpad{6pt + 1em}\raggedright \hspace{6pt} \textbf{Gravity MDS Euclidean} \hspace{6pt}\huxbpad{6pt}} &
\multicolumn{1}{r!{\huxvb{0, 0, 0}{0}}}{\huxtpad{6pt + 1em}\raggedleft \hspace{6pt} 0.795 \hspace{6pt}\huxbpad{6pt}} &
\multicolumn{1}{r!{\huxvb{0, 0, 0}{0}}}{\huxtpad{6pt + 1em}\raggedleft \hspace{6pt} 0.91 \hspace{6pt}\huxbpad{6pt}} &
\multicolumn{1}{r!{\huxvb{0, 0, 0}{0.5}}}{\huxtpad{6pt + 1em}\raggedleft \hspace{6pt} 0.898 \hspace{6pt}\huxbpad{6pt}} &
\multicolumn{1}{r!{\huxvb{0, 0, 0}{0}}}{\huxtpad{6pt + 1em}\raggedleft \hspace{6pt} 0.808 \hspace{6pt}\huxbpad{6pt}} &
\multicolumn{1}{r!{\huxvb{0, 0, 0}{0}}}{\huxtpad{6pt + 1em}\raggedleft \hspace{6pt} 0.957 \hspace{6pt}\huxbpad{6pt}} &
\multicolumn{1}{r!{\huxvb{0, 0, 0}{0.5}}}{\huxtpad{6pt + 1em}\raggedleft \hspace{6pt} 0.939 \hspace{6pt}\huxbpad{6pt}} &
\multicolumn{1}{r!{\huxvb{0, 0, 0}{0}}}{\huxtpad{6pt + 1em}\raggedleft \hspace{6pt} *0.691 \hspace{6pt}\huxbpad{6pt}} &
\multicolumn{1}{r!{\huxvb{0, 0, 0}{0}}}{\huxtpad{6pt + 1em}\raggedleft \hspace{6pt} 0.802 \hspace{6pt}\huxbpad{6pt}} &
\multicolumn{1}{r!{\huxvb{0, 0, 0}{0}}}{\huxtpad{6pt + 1em}\raggedleft \hspace{6pt} 0.79 \hspace{6pt}\huxbpad{6pt}} \tabularnewline[-0.5pt]

\hhline{>{\huxb{0, 0, 0}{0.5}}|>{\huxb{0, 0, 0}{0.5}}|>{\huxb{0, 0, 0}{0.5}}|}
\arrayrulecolor{black}

\multicolumn{1}{!{\huxvb{0, 0, 0}{0}}l!{\huxvb{0, 0, 0}{0.5}}}{\huxtpad{6pt + 1em}\raggedright \hspace{6pt} \textbf{Embedding dot} \hspace{6pt}\huxbpad{6pt}} &
\multicolumn{1}{r!{\huxvb{0, 0, 0}{0}}}{\huxtpad{6pt + 1em}\raggedleft \hspace{6pt} 0.822 \hspace{6pt}\huxbpad{6pt}} &
\multicolumn{1}{r!{\huxvb{0, 0, 0}{0}}}{\huxtpad{6pt + 1em}\raggedleft \hspace{6pt} 0.844 \hspace{6pt}\huxbpad{6pt}} &
\multicolumn{1}{r!{\huxvb{0, 0, 0}{0.5}}}{\huxtpad{6pt + 1em}\raggedleft \hspace{6pt} 0.831 \hspace{6pt}\huxbpad{6pt}} &
\multicolumn{1}{r!{\huxvb{0, 0, 0}{0}}}{\huxtpad{6pt + 1em}\raggedleft \hspace{6pt} 0.864 \hspace{6pt}\huxbpad{6pt}} &
\multicolumn{1}{r!{\huxvb{0, 0, 0}{0}}}{\huxtpad{6pt + 1em}\raggedleft \hspace{6pt} 0.879 \hspace{6pt}\huxbpad{6pt}} &
\multicolumn{1}{r!{\huxvb{0, 0, 0}{0.5}}}{\huxtpad{6pt + 1em}\raggedleft \hspace{6pt} 0.863 \hspace{6pt}\huxbpad{6pt}} &
\multicolumn{1}{r!{\huxvb{0, 0, 0}{0}}}{\huxtpad{6pt + 1em}\raggedleft \hspace{6pt} 0.783 \hspace{6pt}\huxbpad{6pt}} &
\multicolumn{1}{r!{\huxvb{0, 0, 0}{0}}}{\huxtpad{6pt + 1em}\raggedleft \hspace{6pt} 0.809 \hspace{6pt}\huxbpad{6pt}} &
\multicolumn{1}{r!{\huxvb{0, 0, 0}{0}}}{\huxtpad{6pt + 1em}\raggedleft \hspace{6pt} 0.795 \hspace{6pt}\huxbpad{6pt}} \tabularnewline[-0.5pt]

\hhline{>{\huxb{0, 0, 0}{0.5}}|>{\huxb{0, 0, 0}{0.5}}|>{\huxb{0, 0, 0}{0.5}}|}
\arrayrulecolor{black}

\multicolumn{1}{!{\huxvb{0, 0, 0}{0}}l!{\huxvb{0, 0, 0}{0.5}}}{\huxtpad{6pt + 1em}\raggedright \hspace{6pt} \textbf{SVD cosine} \hspace{6pt}\huxbpad{6pt}} &
\multicolumn{1}{r!{\huxvb{0, 0, 0}{0}}}{\huxtpad{6pt + 1em}\raggedleft \hspace{6pt} 0.839 \hspace{6pt}\huxbpad{6pt}} &
\multicolumn{1}{r!{\huxvb{0, 0, 0}{0}}}{\huxtpad{6pt + 1em}\raggedleft \hspace{6pt} 0.785 \hspace{6pt}\huxbpad{6pt}} &
\multicolumn{1}{r!{\huxvb{0, 0, 0}{0.5}}}{\huxtpad{6pt + 1em}\raggedleft \hspace{6pt} 0.761 \hspace{6pt}\huxbpad{6pt}} &
\multicolumn{1}{r!{\huxvb{0, 0, 0}{0}}}{\huxtpad{6pt + 1em}\raggedleft \hspace{6pt} 0.89 \hspace{6pt}\huxbpad{6pt}} &
\multicolumn{1}{r!{\huxvb{0, 0, 0}{0}}}{\huxtpad{6pt + 1em}\raggedleft \hspace{6pt} 0.834 \hspace{6pt}\huxbpad{6pt}} &
\multicolumn{1}{r!{\huxvb{0, 0, 0}{0.5}}}{\huxtpad{6pt + 1em}\raggedleft \hspace{6pt} 0.81 \hspace{6pt}\huxbpad{6pt}} &
\multicolumn{1}{r!{\huxvb{0, 0, 0}{0}}}{\huxtpad{6pt + 1em}\raggedleft \hspace{6pt} 0.792 \hspace{6pt}\huxbpad{6pt}} &
\multicolumn{1}{r!{\huxvb{0, 0, 0}{0}}}{\huxtpad{6pt + 1em}\raggedleft \hspace{6pt} 0.752 \hspace{6pt}\huxbpad{6pt}} &
\multicolumn{1}{r!{\huxvb{0, 0, 0}{0}}}{\huxtpad{6pt + 1em}\raggedleft \hspace{6pt} 0.728 \hspace{6pt}\huxbpad{6pt}} \tabularnewline[-0.5pt]

\hhline{>{\huxb{0, 0, 0}{0.5}}|>{\huxb{0, 0, 0}{0.5}}|>{\huxb{0, 0, 0}{0.5}}|}
\arrayrulecolor{black}

\multicolumn{1}{!{\huxvb{0, 0, 0}{0}}l!{\huxvb{0, 0, 0}{0.5}}}{\huxtpad{6pt + 1em}\raggedright \hspace{6pt} \textbf{Laplacian cosine} \hspace{6pt}\huxbpad{6pt}} &
\multicolumn{1}{r!{\huxvb{0, 0, 0}{0}}}{\huxtpad{6pt + 1em}\raggedleft \hspace{6pt} 0.87 \hspace{6pt}\huxbpad{6pt}} &
\multicolumn{1}{r!{\huxvb{0, 0, 0}{0}}}{\huxtpad{6pt + 1em}\raggedleft \hspace{6pt} 0.837 \hspace{6pt}\huxbpad{6pt}} &
\multicolumn{1}{r!{\huxvb{0, 0, 0}{0.5}}}{\huxtpad{6pt + 1em}\raggedleft \hspace{6pt} 0.801 \hspace{6pt}\huxbpad{6pt}} &
\multicolumn{1}{r!{\huxvb{0, 0, 0}{0}}}{\huxtpad{6pt + 1em}\raggedleft \hspace{6pt} 0.937 \hspace{6pt}\huxbpad{6pt}} &
\multicolumn{1}{r!{\huxvb{0, 0, 0}{0}}}{\huxtpad{6pt + 1em}\raggedleft \hspace{6pt} 0.919 \hspace{6pt}\huxbpad{6pt}} &
\multicolumn{1}{r!{\huxvb{0, 0, 0}{0.5}}}{\huxtpad{6pt + 1em}\raggedleft \hspace{6pt} 0.886 \hspace{6pt}\huxbpad{6pt}} &
\multicolumn{1}{r!{\huxvb{0, 0, 0}{0}}}{\huxtpad{6pt + 1em}\raggedleft \hspace{6pt} 0.804 \hspace{6pt}\huxbpad{6pt}} &
\multicolumn{1}{r!{\huxvb{0, 0, 0}{0}}}{\huxtpad{6pt + 1em}\raggedleft \hspace{6pt} 0.772 \hspace{6pt}\huxbpad{6pt}} &
\multicolumn{1}{r!{\huxvb{0, 0, 0}{0}}}{\huxtpad{6pt + 1em}\raggedleft \hspace{6pt} 0.735 \hspace{6pt}\huxbpad{6pt}} \tabularnewline[-0.5pt]

\hhline{>{\huxb{0, 0, 0}{0.5}}|>{\huxb{0, 0, 0}{0.5}}|>{\huxb{0, 0, 0}{0.5}}|}
\arrayrulecolor{black}

\multicolumn{1}{!{\huxvb{0, 0, 0}{0}}l!{\huxvb{0, 0, 0}{0.5}}}{\huxtpad{6pt + 1em}\raggedright \hspace{6pt} \textbf{Geographic} \hspace{6pt}\huxbpad{6pt}} &
\multicolumn{1}{r!{\huxvb{0, 0, 0}{0}}}{\huxtpad{6pt + 1em}\raggedleft \hspace{6pt} 0.874 \hspace{6pt}\huxbpad{6pt}} &
\multicolumn{1}{r!{\huxvb{0, 0, 0}{0}}}{\huxtpad{6pt + 1em}\raggedleft \hspace{6pt} 0.905 \hspace{6pt}\huxbpad{6pt}} &
\multicolumn{1}{r!{\huxvb{0, 0, 0}{0.5}}}{\huxtpad{6pt + 1em}\raggedleft \hspace{6pt} 0.879 \hspace{6pt}\huxbpad{6pt}} &
\multicolumn{1}{r!{\huxvb{0, 0, 0}{0}}}{\huxtpad{6pt + 1em}\raggedleft \hspace{6pt} 0.888 \hspace{6pt}\huxbpad{6pt}} &
\multicolumn{1}{r!{\huxvb{0, 0, 0}{0}}}{\huxtpad{6pt + 1em}\raggedleft \hspace{6pt} 0.933 \hspace{6pt}\huxbpad{6pt}} &
\multicolumn{1}{r!{\huxvb{0, 0, 0}{0.5}}}{\huxtpad{6pt + 1em}\raggedleft \hspace{6pt} 0.906 \hspace{6pt}\huxbpad{6pt}} &
\multicolumn{1}{r!{\huxvb{0, 0, 0}{0}}}{\huxtpad{6pt + 1em}\raggedleft \hspace{6pt} 0.852 \hspace{6pt}\huxbpad{6pt}} &
\multicolumn{1}{r!{\huxvb{0, 0, 0}{0}}}{\huxtpad{6pt + 1em}\raggedleft \hspace{6pt} 0.874 \hspace{6pt}\huxbpad{6pt}} &
\multicolumn{1}{r!{\huxvb{0, 0, 0}{0}}}{\huxtpad{6pt + 1em}\raggedleft \hspace{6pt} 0.851 \hspace{6pt}\huxbpad{6pt}} \tabularnewline[-0.5pt]

\hhline{>{\huxb{0, 0, 0}{0.5}}|>{\huxb{0, 0, 0}{0.5}}|>{\huxb{0, 0, 0}{0.5}}|}
\arrayrulecolor{black}

\multicolumn{1}{!{\huxvb{0, 0, 0}{0}}l!{\huxvb{0, 0, 0}{0.5}}}{\huxtpad{6pt + 1em}\raggedright \hspace{6pt} \textbf{PPR JSD} \hspace{6pt}\huxbpad{6pt}} &
\multicolumn{1}{r!{\huxvb{0, 0, 0}{0}}}{\huxtpad{6pt + 1em}\raggedleft \hspace{6pt} 0.889 \hspace{6pt}\huxbpad{6pt}} &
\multicolumn{1}{r!{\huxvb{0, 0, 0}{0}}}{\huxtpad{6pt + 1em}\raggedleft \hspace{6pt} 0.85 \hspace{6pt}\huxbpad{6pt}} &
\multicolumn{1}{r!{\huxvb{0, 0, 0}{0.5}}}{\huxtpad{6pt + 1em}\raggedleft \hspace{6pt} 0.838 \hspace{6pt}\huxbpad{6pt}} &
\multicolumn{1}{r!{\huxvb{0, 0, 0}{0}}}{\huxtpad{6pt + 1em}\raggedleft \hspace{6pt} 0.871 \hspace{6pt}\huxbpad{6pt}} &
\multicolumn{1}{r!{\huxvb{0, 0, 0}{0}}}{\huxtpad{6pt + 1em}\raggedleft \hspace{6pt} 0.834 \hspace{6pt}\huxbpad{6pt}} &
\multicolumn{1}{r!{\huxvb{0, 0, 0}{0.5}}}{\huxtpad{6pt + 1em}\raggedleft \hspace{6pt} 0.819 \hspace{6pt}\huxbpad{6pt}} &
\multicolumn{1}{r!{\huxvb{0, 0, 0}{0}}}{\huxtpad{6pt + 1em}\raggedleft \hspace{6pt} 0.865 \hspace{6pt}\huxbpad{6pt}} &
\multicolumn{1}{r!{\huxvb{0, 0, 0}{0}}}{\huxtpad{6pt + 1em}\raggedleft \hspace{6pt} 0.847 \hspace{6pt}\huxbpad{6pt}} &
\multicolumn{1}{r!{\huxvb{0, 0, 0}{0}}}{\huxtpad{6pt + 1em}\raggedleft \hspace{6pt} 0.834 \hspace{6pt}\huxbpad{6pt}} \tabularnewline[-0.5pt]

\hhline{>{\huxb{0, 0, 0}{0.5}}|>{\huxb{0, 0, 0}{0.5}}|>{\huxb{0, 0, 0}{0.5}}|}
\arrayrulecolor{black}

\multicolumn{1}{!{\huxvb{0, 0, 0}{0}}l!{\huxvb{0, 0, 0}{0.5}}}{\huxtpad{6pt + 1em}\raggedright \hspace{6pt} \textbf{PPR cosine} \hspace{6pt}\huxbpad{6pt}} &
\multicolumn{1}{r!{\huxvb{0, 0, 0}{0}}}{\huxtpad{6pt + 1em}\raggedleft \hspace{6pt} 0.92 \hspace{6pt}\huxbpad{6pt}} &
\multicolumn{1}{r!{\huxvb{0, 0, 0}{0}}}{\huxtpad{6pt + 1em}\raggedleft \hspace{6pt} 0.927 \hspace{6pt}\huxbpad{6pt}} &
\multicolumn{1}{r!{\huxvb{0, 0, 0}{0.5}}}{\huxtpad{6pt + 1em}\raggedleft \hspace{6pt} 0.91 \hspace{6pt}\huxbpad{6pt}} &
\multicolumn{1}{r!{\huxvb{0, 0, 0}{0}}}{\huxtpad{6pt + 1em}\raggedleft \hspace{6pt} 0.885 \hspace{6pt}\huxbpad{6pt}} &
\multicolumn{1}{r!{\huxvb{0, 0, 0}{0}}}{\huxtpad{6pt + 1em}\raggedleft \hspace{6pt} 0.918 \hspace{6pt}\huxbpad{6pt}} &
\multicolumn{1}{r!{\huxvb{0, 0, 0}{0.5}}}{\huxtpad{6pt + 1em}\raggedleft \hspace{6pt} 0.896 \hspace{6pt}\huxbpad{6pt}} &
\multicolumn{1}{r!{\huxvb{0, 0, 0}{0}}}{\huxtpad{6pt + 1em}\raggedleft \hspace{6pt} 0.812 \hspace{6pt}\huxbpad{6pt}} &
\multicolumn{1}{r!{\huxvb{0, 0, 0}{0}}}{\huxtpad{6pt + 1em}\raggedleft \hspace{6pt} 0.815 \hspace{6pt}\huxbpad{6pt}} &
\multicolumn{1}{r!{\huxvb{0, 0, 0}{0}}}{\huxtpad{6pt + 1em}\raggedleft \hspace{6pt} 0.797 \hspace{6pt}\huxbpad{6pt}} \tabularnewline[-0.5pt]

\hhline{>{\huxb{0, 0, 0}{0.5}}|>{\huxb{0, 0, 0}{0.5}}|>{\huxb{0, 0, 0}{0.5}}|}
\arrayrulecolor{black}

\multicolumn{1}{!{\huxvb{0, 0, 0}{0}}l!{\huxvb{0, 0, 0}{0.5}}}{\huxtpad{6pt + 1em}\raggedright \hspace{6pt} \textbf{Levy's cosine} \hspace{6pt}\huxbpad{6pt}} &
\multicolumn{1}{r!{\huxvb{0, 0, 0}{0}}}{\huxtpad{6pt + 1em}\raggedleft \hspace{6pt} 0.927 \hspace{6pt}\huxbpad{6pt}} &
\multicolumn{1}{r!{\huxvb{0, 0, 0}{0}}}{\huxtpad{6pt + 1em}\raggedleft \hspace{6pt} 0.926 \hspace{6pt}\huxbpad{6pt}} &
\multicolumn{1}{r!{\huxvb{0, 0, 0}{0.5}}}{\huxtpad{6pt + 1em}\raggedleft \hspace{6pt} 0.911 \hspace{6pt}\huxbpad{6pt}} &
\multicolumn{1}{r!{\huxvb{0, 0, 0}{0}}}{\huxtpad{6pt + 1em}\raggedleft \hspace{6pt} 0.93 \hspace{6pt}\huxbpad{6pt}} &
\multicolumn{1}{r!{\huxvb{0, 0, 0}{0}}}{\huxtpad{6pt + 1em}\raggedleft \hspace{6pt} 0.924 \hspace{6pt}\huxbpad{6pt}} &
\multicolumn{1}{r!{\huxvb{0, 0, 0}{0.5}}}{\huxtpad{6pt + 1em}\raggedleft \hspace{6pt} 0.909 \hspace{6pt}\huxbpad{6pt}} &
\multicolumn{1}{r!{\huxvb{0, 0, 0}{0}}}{\huxtpad{6pt + 1em}\raggedleft \hspace{6pt} 0.861 \hspace{6pt}\huxbpad{6pt}} &
\multicolumn{1}{r!{\huxvb{0, 0, 0}{0}}}{\huxtpad{6pt + 1em}\raggedleft \hspace{6pt} 0.872 \hspace{6pt}\huxbpad{6pt}} &
\multicolumn{1}{r!{\huxvb{0, 0, 0}{0}}}{\huxtpad{6pt + 1em}\raggedleft \hspace{6pt} 0.854 \hspace{6pt}\huxbpad{6pt}} \tabularnewline[-0.5pt]

\hhline{>{\huxb{0, 0, 0}{0.5}}|>{\huxb{0, 0, 0}{0.5}}|>{\huxb{0, 0, 0}{0.5}}|}
\arrayrulecolor{black}

\multicolumn{1}{!{\huxvb{0, 0, 0}{0}}l!{\huxvb{0, 0, 0}{0.5}}}{\huxtpad{6pt + 1em}\raggedright \hspace{6pt} \textbf{Gravity SVD cosine} \hspace{6pt}\huxbpad{6pt}} &
\multicolumn{1}{r!{\huxvb{0, 0, 0}{0}}}{\huxtpad{6pt + 1em}\raggedleft \hspace{6pt} 0.965 \hspace{6pt}\huxbpad{6pt}} &
\multicolumn{1}{r!{\huxvb{0, 0, 0}{0}}}{\huxtpad{6pt + 1em}\raggedleft \hspace{6pt} 0.965 \hspace{6pt}\huxbpad{6pt}} &
\multicolumn{1}{r!{\huxvb{0, 0, 0}{0.5}}}{\huxtpad{6pt + 1em}\raggedleft \hspace{6pt} 0.95 \hspace{6pt}\huxbpad{6pt}} &
\multicolumn{1}{r!{\huxvb{0, 0, 0}{0}}}{\huxtpad{6pt + 1em}\raggedleft \hspace{6pt} 0.955 \hspace{6pt}\huxbpad{6pt}} &
\multicolumn{1}{r!{\huxvb{0, 0, 0}{0}}}{\huxtpad{6pt + 1em}\raggedleft \hspace{6pt} 0.958 \hspace{6pt}\huxbpad{6pt}} &
\multicolumn{1}{r!{\huxvb{0, 0, 0}{0.5}}}{\huxtpad{6pt + 1em}\raggedleft \hspace{6pt} 0.941 \hspace{6pt}\huxbpad{6pt}} &
\multicolumn{1}{r!{\huxvb{0, 0, 0}{0}}}{\huxtpad{6pt + 1em}\raggedleft \hspace{6pt} 0.873 \hspace{6pt}\huxbpad{6pt}} &
\multicolumn{1}{r!{\huxvb{0, 0, 0}{0}}}{\huxtpad{6pt + 1em}\raggedleft \hspace{6pt} 0.883 \hspace{6pt}\huxbpad{6pt}} &
\multicolumn{1}{r!{\huxvb{0, 0, 0}{0}}}{\huxtpad{6pt + 1em}\raggedleft \hspace{6pt} 0.866 \hspace{6pt}\huxbpad{6pt}} \tabularnewline[-0.5pt]

\hhline{>{\huxb{0, 0, 0}{0.5}}|>{\huxb{0, 0, 0}{0.5}}|>{\huxb{0, 0, 0}{0.5}}|}
\arrayrulecolor{black}

\multicolumn{1}{!{\huxvb{0, 0, 0}{0}}l!{\huxvb{0, 0, 0}{0.5}}}{\huxtpad{6pt + 1em}\raggedright \hspace{6pt} \textbf{Levy's dot} \hspace{6pt}\huxbpad{6pt}} &
\multicolumn{1}{r!{\huxvb{0, 0, 0}{0}}}{\huxtpad{6pt + 1em}\raggedleft \hspace{6pt} 0.99 \hspace{6pt}\huxbpad{6pt}} &
\multicolumn{1}{r!{\huxvb{0, 0, 0}{0}}}{\huxtpad{6pt + 1em}\raggedleft \hspace{6pt} 0.998 \hspace{6pt}\huxbpad{6pt}} &
\multicolumn{1}{r!{\huxvb{0, 0, 0}{0.5}}}{\huxtpad{6pt + 1em}\raggedleft \hspace{6pt} 0.983 \hspace{6pt}\huxbpad{6pt}} &
\multicolumn{1}{r!{\huxvb{0, 0, 0}{0}}}{\huxtpad{6pt + 1em}\raggedleft \hspace{6pt} 0.977 \hspace{6pt}\huxbpad{6pt}} &
\multicolumn{1}{r!{\huxvb{0, 0, 0}{0}}}{\huxtpad{6pt + 1em}\raggedleft \hspace{6pt} 1.015 \hspace{6pt}\huxbpad{6pt}} &
\multicolumn{1}{r!{\huxvb{0, 0, 0}{0.5}}}{\huxtpad{6pt + 1em}\raggedleft \hspace{6pt} 0.997 \hspace{6pt}\huxbpad{6pt}} &
\multicolumn{1}{r!{\huxvb{0, 0, 0}{0}}}{\huxtpad{6pt + 1em}\raggedleft \hspace{6pt} 0.845 \hspace{6pt}\huxbpad{6pt}} &
\multicolumn{1}{r!{\huxvb{0, 0, 0}{0}}}{\huxtpad{6pt + 1em}\raggedleft \hspace{6pt} 0.86 \hspace{6pt}\huxbpad{6pt}} &
\multicolumn{1}{r!{\huxvb{0, 0, 0}{0}}}{\huxtpad{6pt + 1em}\raggedleft \hspace{6pt} 0.841 \hspace{6pt}\huxbpad{6pt}} \tabularnewline[-0.5pt]

\hhline{>{\huxb{0, 0, 0}{0.5}}|>{\huxb{0, 0, 0}{0.5}}|>{\huxb{0, 0, 0}{0.5}}|}
\arrayrulecolor{black}
\end{tabular}
\end{threeparttable}\par\end{centerbox}

\end{table}

\end{landscape}

\begin{table}[!htbp] \centering 
  \caption{\textbf{SemAxis captures prestige across multiple countries}.
  The Spearman's $\rho$ between the prestige axis generated by SemAxis and the ranking of universities according to their fractional mean normalized citation score in the Leiden rankings~\autocite{waltman2012leidenrankings}.
  Shown for the top ten countries by number of universities represented in the Leiden rankings. 
  The prestige axis is generated for each country using their top and bottom 5 ranked universities.
  We note that the lowest value, $\rho=0.458$ is for China, which may be a consequence of the difficulty in disambiguating the names of authors with Chinese names.
  } 
  \label{table:supp:semaxis_prestige_bycountry} 
\begin{tabular}{@{\extracolsep{5pt}} ccc} 
\\[-1.8ex]\hline 
\hline \\[-1.8ex] 
Country & Count of orgs & Spearman's $\rho$ \\ 
\hline \\[-1.8ex] 
USA & $172$ & $0.763$ \\ 
CHN & $165$ & $0.458$ \\ 
DEU & $50$ & $0.601$ \\ 
GBR & $45$ & $0.915$ \\ 
JPN & $41$ & $0.651$ \\ 
ITA & $40$ & $0.543$ \\ 
KOR & $35$ & $0.756$ \\ 
ESP & $34$ & $0.889$ \\ 
CAN & $27$ & $0.761$ \\ 
AUS & $26$ & $0.718$ \\ 
\hline \\[-1.8ex] 
\end{tabular} 
\end{table}

%
%
%
%
%
%
%
%
\newpage

%
%
\begin{figure}[p!]
	\centering
	\includegraphics[width=\textwidth]{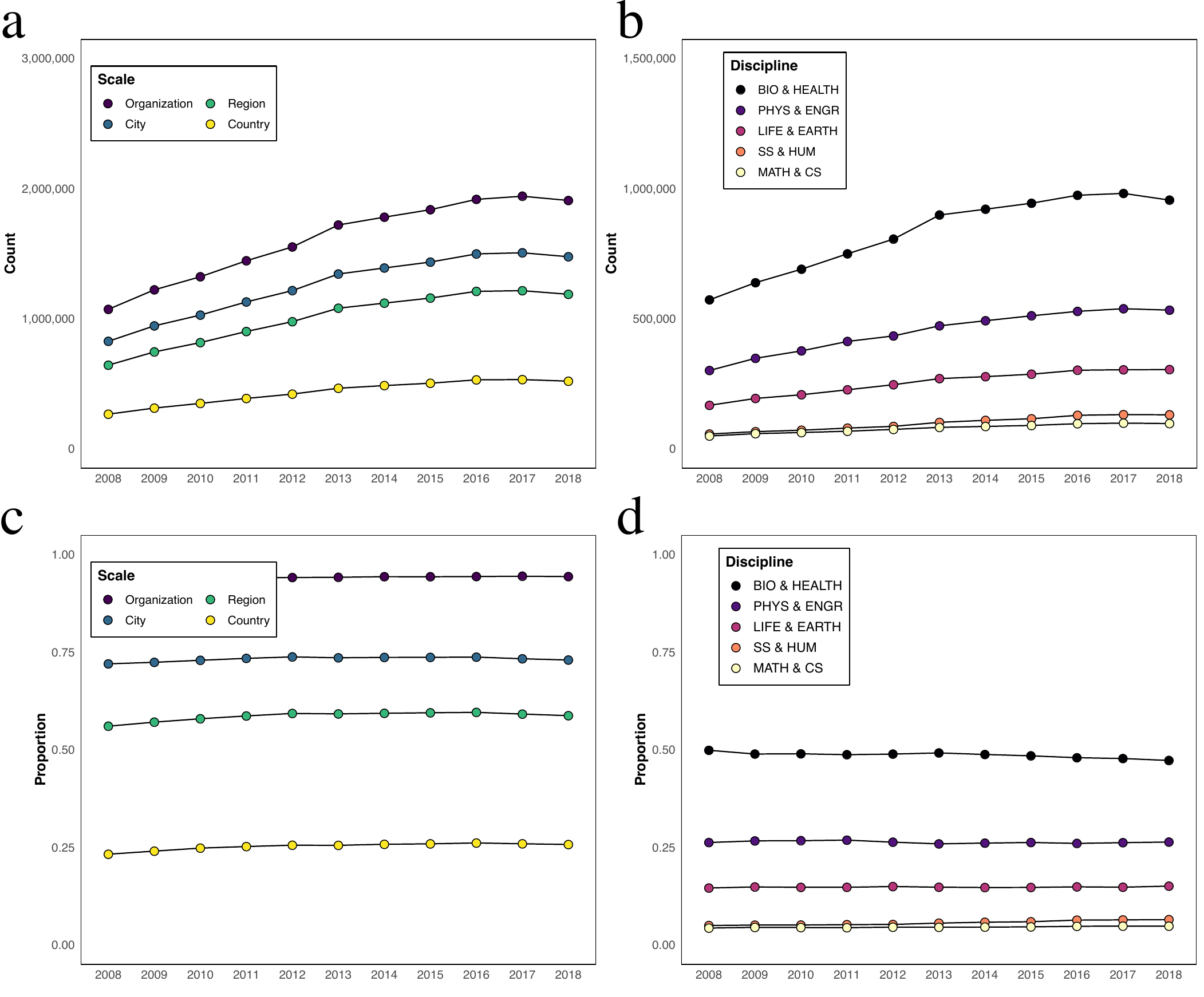}
	\caption{
		\textbf{Publications over time.}
		\textbf{a.}
		The number of papers published by mobile authors has been steadily increasing from 2008 to 2017, with a small decrease in 2018,  which may be due to an artifact of the Web of Science indexing process.
		Lines correspond to publications by mobile authors, by authors with affiliations in at least two cities, at least two regions, and at least two countries.
		We did not find major changes in the publication patterns of mobile authors during this time period.
		\textbf{b.}
		Lines correspond to the proportion of publications classified as Biology and Health (black), Physics and Engineering (purple), Life and Earth Science (magenta), Social Science and Humanities (orange), and Math and Computer Science (yellow).
		The rate of publication in Biology and Health has leveled since about 2013, whereas the rate of publication in other fields has steadily increased.
		\textbf{c.}
		While the absolute count of publications has increased, the percentage of mobile scholars, and those with affiliations in at least least two cities, regions, or countries, as a proportion of all publications, has remained stable over time.
		\textbf{d.}
		The proportion of authors' publications across fields has largely remained steady.
		Biology and Health Science has comprised the majority of publications across nearly all years but has steadily declined in proportion.
		However, the proportion of Social Science and Humanities publications has been steadily increasing.
	}
	\label{fig:supp:pubs_over_time}
\end{figure}

%
%
\begin{figure}[p!]
	\centering
	\includegraphics[width=\textwidth]{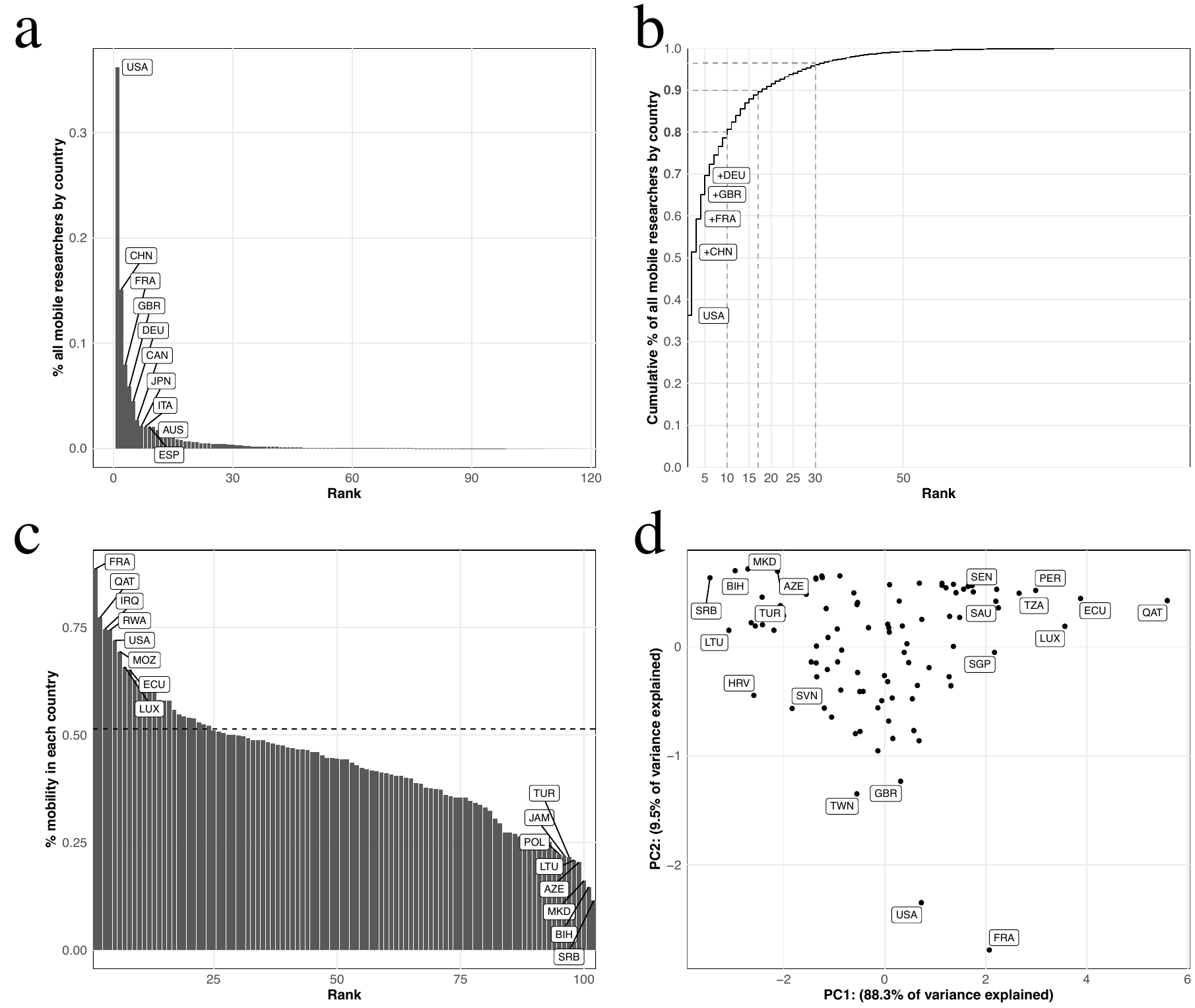}
	\caption{
		\textbf{Extent and nature of mobility by country.}
		\textbf{a.}
		The proportion of all mobile researchers contributed by each country.
		Over 30\% of all mobile researchers have been affiliated with organizations in the U.S. during the period of study.
		\textbf{b.}
		Cumulative distribution of data shown in (\textbf{a}).
		The U.S., China, and France, the U.K., and Germany comprise about 70\% of all mobile researchers.
		\textbf{c.}
		The proportion of each country's researchers who are mobile.
		The dashed line indicates the proportion of all researchers in the data who are mobile.
		France, followed by Qatar and the U.S. have the highest proportion of mobile researchers.
		\textbf{d.} First two principal components of four variables: proportion of researchers in each country mobile across organizations, proportion mobile across cities, proportion mobile across regions, and proportion mobile across countries.
		The countries are roughly sorted in order of the number of mobile researchers and the fraction of international mobile researchers in the first and second principal components, which are indicated by PC1 and PC2, respectively.
		PC1 explains 88.3\% of the total variance, whereas PC2 explains 9.5\% of the total variance.
	}
	\label{fig:supp:descriptives}
\end{figure}

\newpage
%
%
\begin{figure}[p!]
	\centering
	\includegraphics[width=0.8\textwidth]{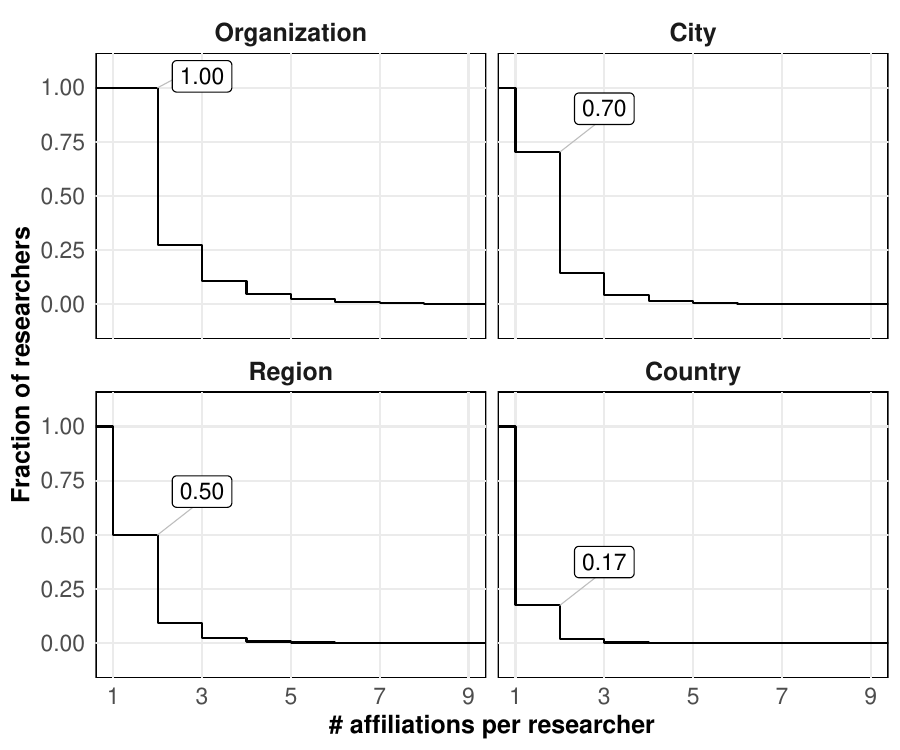}
	\caption{
		\textbf{Reverse cumulative-distribution function of mobile researchers by geographic scale.}
		\textbf{a.}
		Survival probability of mobile researchers with respect to the number of organizations in the their affiliation trajectory.
		All mobile authors were affiliated with at least two organizations (i.e., survival probability of one) and about 25.0\% were affiliated with three or more.
		\textbf{b.} About 70\% of mobile authors listed at least two cities represented in their career trajectories.
		\textbf{c.} 50\% of mobile authors have two or more regions represented in their career trajectories.
		\textbf{d.} Only 17\% of mobile authors had two or more countries represented in their career trajectories.
	}
	\label{fig:supp:rcdf_by_scale}
\end{figure}

\newpage
%
%
\begin{figure}[p!]
	\centering
	\includegraphics[width=0.8\textwidth]{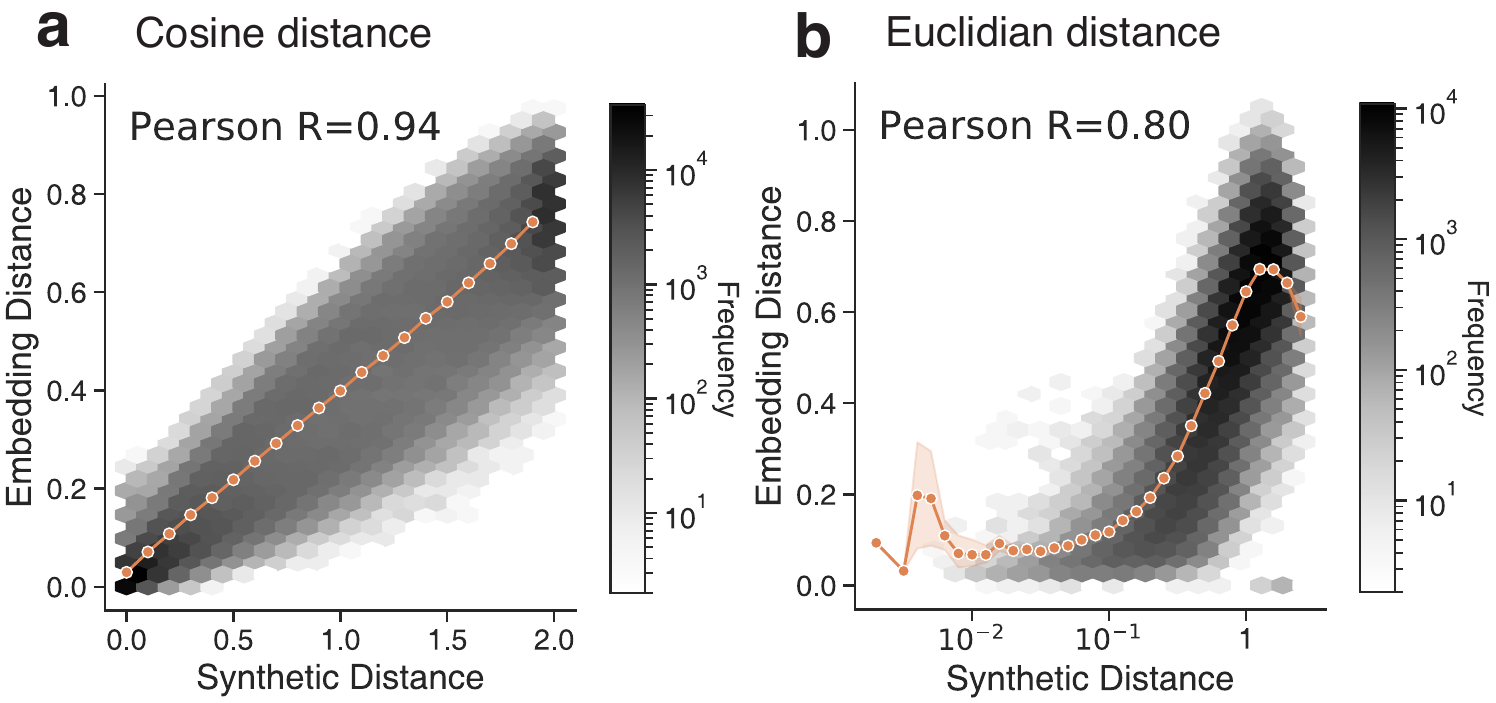}
	    \caption{\textbf{Neural embedding recovers the underlying distance from trajectories. }\textbf{a.} The trajectories generated with cosine distance an \textbf{b.} The trajectories generated with Euclidean distance. Illustratively, we show the result of the representative iteration of the experiment whose correlation is close to the average. The embedding distance shows a good agreement with underlying distance. Orange dots represent the average embedding distance across binned synthetic distances with standard error.
    }
	\label{fig:supp:synthetic_exp}
\end{figure}

\newpage
%
%
\begin{figure}[p!]
	\centering
	\includegraphics[width=0.8\textwidth]{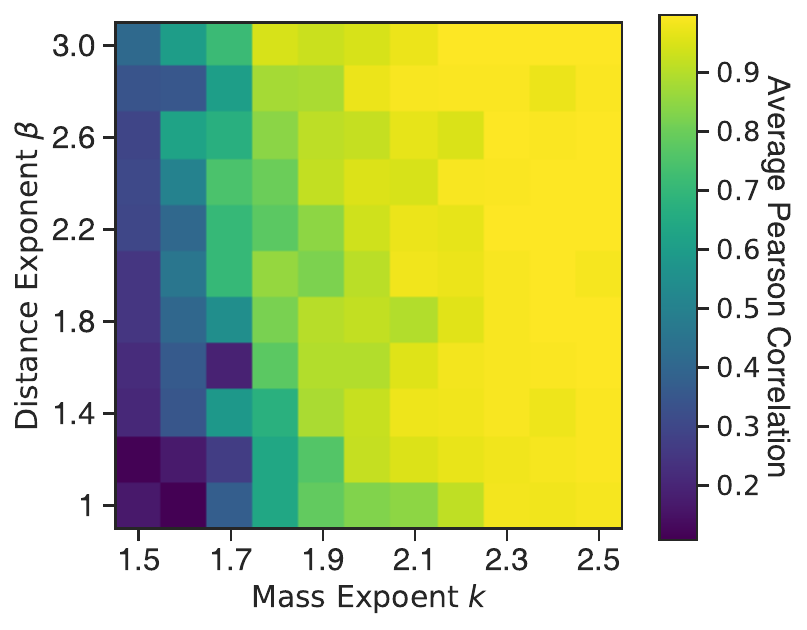}
	    \caption{\textbf{Neural embedding consistently recovers the underlying distance from trajectories across diverse hyperparameter settings.} We repeat our synthetic benchmark varying mass exponent $k$ and distance exponent $\beta$. The color of each cell represents the average Pearson correlation of the synthetic distance and embedding distances over ten repetitions with given $k$ and $\beta$.}
	\label{fig:supp:synthetic_robust}
\end{figure}

\begin{figure}[p!]
	\centering
	\includegraphics[width=\linewidth]{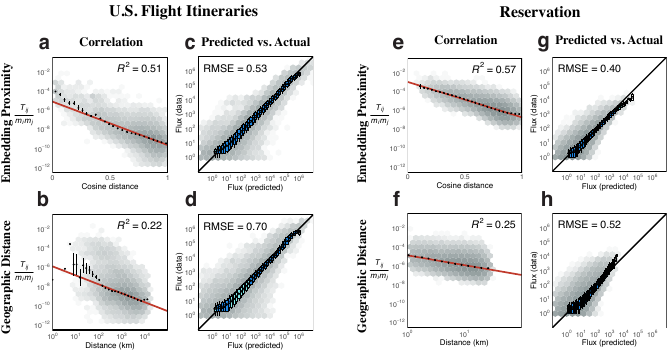}
	\caption{
		\textbf{Neural embedding provides functional distance that improves predictive power of the gravity model of migration best in distnct human trajectory datasets }
		 \textbf{a.} Embedding distance better explains the expected flux of passengers in U.S. flight itineraries than does geographic distance (\textbf{b}).
		The red line is the line of the best fit.
		Black dots are mean flux across binned distances.
		99\% confidence intervals are plotted for the mean flux in each bin.
		Correlation is calculated on the data in the log-log scale ($p < 0.0001$ across all fits).
		The lightness of each hex bin indicates the frequency of organization pairs within it.
		\textbf{d.} Predictions of flux between airport pairs made using embedding distance outperform those made using geographic distance (\textbf{e}).
		Box-plots show the distribution of actual flux for binned values of predicted flux.
		Box color corresponds to the degree to which the distribution overlaps with $y = x$.
		``RMSE'' is the root-mean-squared error between the actual and predicted values.
	Embedding distance consistently produces powerful functional distance for Korean accommodation reservations (\textbf{f-h})}
	\label{fig:supp:gravity_performance_si}
\end{figure}

%
%
\begin{figure}[p!]
	\centering
	\includegraphics[width=\textwidth]{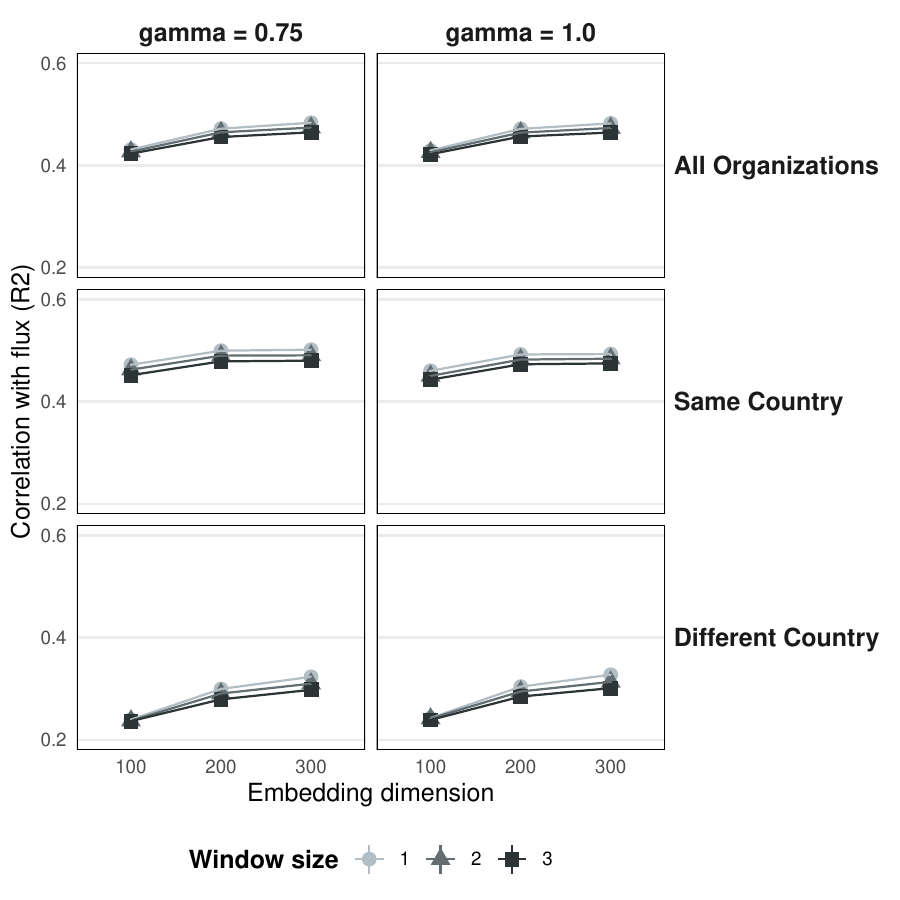}
	\caption{
		\textbf{Larger dimensions, smaller window size improves embedding performance.}
		The correlation, or amount of flux explained by the embedding distance with varying skip-gram negative sampling hyperparameters.
		Window size refers to $w$, the size of the context window that defines the context in a trajectory.
		Smaller window sizes result in an embedding that explains more flux.
		Embedding dimensions refer to the size of the embedding vector.
		Larger vectors perform better, though with little difference between 200 and 300.
		Gamma refers to the $\gamma$ parameter in \textit{word2vec}, which shapes the negative sampling distribution.
		A value of $\gamma = 0.75$ is the default for \textit{word2vec}.
		There is virtually no difference in performance based on $\gamma$.
		All variants perform better on same-country organization pairs, and worse on different-country pairs, than on all pairs of  of organizations than on all pairs.
		Embeddings with larger dimensions outperform mid-size embeddings for the different-country case.
	}

	\label{fig:supp:hyperparameter}
\end{figure}

%
%
\begin{figure}[ht!]
	\centering
	\includegraphics[width=0.95\textwidth]{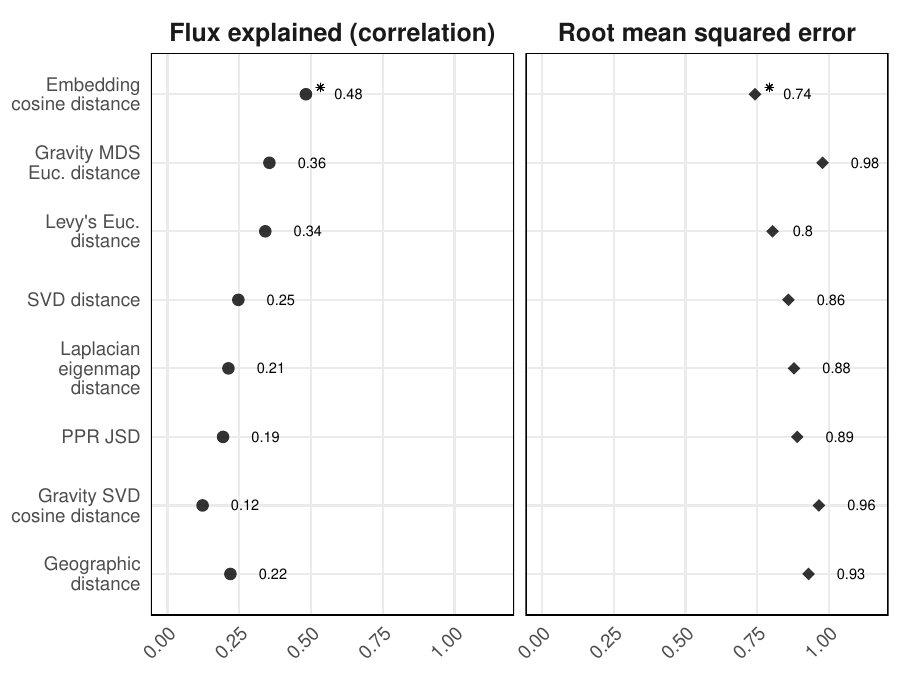}
	\caption{
	\textbf{Neural embeddings outperform baselines for scientific mobility.}
		Cosine distance between embedding vectors generated with \textit{word2vec} explains more of the flux, and better predicts flux when used in the gravity model of mobility than geographic distance and other baselines.
		Shown is the correlation between the flux and embedding distances, measured with $R^{2}$ (left), and the prediction error when using the distance as input to the gravity model of mobility.
    The asterisk denotes the top-performing metric.
		For prediction error, we show results based on both the exponential and power-law forms of the gravity model.
		All embedding-based methods use dimensions of 300.
		Here, embedding distance is obtained from neural embeddings learned with window size of 1 and $\gamma = 1$.
		In all cases, organization population is defined as the mean annualized number of unique mobile and non-mobile authors, and flux is calculated for all global mobility.
		Baselines include the top-performing distance metrics calculated between vectors obtained by personalized-page rank (PPR), singular value decomposition (SVD), laplacian eigenmap, direct-factorization following Levy's approach~\autocite{levy2014neural}, and direct optimization of the gravity model using SVD and multidimensional scaling (MDS), as well as the geographic distance between organizations.
		Embedding distance better explains and predicts flux than any other baseline, though there is some variation by experimental parameters (Table~\ref{supp:table:r2_table}, Table~\ref{supp:table:rmse_exp_table}, and Table~\ref{supp:table:rmse_power_table}).
	}
	\label{fig:supp:distancemetrics}
\end{figure}

%
%
\begin{figure}[p!]
	\centering
	\includegraphics[width=\textwidth]{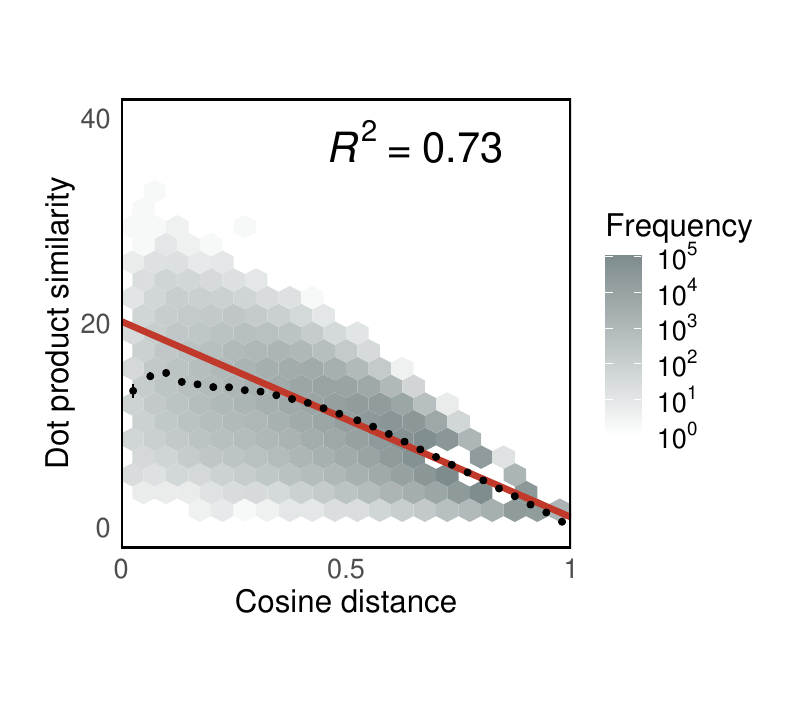}
	\caption{
		\textbf{Cosine distance is correlated with dot product similarity.}
		We find a relatively high correlation between the embedding distance---one minus the cosine similarity---and the dot product similarity between organization vectors ($R^2 = 0.73$).
		Color of each hex bin indicates the frequency of organization pairs.
				The red line is the line of the best fit.
		Black dots are mean flux across binned distances.
		99\% confidence intervals are plotted for the mean flux in each bin based on a normal distribution.
		Correlation is calculated on the data in the log-log scale ($p < 0.0001$).
	}
	\label{fig:supp:cosdot}
\end{figure}

%
%
\begin{figure}[p!]
	\centering
	\includegraphics[width=\textwidth]{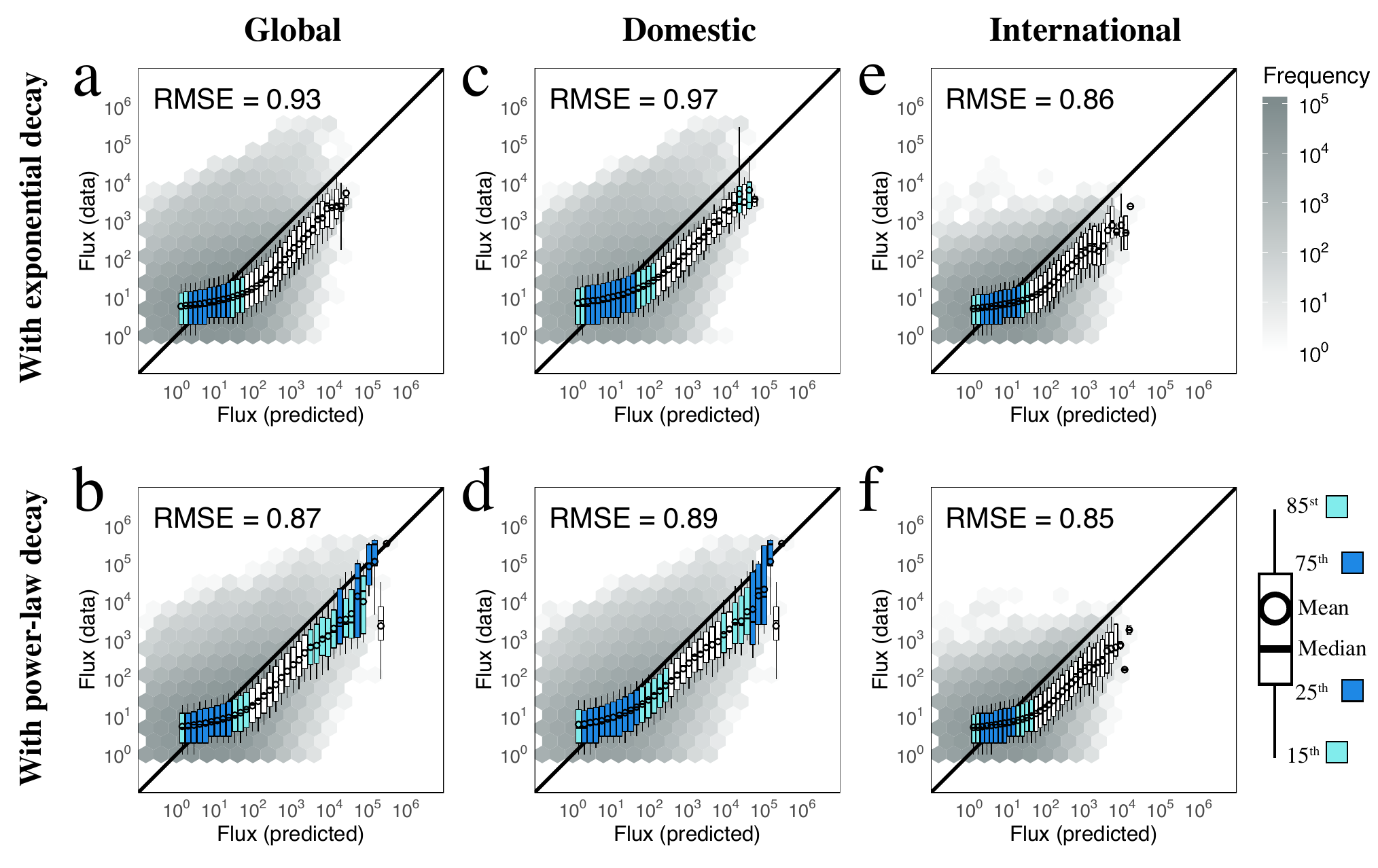}
	\caption{
		\textbf{For geographic distance, the power-decay gravity model is better.}
		Flux between organization pairs predicted by the gravity model with different distance decay functions, i.e., exponential decay function (\textbf{a}) and power-law decay function (\textbf{b}) using geographic distance.
		Boxplots show distribution of actual flux for binned values of predicted flux.
		Box color corresponds to the degree to which the distribution overlaps with $x = y$;
		a perfect prediction yields all points on the black line.
		``RMSE'' is the root-mean-squared error between the actual and predicted values.
		Shown for all pairs of organization (\textbf{a-b}), domestic (\textbf{c-d}), and international only (\textbf{e-f}) mobility.
		The gravity model with the power-decay function outperforms that with an exponential decay function.
	}
	\label{fig:supp:predict_geo_bymodel}
\end{figure}

%
%
\begin{figure}[p!]
	\centering
	\includegraphics[width=\textwidth]{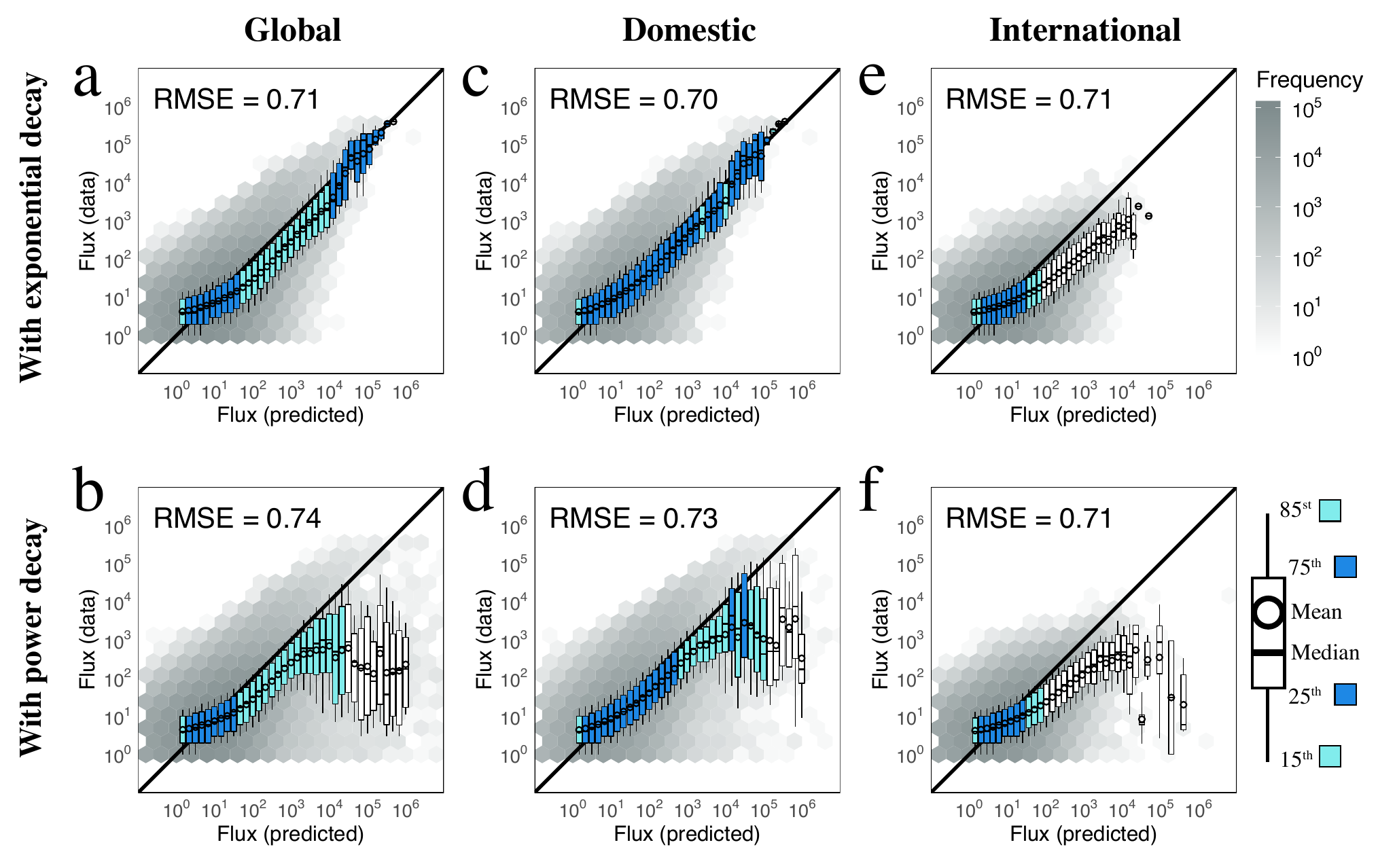}
	\caption{
		\textbf{For embedding distance, the exponential-decay gravity model is slightly better.}
		Flux between organization pairs predicted by the gravity model with different distance decay functions, i.e., exponential decay function (\textbf{a}) and power-law decay function (\textbf{b}) using embedding distance.
		Boxplots show the distribution of actual flux for binned values of predicted flux.
		Box color corresponds to the degree to which the distribution overlaps with $x = y$;
		a perfect prediction yields all points on the black line.
		``RMSE'' is the root-mean-squared error between the actual and predicted values.
		Shown for all pairs of organization (\textbf{a-b}), domestic (\textbf{c-d}), and international only (\textbf{e-f}) mobility.
		The gravity model with the exponential decay function slightly outperforms that with a power-decay function except in the case of international-only mobility, for which power-decay performs slightly better.
	}
	\label{fig:supp:predict_emb_bymodel}
\end{figure}

%
%
\begin{figure}[p!]
	\centering
	\includegraphics[width=\textwidth]{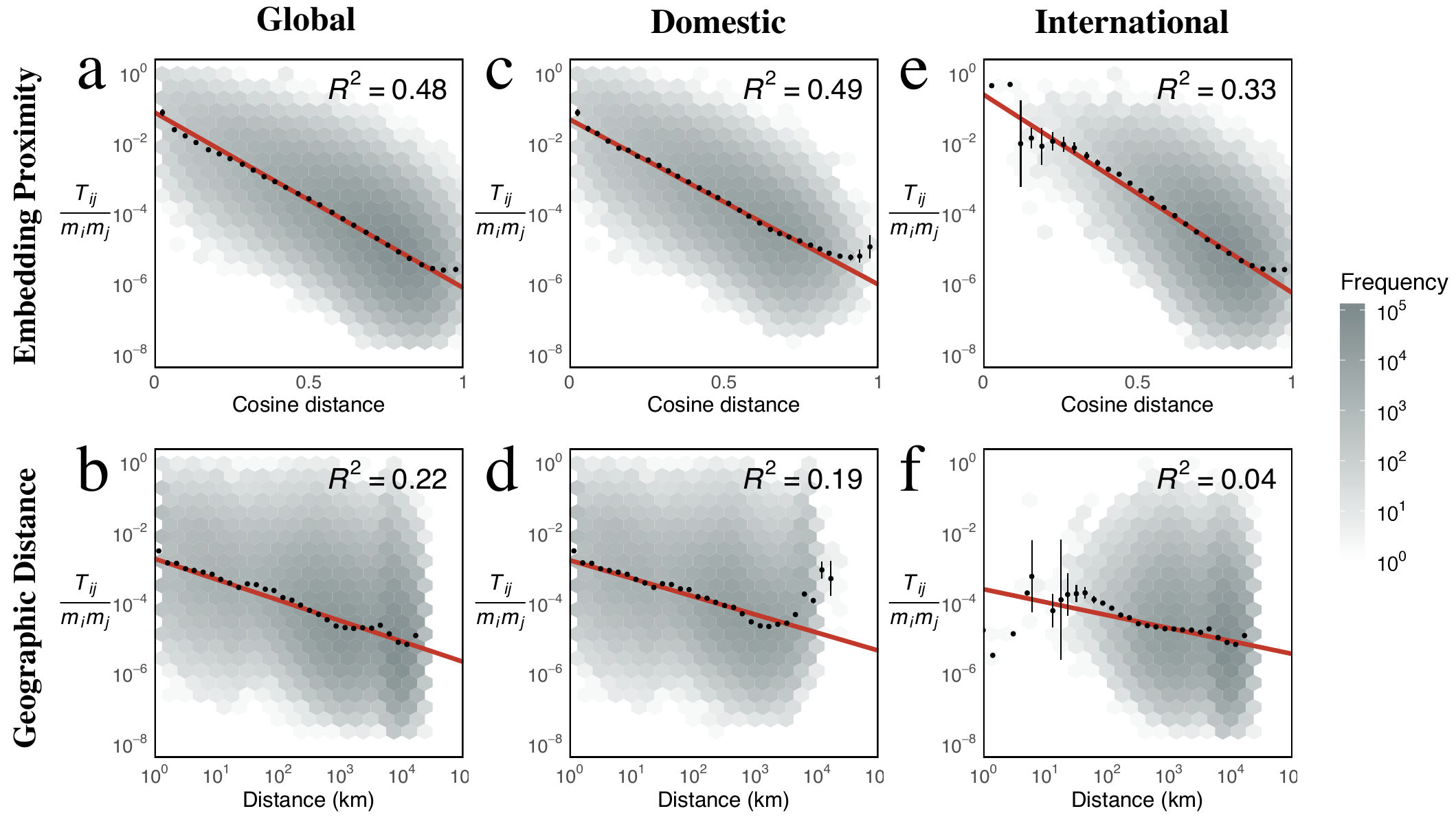}
	\caption{
		\textbf{Embedding distance explains more variance for global, within, and across country flux than geographic distance.}
		\textbf{a.}
		Embedding distance explains more flux than geographic distance (\textbf{b}).
		The red line is the line of the best fit.
		Black dots are mean flux across binned distances.
		99\% confidence intervals are plotted for the mean flux in each bin based on a normal distribution.
		Correlation is calculated on the data in the log-log scale ($p < 0.0001$ across all fits).
		Color of each hex bin indicates frequency of organization pairs.
		Results here are identical to those shown in manuscript.
		\textbf{c-d.}	embedding distance explains more variance when considering only within-country organization pairs.
		\textbf{e-f.} embedding distance is more robust than geographic distance when considering only across-country organization pairs.}
	\label{fig:supp:gravity_correlation}
\end{figure}

%
%
\begin{figure}[p!]
	\centering
	\includegraphics[width=\textwidth]{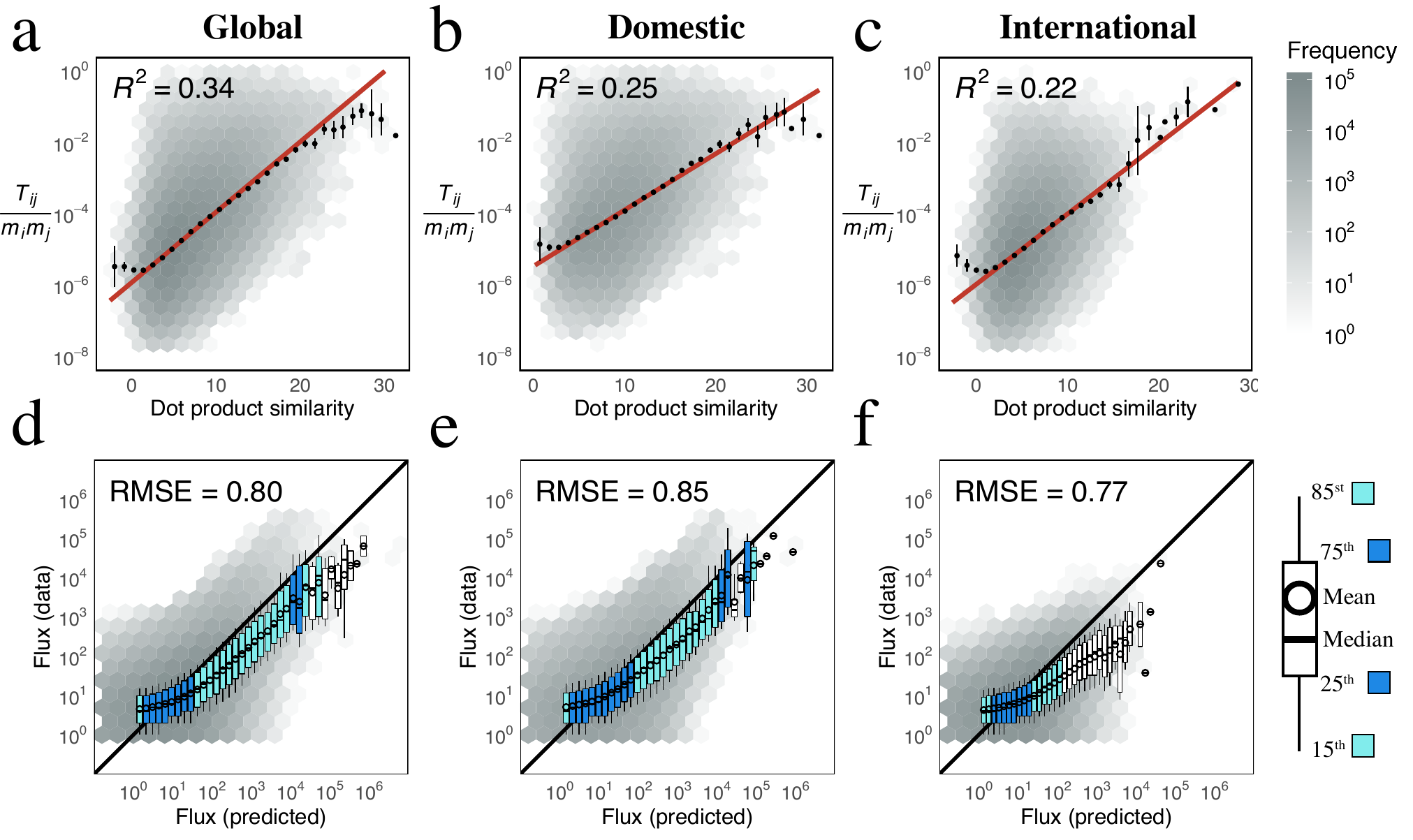}
	\caption{
		\textbf{Examine gravity model with dot product on the embedding space.}
		Performance of dot product similarities in explaining and predicting mobility.
		Similarity scores are calculated as the pairwise dot product between organizational vectors.
		Dot product similarity performs better than geographic distance, though worse than cosine similarity in explaining global mobility (\textbf{a}), or domestic (\textbf{b}) or international (\textbf{c}) country mobility.
		The red line is the line of the best fit.
		Black dots are mean flux across binned distances.
		99\% confidence intervals are plotted for the mean flux in each bin based on a normal distribution.
		Correlation is calculated on the data in the log-log scale ($p < 0.0001$ across all fits).
		Color indicates frequency of organization pairs within each hex bin.
		Similarly, PPR distance performs comparably to geographic distance in predicting global (\textbf{d}), domestic (\textit{e}) and international (\textbf{f}) scientific mobility.
		Boxplots show distribution of actual flux for binned values of predicted flux.
		Box color corresponds to the degree to which the distribution overlaps $x = y$;
		a perfect prediction yields all points on the black line.
		``RMSE'' is the root-mean-squared error between the actual and predicted values.
	}
	\label{fig:supp:gravity_dot}
\end{figure}

%
%
\begin{figure}[p!]
	\centering
	\includegraphics[width=\textwidth]{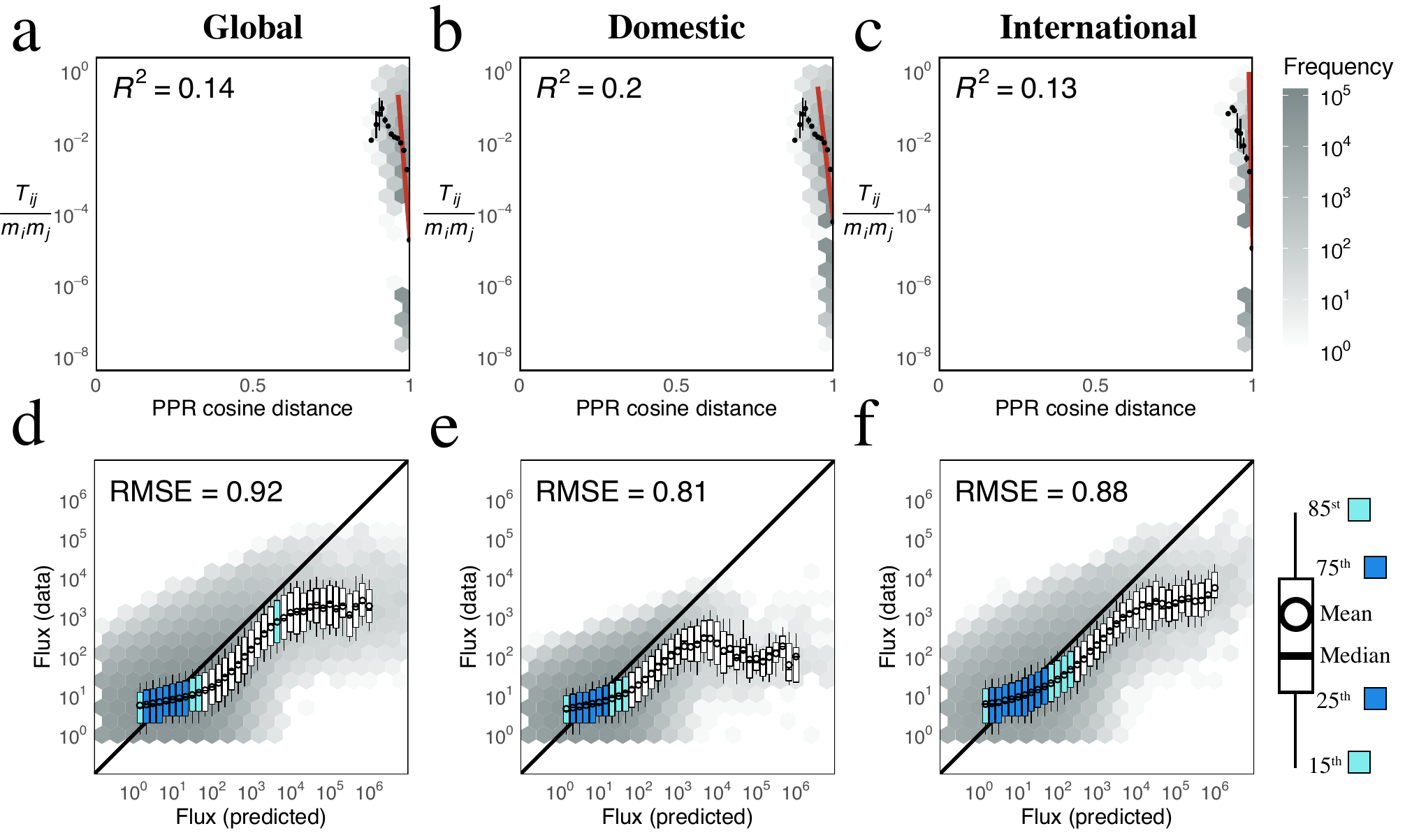}
	\caption{
		\textbf{Personalized page rank with cosine distance.}
		Performance of personalized page rank scores in explaining and predicting mobility.
		Personalized page rank is calculated for the underlying mobility network, and distance measured as the cosine distance between PPR probability distribution vectors.
		PPR cosine distance performs roughly similar to geographic distance in explaining global(\textbf{a}), domestic (\textbf{b}), or international (\textbf{c}) country mobility.
		The red line is the line of the best fit.
		Black dots are mean flux across binned distances.
		99\% confidence intervals are plotted for the mean flux in each bin based on a normal distribution.
		Correlation is calculated on the data in the log-log scale ($p < 0.0001$ across all fits).
		Color of hex bind indicates frequency of organization pairs.
		Similarly, PPR distance performs comparably to geographic distance in predicting global (\textbf{d}), domestic (\textit{e}) and international (\textbf{f}) scientific mobility.
		Boxplots show distribution of actual flux for binned values of predicted flux.
		Box color corresponds to the degree to which the distribution overlaps $x = y$;
		a perfect prediction yields all points on the black line.
		``RMSE'' is the root-mean-squared error between the actual and predicted values.
	}
	\label{fig:supp:gravity_pprcos}
\end{figure}

%
%

\begin{figure}[p!]
	\centering
	\includegraphics[width=\textwidth]{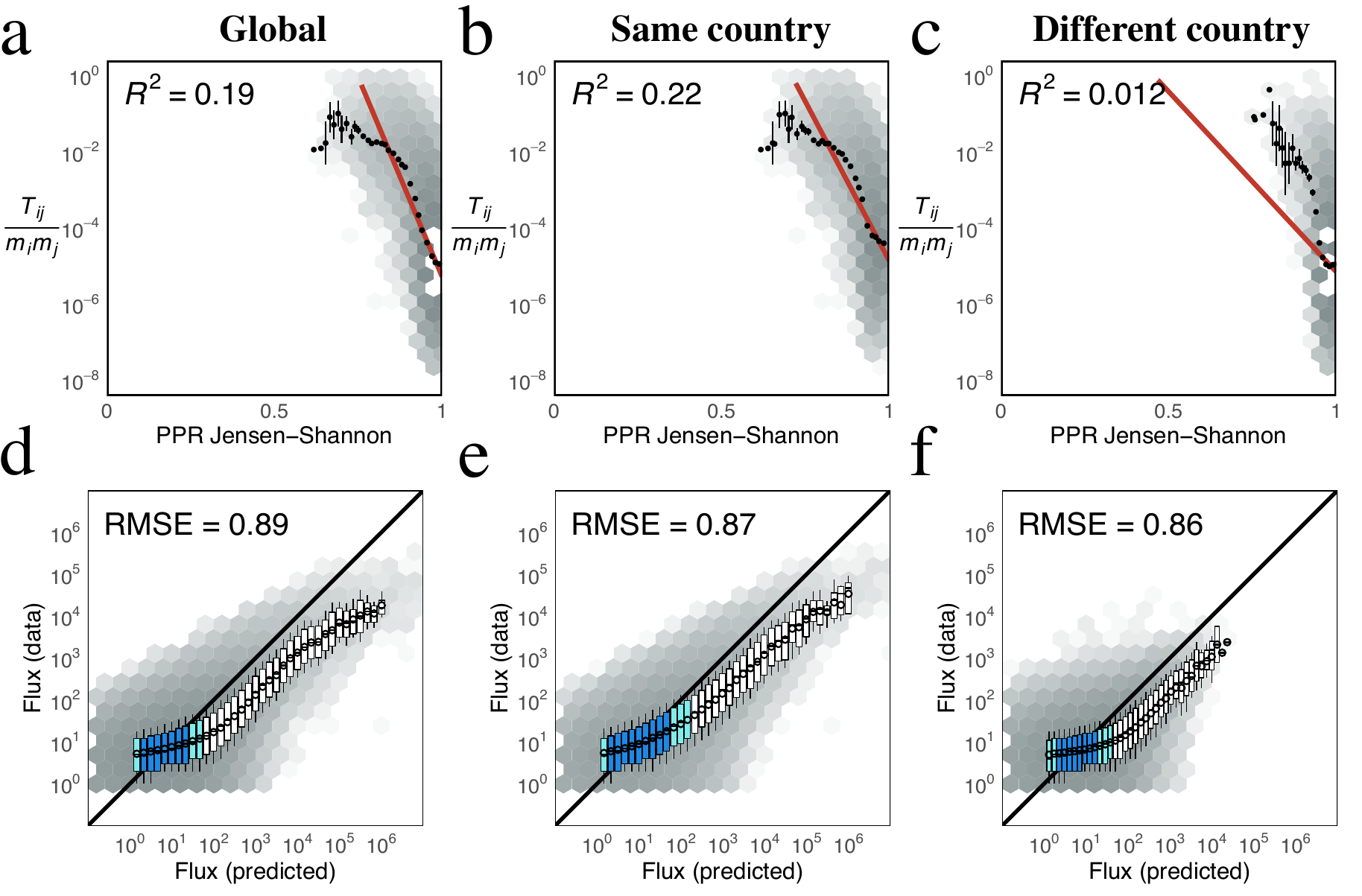}
	\caption{
		\textbf{Personalized page rank with Jensen-Shannon Divergence.}
		Performance of personalized page rank scores in explaining and predicting mobility.
		Personalized page rank is calculated for the underlying mobility network, and distance measured as the Jensen-Shannon Divergence (JSD) between PPR probability distribution vectors.
		PPR JSD performs roughly similar to geographic distance in explaining global mobility (\textbf{a}), or domestic (\textbf{b}) or international (\textbf{c}) country mobility.
		Overall, PPR JSD explains more variance in mobility than using cosine distance (Fig.~\ref{fig:supp:gravity_pprcos}), except for international mobility, for which cosine similarity out-performs JSD.
		The red line is the line of the best fit.
		Black dots are mean flux across binned distances.
		99\% confidence intervals are plotted for the mean flux in each bin based on a normal distribution.
		Correlation is calculated on the data in the log-log scale ($p < 0.0001$ across all fits).
		Color of hex bind indicates frequency of organization pairs.
		Similarly, PPR JSD performs comparably to geographic distance in predicting global (\textbf{d}), domestic (\textit{e}) and international (\textbf{f}) scientific mobility.
		Boxplots show distribution of actual flux for binned values of predicted flux.
		Box color corresponds to the degree to which the distribution overlaps $x = y$;
		a perfect prediction yields all points on the black line.
		``RMSE'' is the root-mean-squared error between the actual and predicted values.
	}
	\label{fig:supp:gravity_pprjsd}
\end{figure}

%
%
\begin{figure}
    \centering
    \includegraphics{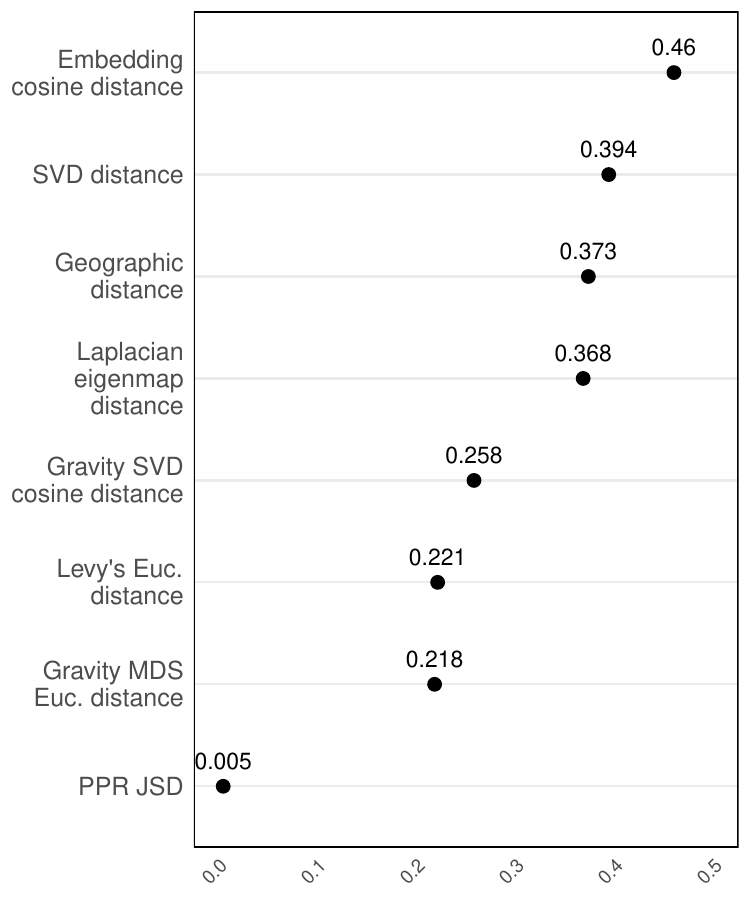}
    \caption{\textbf{Embedding distance outperforms alternatives in common part of commuters (CPC).}
    The CPC measure is calculated for both the power and exponential form of the gravity model, with the highest value reported for each distance type. 
    }
    \label{fig:supp:cpc-performance}
\end{figure}

%
%
\begin{figure}[hp!]
	\centering
	\includegraphics[width=\textwidth]{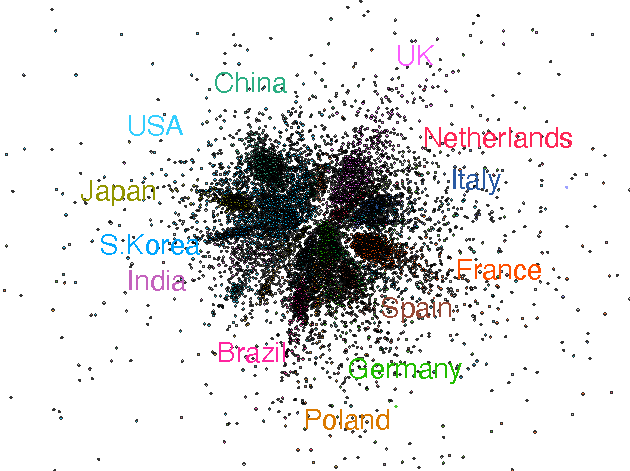}
	\caption{
		\textbf{Visualization of global mobility network.}
		The network demonstrates country-level structure, but not at the detail or the extent of the global UMAP projection (Fig. 2a).
		Each node corresponds to an organization, whereas  weighted edges (not shown) correspond to the flow of mobile researchers between the two organization.
		Nodes are colored by the country of the organization.
		Nodes are positioned using the Force Atlas layout algorithm.
	}
	\label{fig:supp:network_vis}
\end{figure}

%
%
\begin{figure}[hp!]
	\centering
	\includegraphics[width=\textwidth]{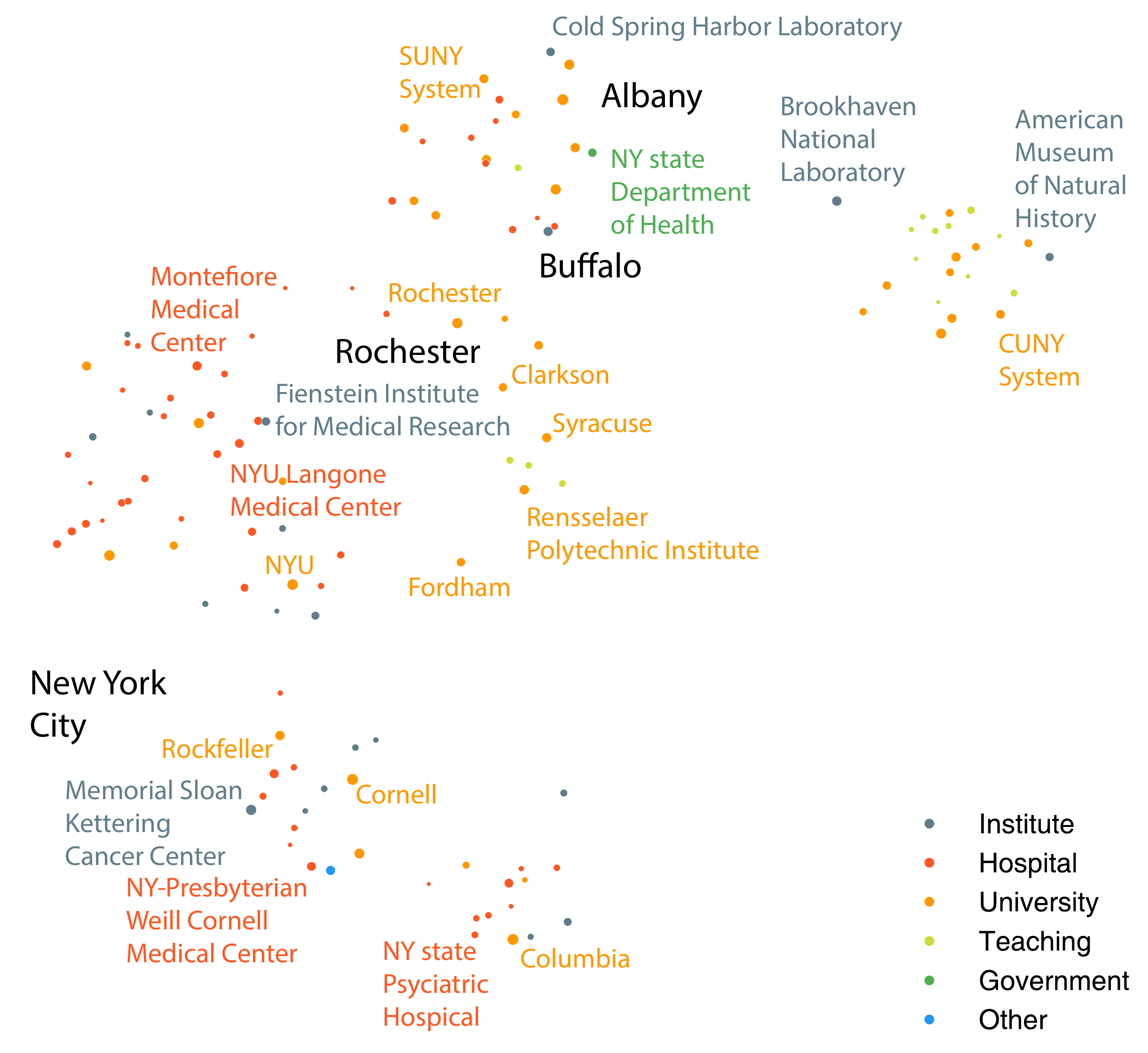}
	\caption{
		\textbf{UMAP Projection of organizations in New York.}
		Each point corresponds to an organization and its size indicates the average annual number of mobile and non-mobile authors affiliated with that organization from 2008 to 2019.
		Color indicates the sector.
	}
	\label{fig:supp:proj_newyork}
\end{figure}

%
%
\begin{figure}[hp!]
	\centering
	\includegraphics[width=\textwidth]{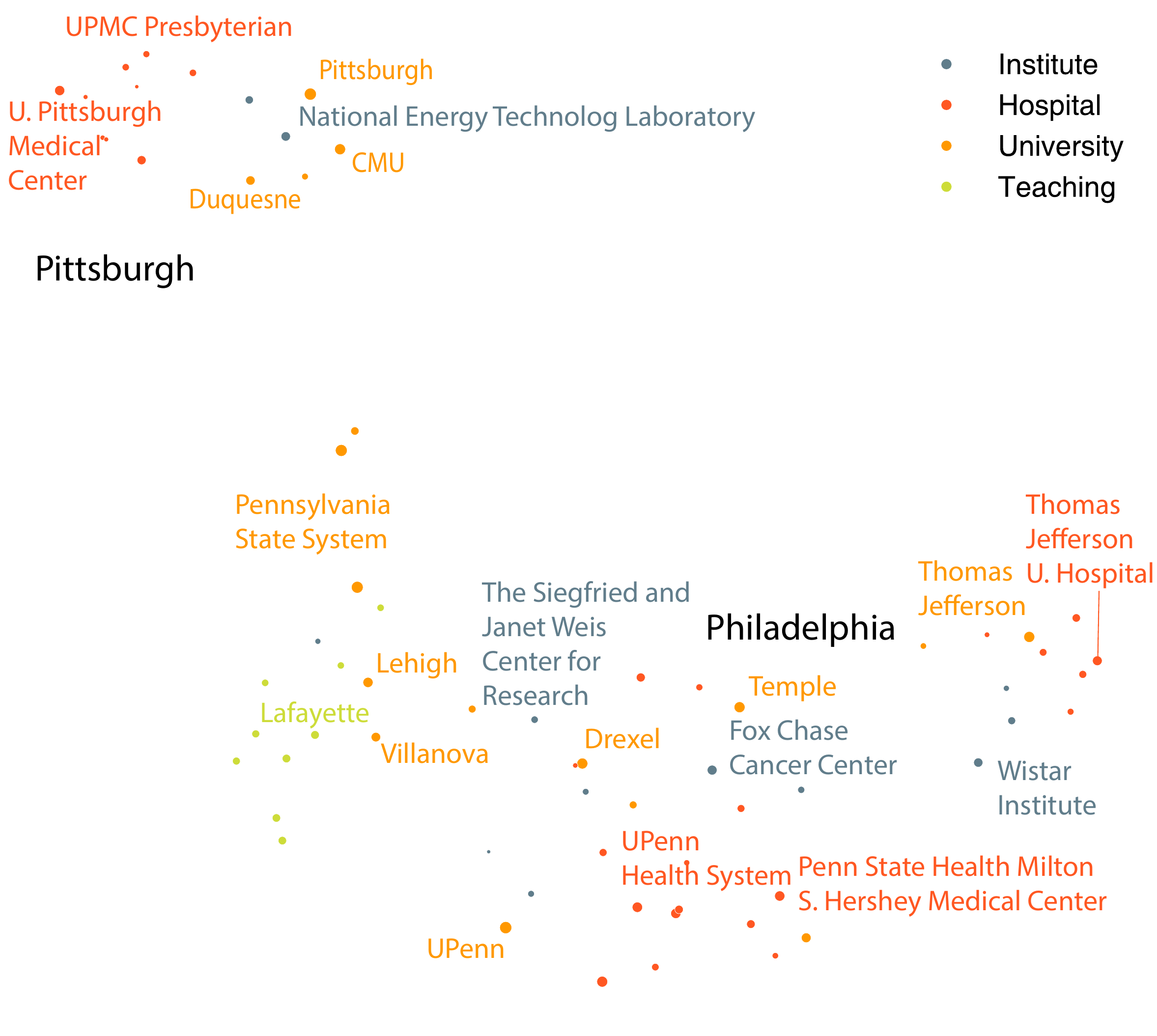}
	\caption{
		\textbf{UMAP Projection of organizations in Pennsylvania.}
		UMAP projection of the embedding space of organizations in Pennsylvania reveals clustering based on geography, sector, and academic prestige.
		Each point corresponds to an organization and its size indicates the average annual number of mobile and non-mobile authors affiliated with that organization from 2008 to 2019.
		Color indicates the sector.
	}
	\label{fig:supp:proj_pennsylvania}
\end{figure}

%
%
\begin{figure}[hp!]
	\centering
	\includegraphics[width=\textwidth]{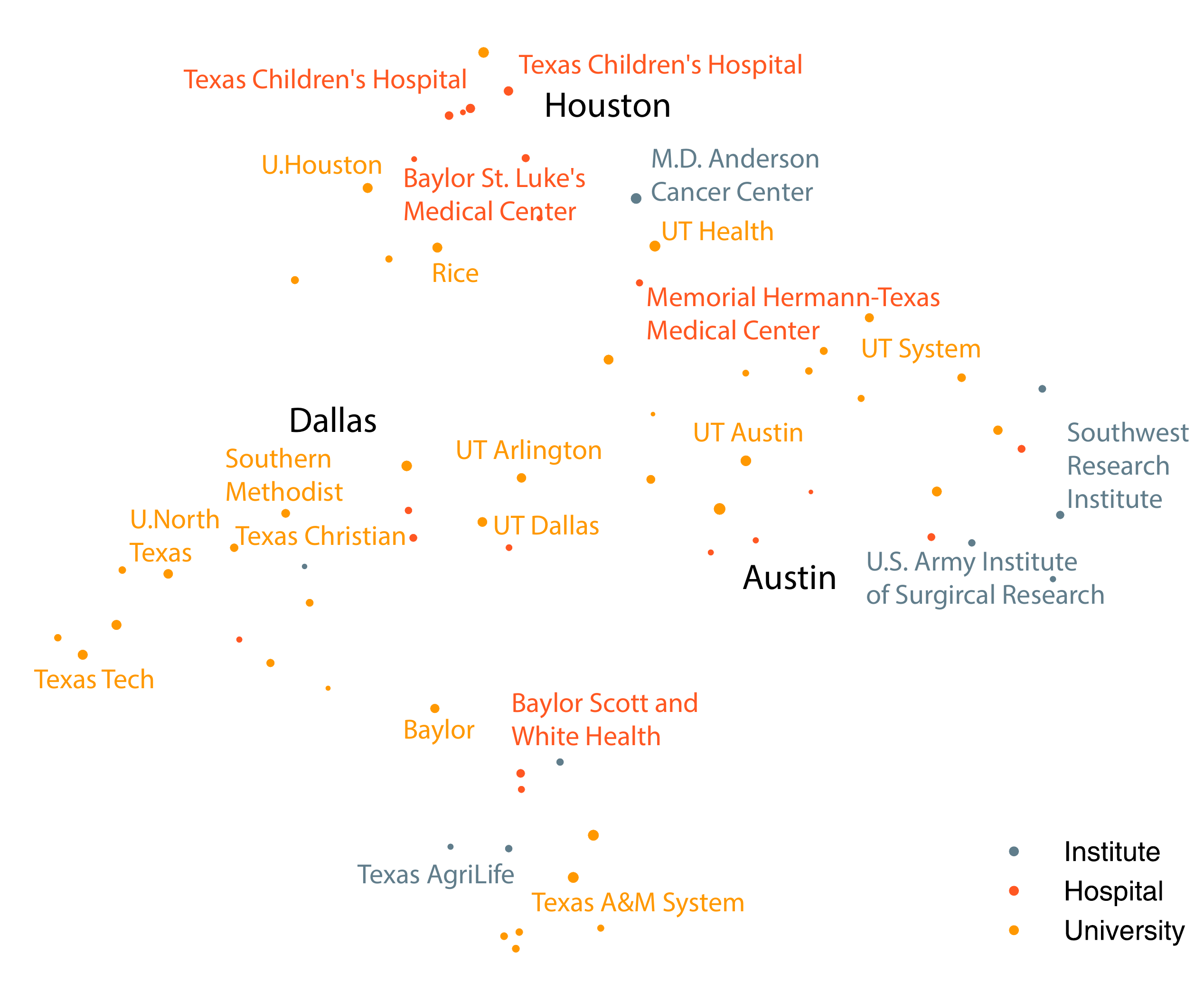}
	\caption{
		\textbf{UMAP Projection of organizations in Texas.}
		Each point corresponds to an organization and its size indicates the average annual number of mobile and non-mobile authors affiliated with that organization from 2008 to 2019.
		Color indicates the sector.
	}
	\label{fig:supp:proj_texas}
\end{figure}

%
%
\begin{figure}[hp!]
	\centering
	\includegraphics[width=\textwidth]{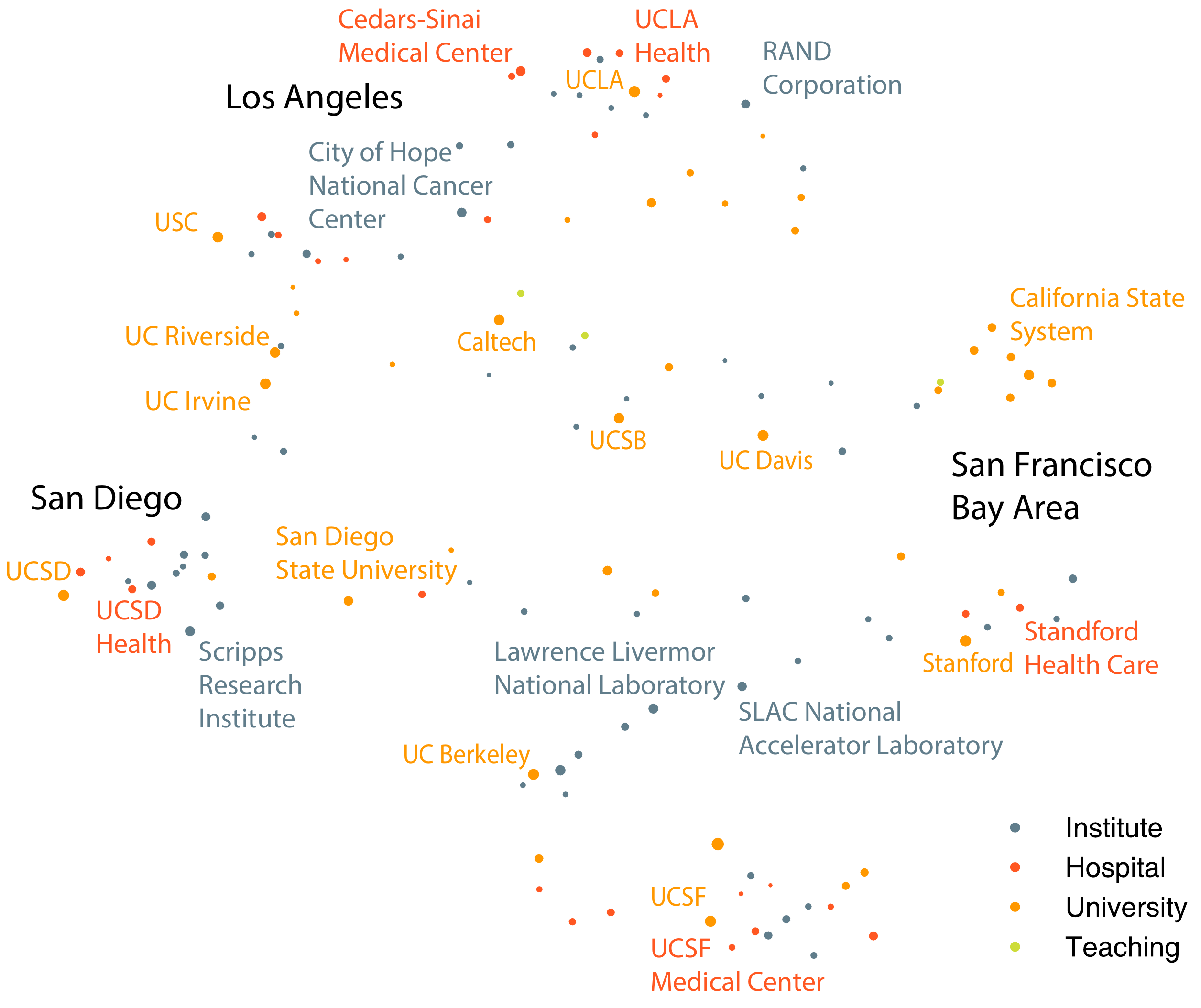}
	\caption{
		\textbf{UMAP Projection of organizations in California.}
		Each point corresponds to an organization and its size indicates the average annual number of mobile and non-mobile authors affiliated with that organization from 2008 to 2019.
		Color indicates the sector.
	}
	\label{fig:supp:proj_california}
\end{figure}

%
%
%
%

%
%
\begin{figure}[hp!]
	\centering
	\includegraphics[width=\textwidth]{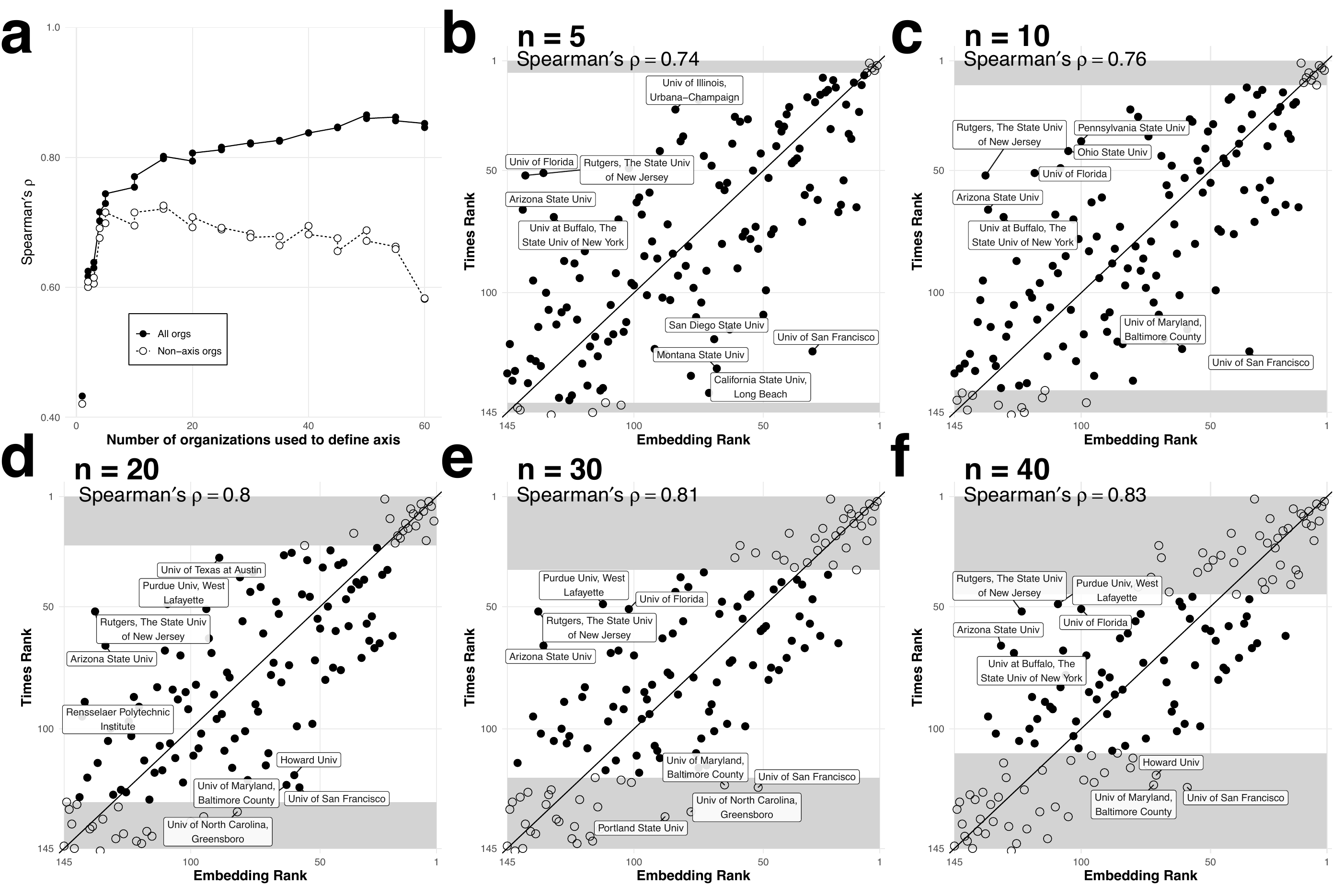}
	\caption{
		\textbf{SemRank hierarchy is robust.}
		\textbf{a.} Spearman's $\rho$ ($n = 143$) between Times prestige rank and embedding rank derived using SemAxis, with poles defined using the top and bottom (geographically matched) ranked universities.
		Black points show spearman correlation using all organizations; white points show correlation using only universalizes not aggregated in the poles.
		Including more universities improves performance, but quickly saturates after around five universities.
		\textbf{b - f.} Comparison between the Times and SemAxis ranks of universities, by the number of universities used to define the poles (n).
		White points are those top and bottom 20 universities aggregated to define the ends of the axis.
		The grey box corresponds to the top 20 and bottom 20 ranks.
	  	Spearman's $\rho$ details the estimate from Spearman correlation between the two rankings using all universities, including those used to define the ends of each axis.
	  	All correlations are significant with $p < 0.0001$.
	}
	\label{fig:supp:semaxis_compare}
\end{figure}

%
%
\begin{figure}[hp!]
	\centering
	\includegraphics[width=\textwidth]{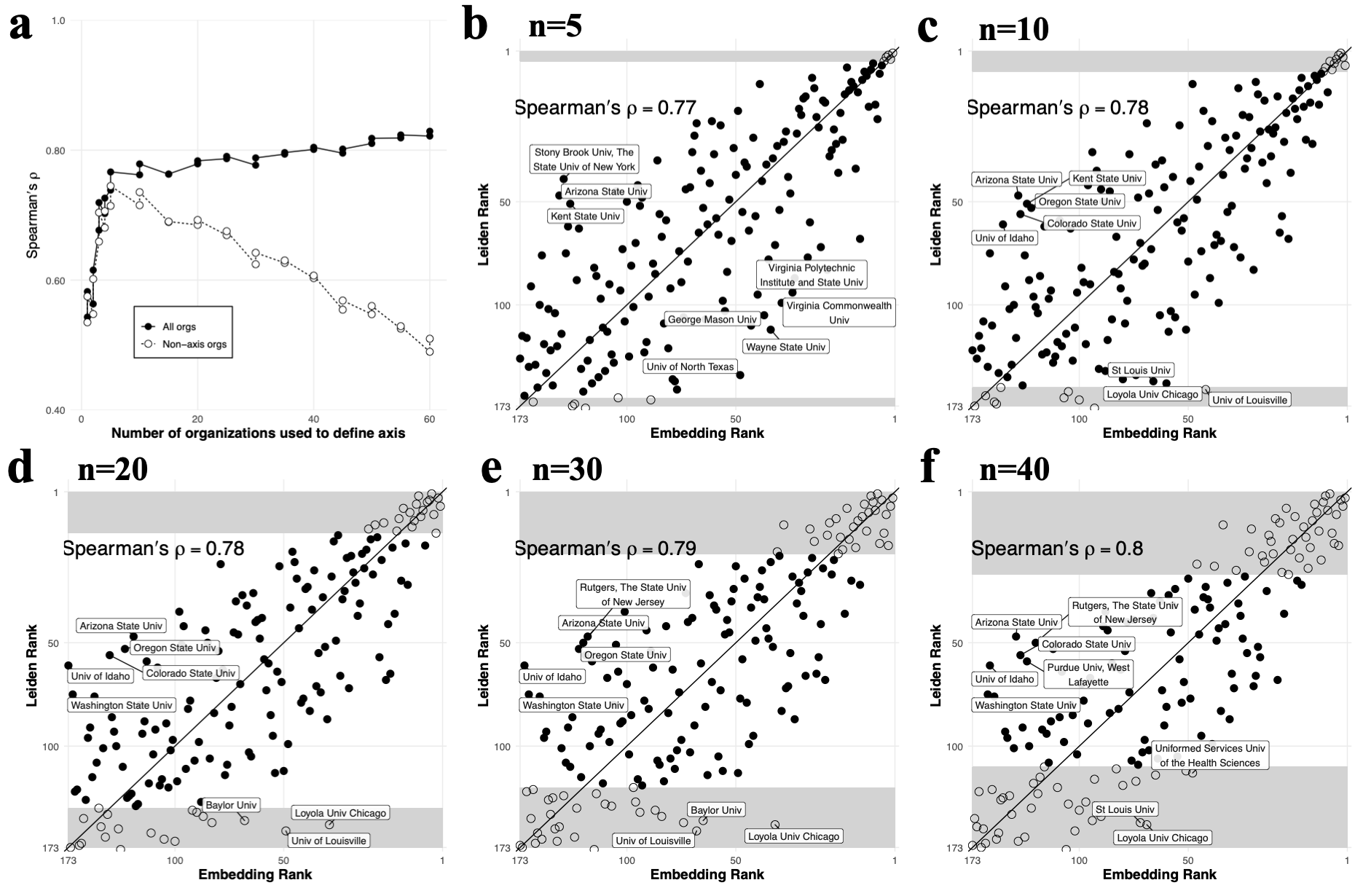}
	\caption{
		\textbf{SemRank hierarchy holds when compared to Leiden Ranking.}
		\textbf{a.} Spearman's $\rho$ ($n = 143$) between Leiden ranking of U.S. universities derived using SemAxis, with poles defined using the top and bottom (geographically matched) ranked universities.
		Black points show Spearman correlation using all organizations; white points show correlation using only universities not aggregated in the poles.
		Including more universities improves performance, but quickly saturates after around five universities.
		\textbf{b - f.} Comparison between the Leiden and SemAxis ranks of universities, by the number of universities used to define the poles (n).
		White points are the top and bottom 20 universities aggregated to define the ends of the axis.
		The grey box corresponds to the top 20 and bottom 20 ranks.
	  	Spearman's $\rho$ details the estimate from Spearman's rank correlation between the two rankings using all universities, including those used to define the ends of each axis.
	  	All correlations are significant with $p < 0.0001$.
	}
	\label{fig:supp:semaxis_compare_leiden}
\end{figure}

%
%
\begin{figure}[hp!]
	\centering
	\includegraphics[width=\textwidth]{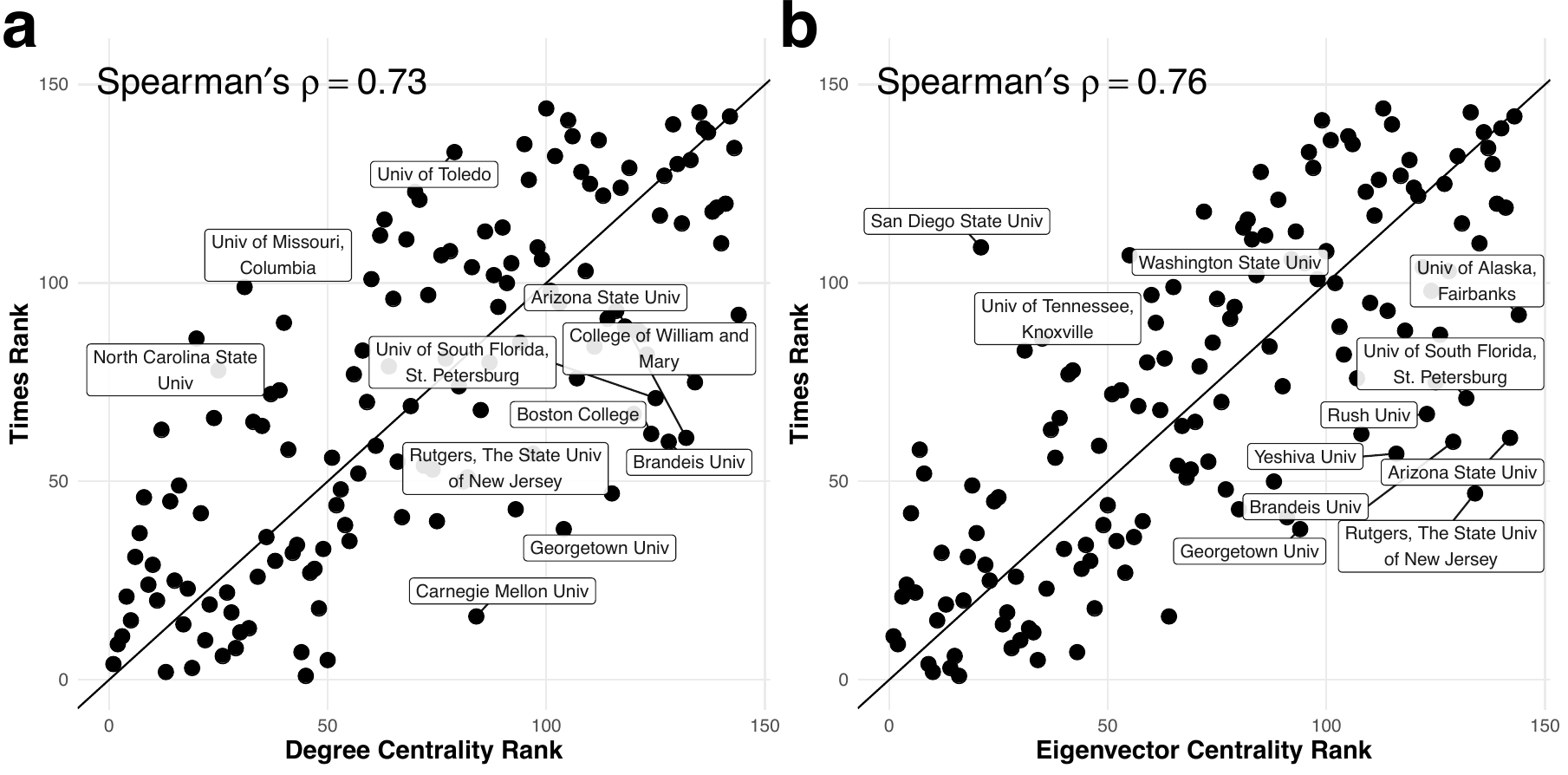}
	\caption{
		\textbf{Network centrality is strongly correlated with Times ranking.}
		Comparison between the ranking of organizations by their network-centrality rank and their rank in the 2018 Times Higher Education ranking of U.S. Universities .
		The Times rank is correlated with degree centrality rank (\textbf{a}) with Spearman's $\rho = 0.73$, and is correlated with the eigenvector centrality rank (\textbf{b}) with Spearman's $\rho = 0.76$.
		All correlations are significant with $p < 0.0001$.
	}
	\label{fig:supp:centrality_times_compare}
\end{figure}

%
%
\begin{figure}[hp!]
	\centering
	\includegraphics[width=\textwidth]{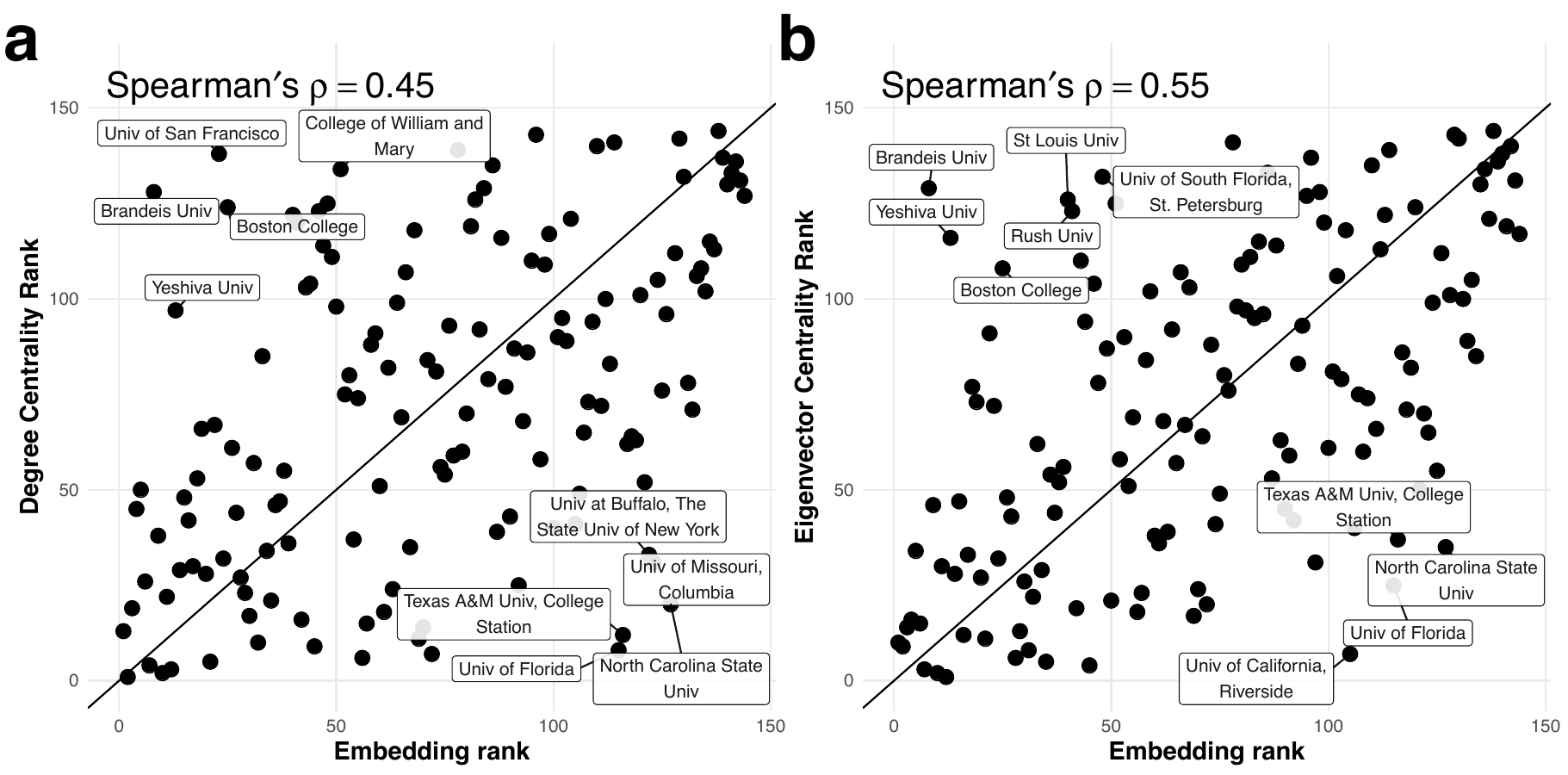}
	\caption{
		\textbf{Network centrality less correlated with Embedding rank.}
		Comparison between the ranking of organizations by their network-centrality rank and the embedding rank derived with SemAxis with poles defined using the top five to geographically-matched bottom five universities ranked by the 2018 Times Higher Education ranking of U.S. Universities .
		Embedding rank is correlated with degree centrality rank (\textbf{a}) with Spearman's $\rho = 0.45$, and is correlated with the eigenvector centrality rank (\textbf{b}) with Spearman's $\rho = 0.55$.
		All correlations are significant with $p < 0.0001$.
	}
	\label{fig:supp:centrality_semaxis_compare}
\end{figure}

%
%
\begin{figure}[hp!]
	\centering
	\includegraphics[width=\textwidth]{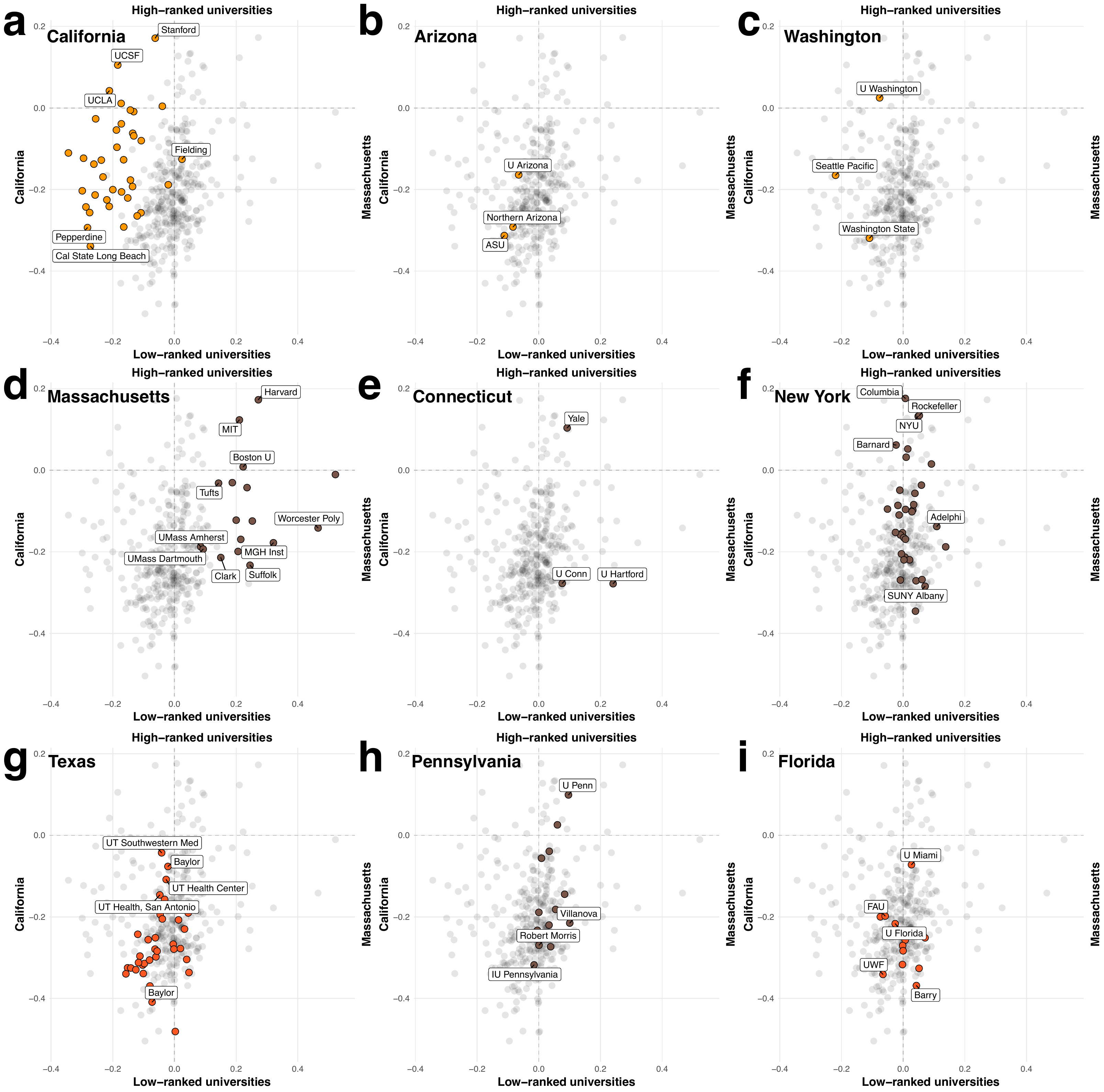}
	\caption{
		\textbf{Geography and prestige SemAxis by U.S. state.}
		SemAxis projection along two axes, comparing California to Massachusetts universities (left to right), and between the top 20 and geographically-matched bottom 20 universities ranked by the 2018 Times Higher Education ranking of U.S. Universities (bottom to top).
		Points correspond to universities shown for California (\textbf{a}), Arizona (\textbf{b}), Washington (\textbf{c}), Massachusetts (\textbf{d}), Connecticut (\textbf{e}), New York (\textbf{f}), Texas (\textbf{g}), Pennsylvania (\textbf{h}), and Florida (\textbf{i}).
		Grey points correspond to all other U.S. universities.
		Full organization names listed in Table~\ref{table:supp:orglabels}.
	}
	\label{fig:supp:semaxis_states}
\end{figure}

%
%
\begin{figure}[hp!]
	\centering
	\includegraphics[width=\textwidth]{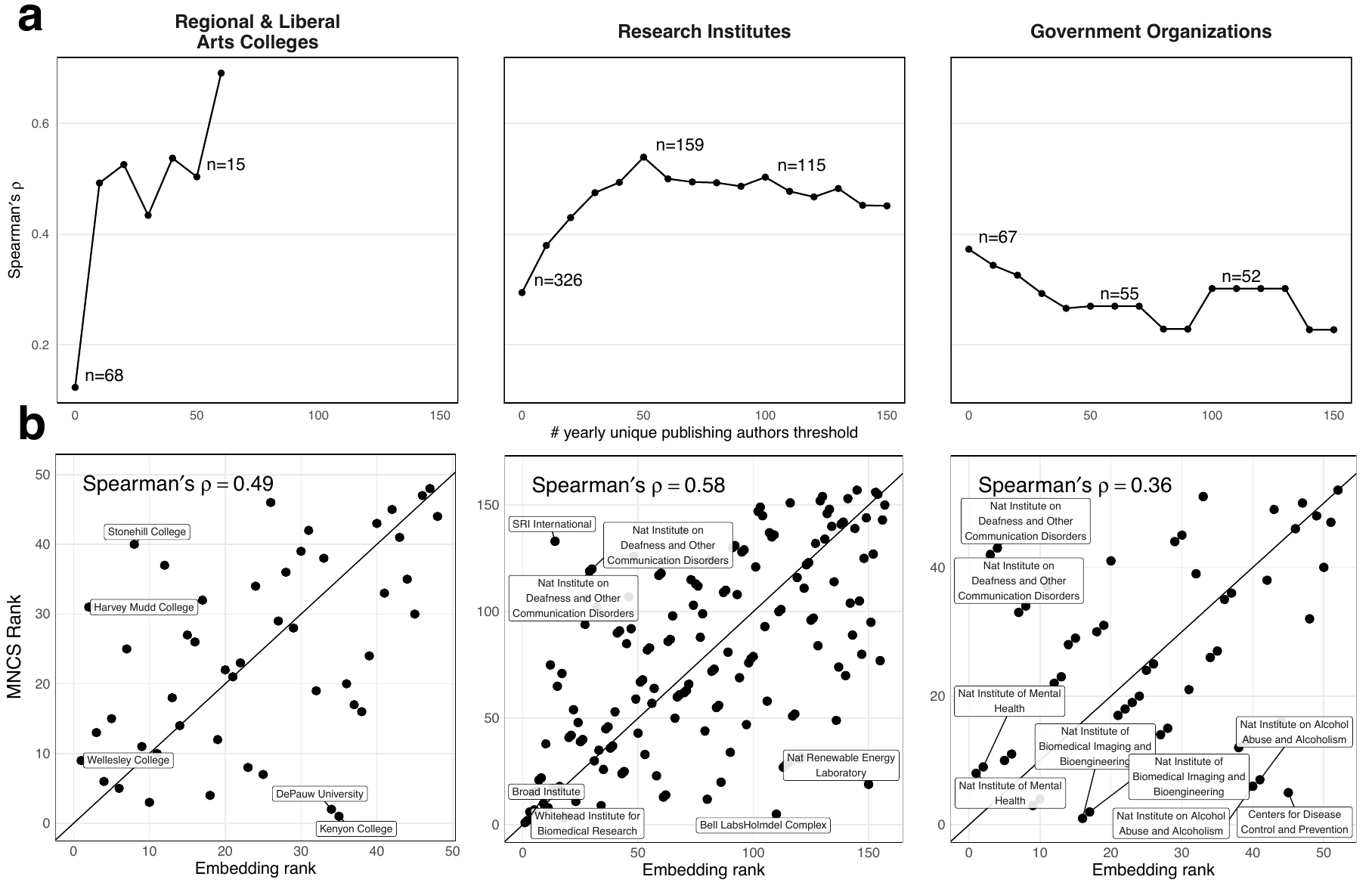}
	\caption{
		\textbf{SemAxis reconstructs publication impact in non-university sectors.}
		Comparison between the ranking of organizations in each non-university sector by their citation impact and the embedding rank.
		Citation impact is calculated as the mean-normalized citation score using papers published in the Web of Science database between 2008 and 2019.
		The embedding rank is derived by first projecting non-university organizations onto the SemAxis axis formed with poles defined using the top five to geographically-matched bottom five universities ranked by the 2018 Times Higher Education ranking of U.S. Universities.
		\textbf{a} Shows how the correlation between the citation impact and SemAxis rankings differ while varying the size threshold for including an organization.
		Size is calculated as the mean annualized number of unique authors publishing with that organization.
		Annotations show the number of organizations remaining at thresholds of 0, 50, and 100.
		\textbf{b}. Comparison of organizations using a size threshold of 10 for regional and liberal arts colleges, and 50 for research institutes and government organizations;
		these thresholds were chosen as points thresholds of stability in \textbf{a}.
		The impact rank is correlated with the embedding rank for regional and liberal arts colleges with Spearman's $\rho = 0.49\,(n = 48)$, research institutes with Spearman's $\rho = 0.58\,(n = 159)$, and for government organizations with Spearman's $\rho = 0.36\,(n = 55)$.
		All correlations are significant with $p < 0.001$.
	}
	\label{fig:supp:sector_semaxis_impact}
\end{figure}

%
%
\clearpage
\begin{figure}
    \centering
    \includegraphics[width=\textwidth]{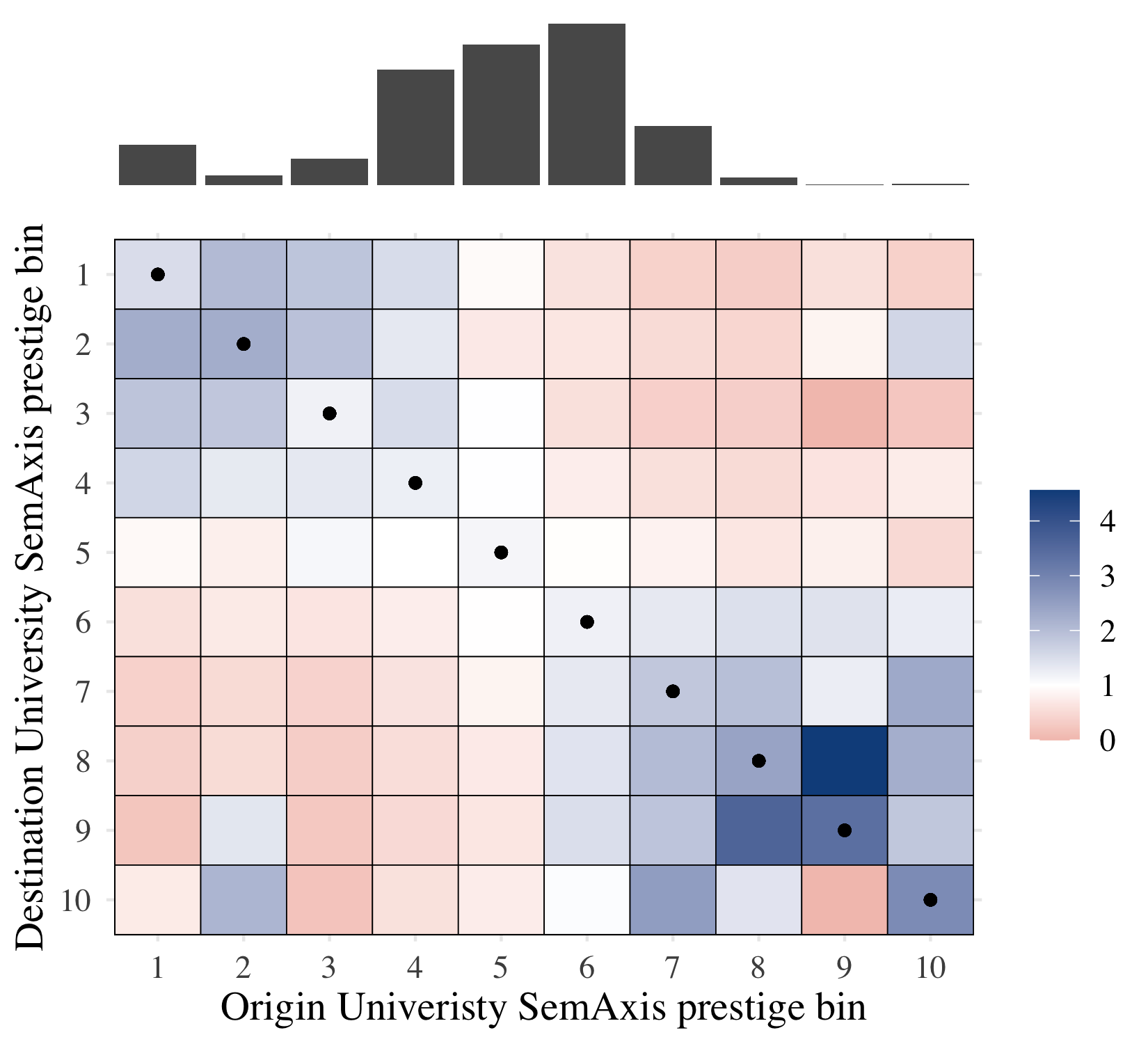}
    \caption{
    \textbf{Transitions by SemAxis prestige.}
    Information is sourced for over 100,000 mobile researchers in the United States for whom clear migrations could be identified.
    SemAxis rank is derived by setting the ``elite'' pole set according to the top 5 universities by the Times ranking.
    The continuous range of the SemAxis rank is divided into 10 equally spaced bins.
    Each cell shows the normalized transitions between the bins of the origin and destination university following the procedure outlined in previous research~\autocite{park_2019_labor}.
    Blue indicates that the number of transitions between the bins' is higher than expected when assuming a null model based only on their sizes and total in and out flows, whereas red indicates lower than expected transitions. 
    Dots inside a cell indicate the diagonal. 
    The histogram at the top provides a rough estimation of the distribution of total number of researchers affiliated with universities in each bin. 
    We observe two blocks where internal mobility is over-represented among the most elite, and the least elite universities, whereas middle-ranking universities (accounting for the majority of researchers) show weak affinity in either direction. 
    }
    \label{fig:supp:semaxis-individual}
\end{figure}

%
%

%
%
\begin{figure}[hp!]
	\centering
	\includegraphics[width=\textwidth]{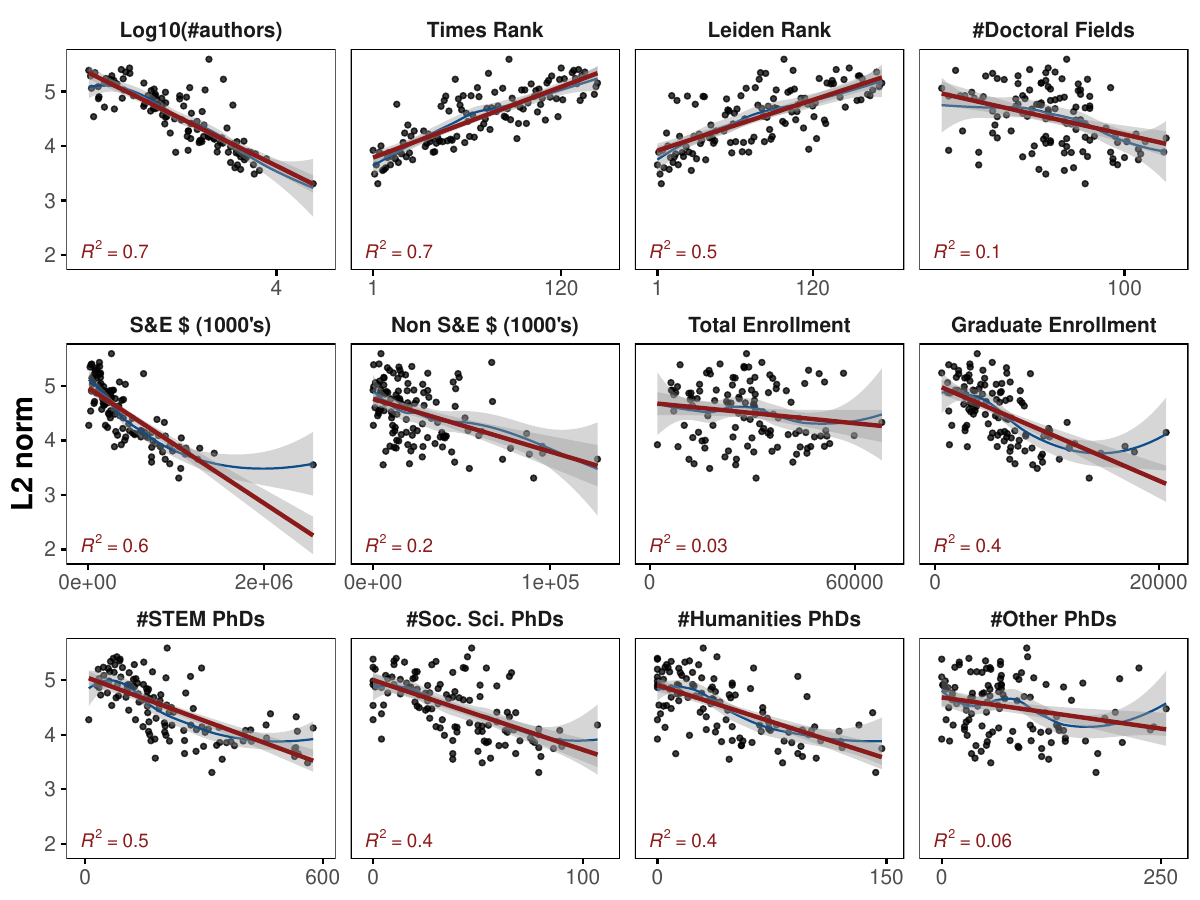}
	\caption{
		\textbf{Factors relating to the L2 norm of vectors for U.S. universities}
		Correlation between the L2 norm of organization embedding vectors of U.S. universities and characteristics of U.S. universities.
		Dots correspond to organizations.
		The red line is the line of  the best fit with corresponding 99\% confidence intervals.
		Red text is the regression estimate.
		The blue line is the loess regression line with 99\% confidence intervals.
		Number of authors is the average annual count of unique mobile and non-mobile authors.
		Rankings are derived from the Times Ranking of World Universities, and the Leiden Rankings of Universities.
		Remaining variables come from the Carnegie Classification of Higher Education Institutions.
		The factors that best explain $s_i$ are the number of authors, the rank, the amount of Science and Engineering (S\&E) funding, and the number of doctorates granted.
	}
	\label{fig:supp:length_vs_metaInfo}
\end{figure}

%
%
\begin{figure}[h!]
	\centering
	\includegraphics[width=0.8\textwidth]{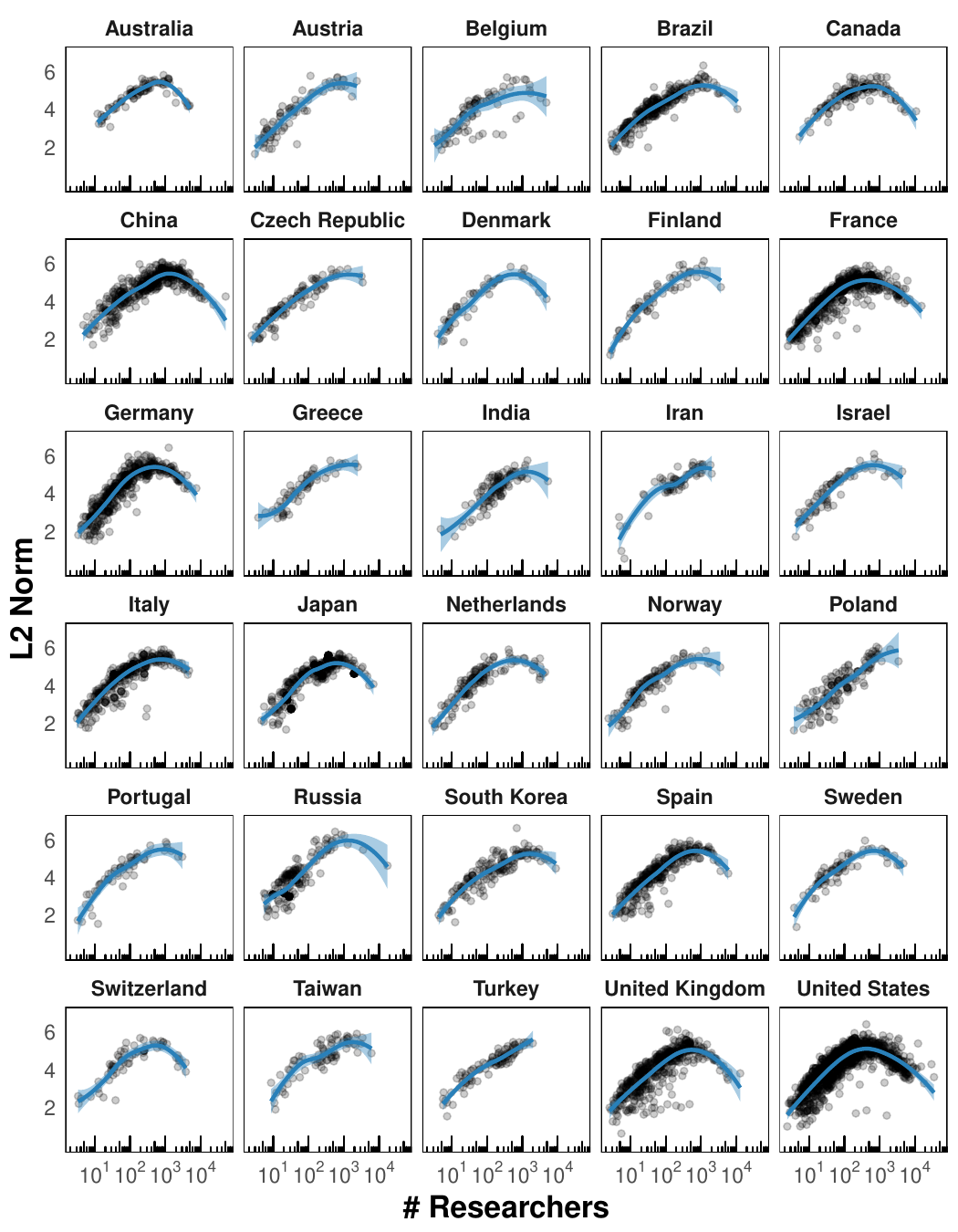}
	\caption{
		\textbf{Concave-curve repeats across most of 30 countries with most researchers.}
		Size (L2 norm) of organization embedding vectors compared to their number of researchers for U.S. universities.
		Loess regression line is shown for each country with 99\% confidence intervals.
		Countries shown are the 30 with the largest number of total unique mobile and non-mobile researchers.
	}
	\label{fig:concave30}
\end{figure}

%
%
\begin{figure}[h!]
	\centering
	\includegraphics[width=\textwidth]{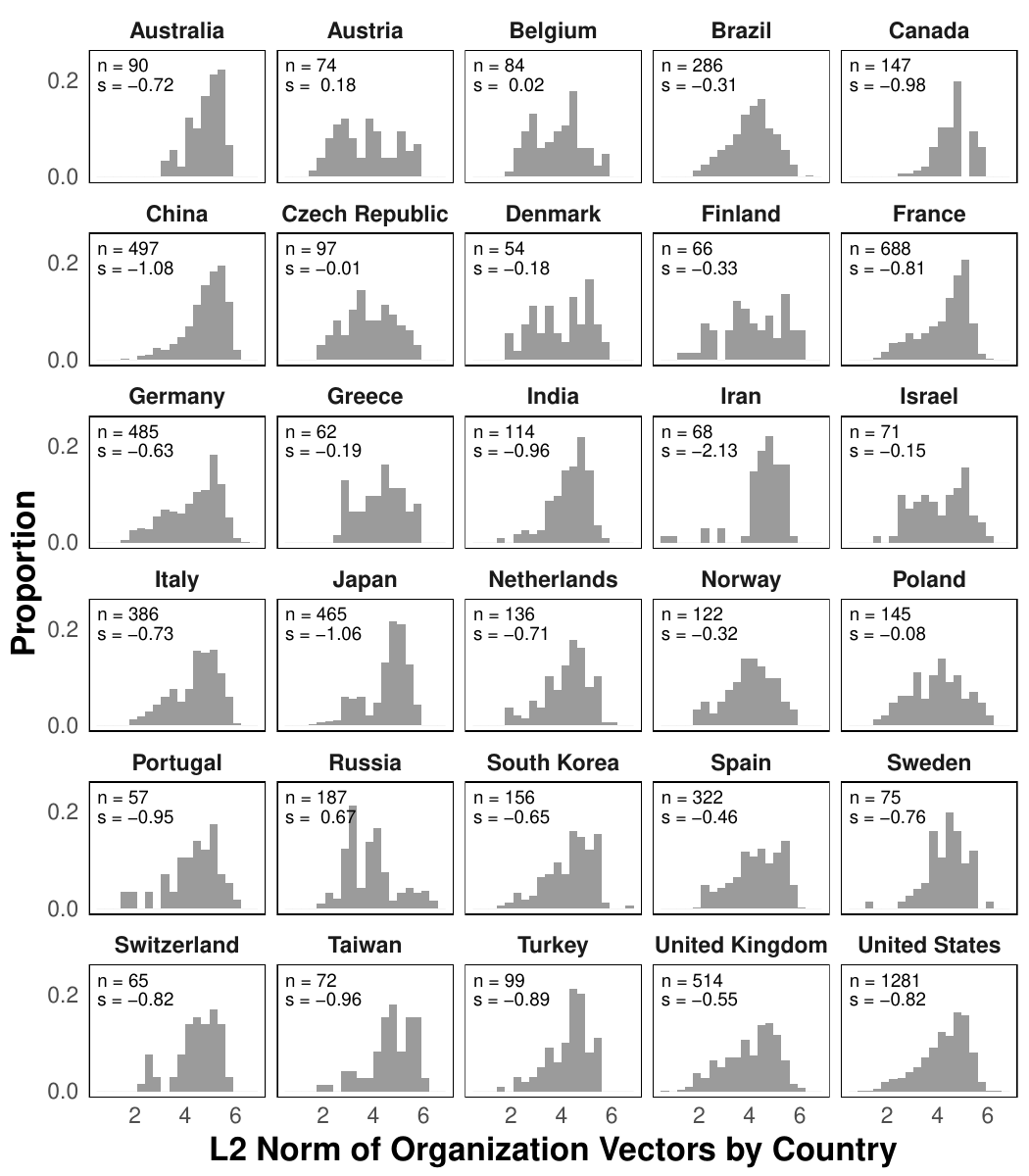}
	\caption{
		\textbf{Distribution of organization embedding vector norms by country.}
		Histogram showing the distribution of L2 norm values of organization embedding vectors in each of the 30 countries with the largest number of total unique mobile and non-mobile researchers.
		Text in each panel shows the number of organizations in the country (n) and the GINI index of inequality of the distribution (g);
		a small GINI index indicates that the L2 norms of organizations are more balanced, whereas a high GINI value indicates that they are more unequal.
	}
	\label{fig:norm_dist}
\end{figure}

\begin{figure}[h]
	\centering
	\includegraphics[width=\textwidth]{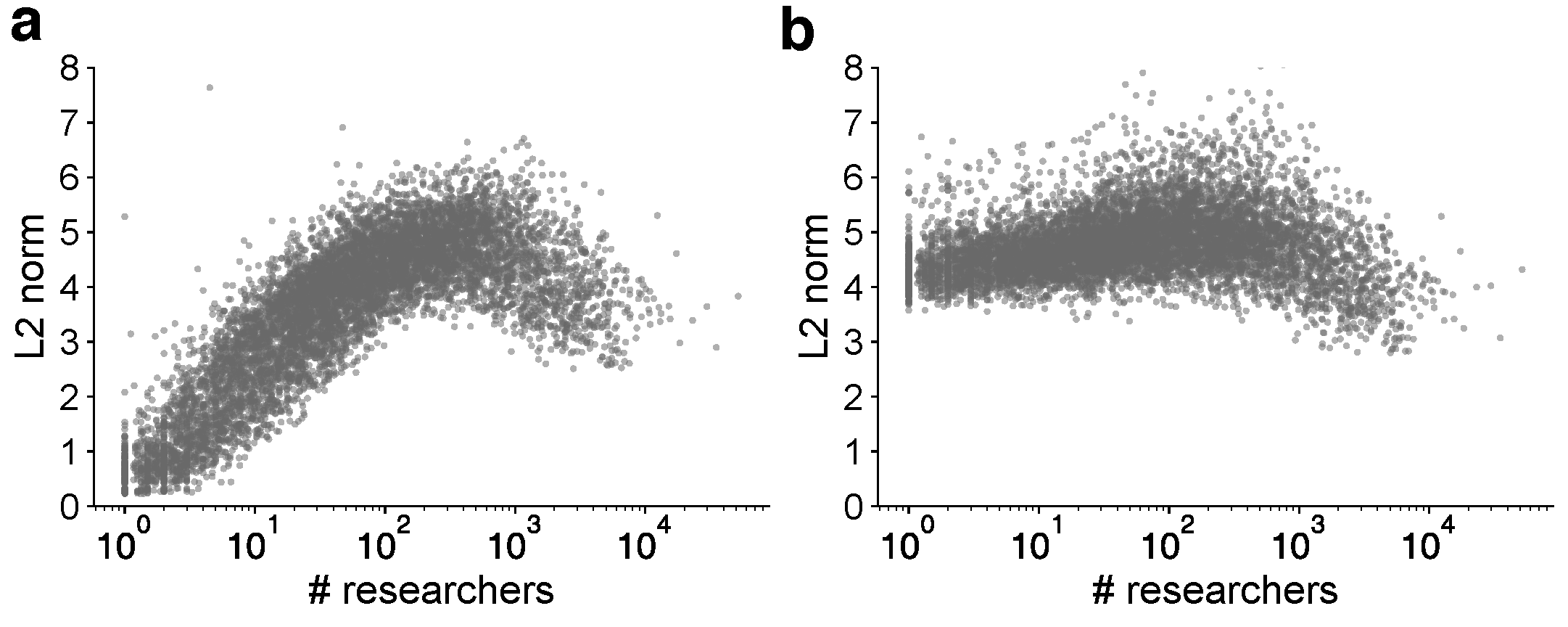}
	\caption{
		\textbf{Relationship between organization size and vector L2-norm for synthetic trajectories}
        The x-axis corresponds to the size of an organization, measured as the number of distinct publishing researchers, taken from the real scientific migration data. 
        The y-axis shows the L2-norm calculated for each organization's vector embedded using \textit{word2vec}.
        Each panel shows vectors generated on synthetic trajectories generated by a process of initializing random walkers (with five steps) on the real weighted organization co-occurence network.
        (a) random walkers are initialized at each node proportional to the size of the organization.
        (b) random walkers are initialized with uniform probability across all organizations.
	}
	\label{fig:supp:real_network}
\end{figure}

\begin{figure}[h]
	\centering
	\includegraphics[width=\textwidth]{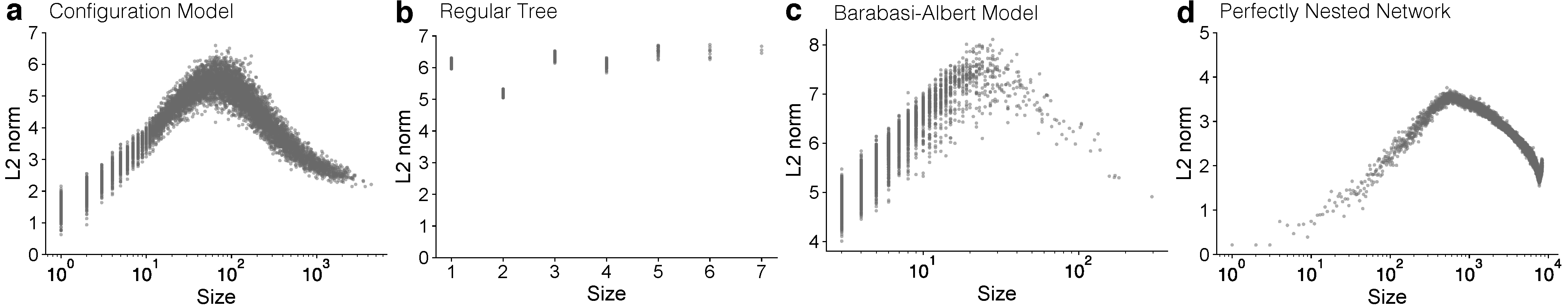}
	\caption{
		\textbf{Experiments modelling synthetic network data}. 
        The relationship between the size of a node (its degree) and the L2-norm of its embedding vector, generated through the process outlined in the text. 
        Shown for \textbf{a.} the configuration model, \textbf{b.} regular tree, \textbf{c.}  Barabási-Albert model, and \textbf{d.} a perfectly nested network. 
        The regular tree shows the results split by the depth of each node.
        The curvilinear relationship emerges in all but the regular tree model, and is particularly prominent in the perfectly nested network.
	}
	\label{fig:supp:model_network}
\end{figure}

\end{document}